\documentclass[table]{gtech}

\PassOptionsToPackage{table,usenames,dvipsnames}{xcolor}

\usepackage[T1]{fontenc}
\usepackage[utf8]{inputenc}

\RequirePackage{tgpagella}
\RequirePackage{mathpazo}
\RequirePackage{inconsolata}
\usepackage{biolinum}

\usepackage{amsmath,amssymb,amsfonts,amsthm,bm}
\usepackage{nicefrac}
\usepackage{algorithm}
\usepackage{algpseudocode}

\usepackage{xcolor}
\usepackage[most]{tcolorbox}
\usepackage{colortbl}

\usepackage{graphicx}
\usepackage{wrapfig}
\usepackage{float}
\usepackage{subcaption}
\usepackage{caption}
\usepackage{adjustbox}
\usepackage{multicol}
\usepackage{fontawesome5}

\usepackage{array}
\usepackage{booktabs}
\usepackage{multirow}
\usepackage{bigdelim}
\usepackage{longtable}
\usepackage{tabularray}
\usepackage{tabulary}
\usepackage{makecell}
\usepackage{threeparttable}
\usepackage{tablefootnote}
\usepackage{arydshln}
\usepackage{tabularx}
\usepackage{siunitx}

\usepackage{enumitem}
\usepackage{url}
\usepackage{hyperref}
\usepackage{microtype}
\usepackage{xspace}

\usepackage{pifont}
\usepackage{bbding}
\usepackage{fontawesome5}

\usepackage{datatool}
\usepackage[normalem]{ulem} 

\theoremstyle{definition}
\newtheorem{definition}{Definition}[section]

\newcolumntype{C}[1]{>{\centering\arraybackslash}m{#1}}

\newcommand{\paratitle}[1]{\vspace{1.5ex}\noindent\textbf{#1}}

\newcommand{\ignore}[1]{}

\definecolor{CQColor}{rgb}{0.0,0.0,1.0}

\newlength\savewidth

\newcommand{\tablestyle}[2]{%
  \setlength{\tabcolsep}{#1}%
  \renewcommand{\arraystretch}{#2}%
  \centering\footnotesize
}

\newcommand{\NAME}{{\fontfamily{LinuxBiolinumT-LF}\selectfont\textbf{Falcon-2.0}}\xspace}

\newcommand{\Orbit}{\textbf{ORBIT}\xspace}

\newcommand{\icon}{\raisebox{-2pt}{\includegraphics[width=1.5em]{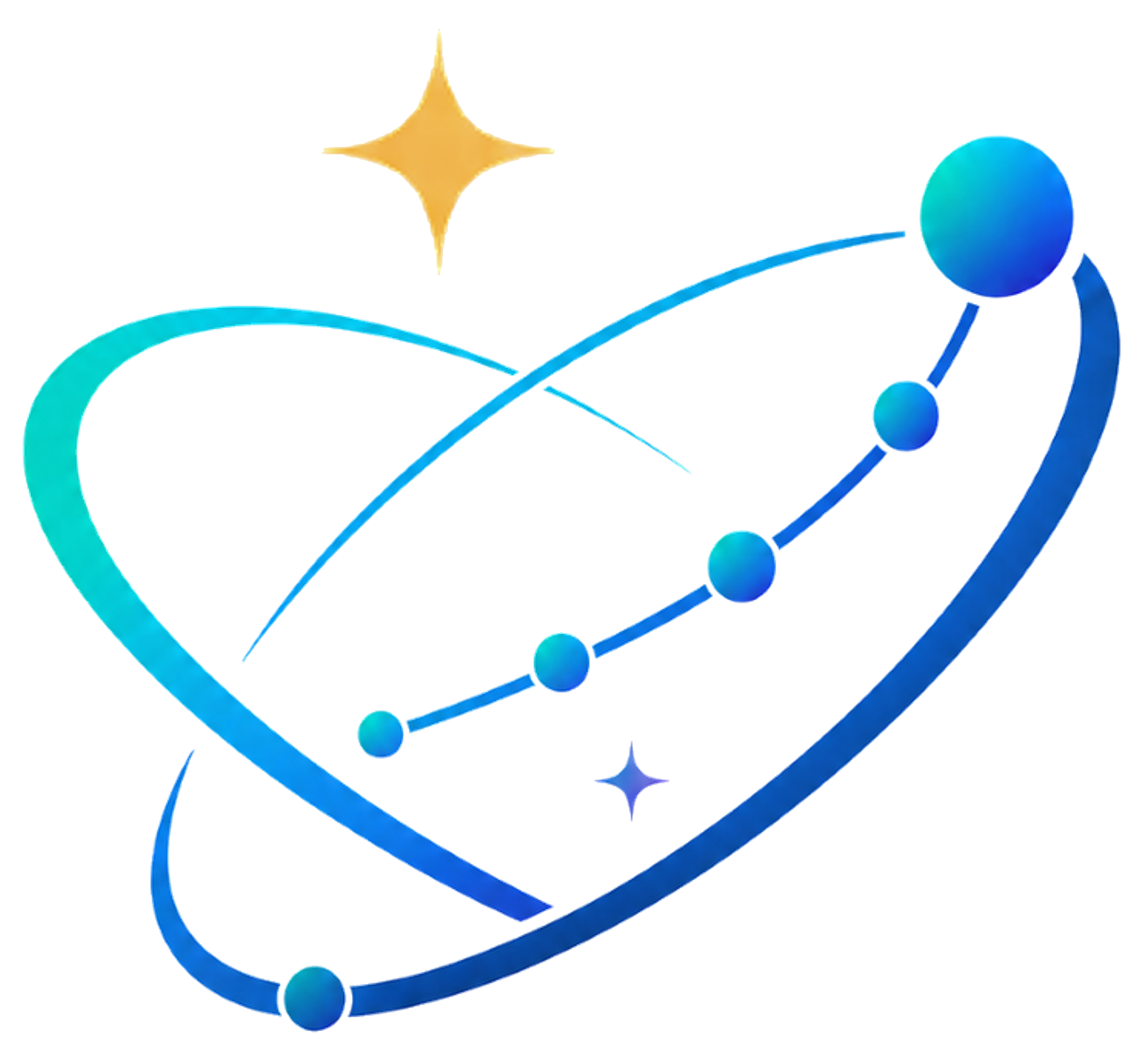}}\xspace}

\renewcommand{\title}[1]{%
  \newcommand{\titlelist}{{\huge\fontfamily{optimistic}\selectfont #1}}%
}

\title{
  \Large{\textcolor[HTML]{0369ff}{\NAME} Technical Report} \\[0.5em]
  \huge{\icon Into the \textcolor[HTML]{0369ff}{\Orbit} for Time Series: Training Regimes for Foundation Models}
}

\author[*]{Hongjie Xia}
\author[*]{Yiding Liu}
\author[*]{Yifan Hu}
\author[*]{Peiyuan Liu}
\author[\dagger]{Zewei Dong}

\contribution[*]{
  Equal Contribution
}
\contribution[\dagger]{Corresponding Author \\ \ \faEnvelope[regular]\ \email{\{xiahongjie.xhj,\ yiding.lyd,\ hyf476357,\ peiyuan.liu,\ zewei.dong\}@ant-intl.com}}

\abstract{

Time series foundation models (TSFMs) have advanced primarily through architectural innovations such as group attention, flow matching, serial-token prediction, and mixture-of-experts scaling. 
However, the training regimes governing data exposure over large-scale heterogeneous corpora remain comparatively under-explored. Consequently, the effective pre-training distribution is often poorly controlled along four coupled axes: cross-domain imbalance, frequency-dependent context requirements, variable prediction horizons, and missingness. 
To address this problem, we introduce \Orbit{} (\textbf{O}mni-\textbf{R}ange \textbf{B}ootstrap \textbf{I}ncremental \textbf{T}raining), a training paradigm that explicitly controls the effective pre-training distribution. \Orbit{} comprises two key components. First, \textbf{Bootstrap Multi-Level Sampling} converts prescribed dataset weights into a global source allocation and constructs per-dataset sample indices through stochastic selection of time series records, target variables, context windows, and prediction horizons. Second, \textbf{Omni-Range Incremental Training} assembles and incrementally consumes these variable-length examples throughout a single training run, allowing diverse context lengths and prediction horizons to coexist without stage-specific schedules. 
Under \Orbit{}, we train \NAME{}, a deliberately simple univariate encoder-only Transformer featuring missingness-aware triple-channel patch tokenization and direct multi-patch quantile prediction. 
Beyond this backbone, we further investigate how representation rank evolves through Transformer depth and introduce \textbf{Rank-Guided Cross-Depth Alignment}, a training-only objective that uses late-layer states as stop-gradient references for shallow representations and adds no inference cost. Our analysis relates the alignment error to the transfer of non-negligible spectral modes across depth. 
Evaluations on GIFT-Eval and fev-bench show that \NAME{} achieves strong zero-shot forecasting performance across diverse domains and frequencies. Ablations confirm the importance of stochastic sample construction and simultaneous exposure to diverse context lengths and prediction horizons.
\NAME{} is publicly released to support future research.
}

\date{\today\vspace{-1mm}}
\gtechdata[Code]{\url{https://github.com/ant-intl/Falcon-TST}\vspace{-1mm}}
\gtechdata[Model]{\url{https://pypi.org/project/falcon-tst/}}

\begin{document}

\maketitle

\section{Introduction}
\label{sec:intro}


Time series foundation models (TSFMs) have recently emerged as a promising paradigm for learning generalizable forecasting capabilities from large-scale and heterogeneous temporal corpora~\citep{woo2024moirai,das2023timesfm,ansari2024chronos,cohen2026toto,olber2025tirex}. Compared with traditional task-specific forecasting models~\citep{zhou2022fedformer,hu2025timefilter,hu2025adaptive,liu2024itransformer, nie2023patchtst}, TSFMs aim to provide zero-shot or few-shot forecasting across diverse domains, frequencies, and forecasting settings. 
Recent progress has been driven primarily by architectural innovations, including improved tokenization strategies, attention mechanisms, generative objectives, long-context modeling, and large-scale model scaling~\citep{liu2026timers1, ansari2025chronos2, grinsztajn2026tabpfn3,khwaja2026toto2,podest2026tirex2,liu2026falconx}. 
Although these developments have substantially improved forecasting performance, the training regimes that govern the exposure of heterogeneous temporal corpora during pre-training remain comparatively under-explored.

Unlike language or vision foundation models, time series corpora naturally exhibit substantial heterogeneity in their sources, temporal resolutions, observation patterns, and forecasting requirements~\citep{kottapalli2025foundationmodelstimeseries,hulandscape}. Importantly, the capability of a TSFM is not solely determined by the nominal size of the collected corpus, but by the \emph{effective pre-training distribution} induced by the training pipeline~\citep{syed2026position}. This distribution is shaped by which datasets, time series records, variables, context windows, and forecasting horizons are sampled, as well as how missing observations and invalid targets are handled throughout optimization~\citep{yeh2023toward}. Therefore, designing a scalable TSFM requires not only increasing corpus size or model capacity, but also controlling the distribution of training experiences presented to the model.

This challenge introduces several coupled difficulties for large-scale TSFM pre-training. First, heterogeneous datasets often contain highly imbalanced sources, causing dominant datasets to determine the optimization trajectory while under-represented domains receive insufficient exposure~\citep{shao2024exploring,shao2025heterogeneity}. Second, different temporal resolutions and recording frequencies naturally require different historical context ranges, making fixed context training insufficient for broad temporal generalization~\citep{zhang2022self,chung2024time}. Third, forecasting tasks involve diverse prediction horizons, while existing models often rely on horizon-specific training or multi-stage context extension~\citep{lim2021temporal}. Finally, missing observations are pervasive in real-world time series and require consistent treatment across normalization, tokenization, attention computation, and loss optimization~\citep{hu2026existence,che2018recurrent}. These challenges indicate that effective TSFM training requires a principled strategy for controlling data exposure, temporal coverage, and supervision quality simultaneously.

To address these challenges, we introduce \Orbit{} (\textbf{O}mni-\textbf{R}ange \textbf{B}ootstrap \textbf{I}ncremental \textbf{T}raining), a training paradigm designed to explicitly control the effective pre-training distribution of heterogeneous time series corpora. 
ORBIT separates the construction of forecasting examples from their consumption during optimization through two complementary components.
\ding{182} Bootstrap Multi-Level Sampling controls data exposure hierarchically. At the corpus level, prescribed domain-aware dataset weights are converted into an ordered global training stream using a low-discrepancy greedy blending rule. Within each retained dataset, Bootstrap Stochastic Sampling constructs an offline sample index through four stochastic selections: a valid time series record, a target variable, a context window, and a compatible prediction horizon. Each indexed example is represented by a five-tuple containing its record, variable, context start, context end, and prediction horizon. In contrast to sequential traversal and sliding-window enumeration used in existing TSFM training pipelines~\citep{goswami2024moment,shi2024timemoe,liu2025timerxl}, this construction explicitly randomizes forecasting configurations while preserving reproducibility and efficient random access.
\ding{183} Omni-Range Incremental Training consumes the resulting variable-length examples throughout a single training run. Context lengths and prediction horizons are sampled once during sample-index construction rather than resampled during optimization. During batch assembly, contexts are left-padded and targets are right-padded to their respective mini-batch maxima, with attention and loss masks excluding unsupported positions. The globally interleaved sample stream is then consumed incrementally from memory-mapped storage. Short and long contexts, together with short and long prediction horizons, therefore coexist throughout optimization, avoiding the separate context-extension stages or horizon-specific training schedules adopted by several existing approaches~\citep{ansari2025chronos2,liu2026timers1,shi2024timemoe}.

Under \Orbit{}, we train \NAME{}, a deliberately simple encoder-only Transformer designed to investigate the effectiveness of the proposed training paradigm beyond architectural scaling alone. 
Rather than introducing highly specialized modules or task-specific architectural branches, \NAME{} adopts a unified patch-based forecasting framework that provides a simple and scalable interface for heterogeneous pre-training following the univariate encoder formulation of Chronos-2~\citep{ansari2025chronos2}. Specifically, \NAME{} treats each variable as an independent temporal sequence and employs a missingness-aware triple-channel patch tokenization scheme to distinguish observed values, missingness patterns, and temporal information. This design allows the model to consistently process incomplete observations and diverse training examples generated by \Orbit{}. Furthermore, \NAME{} adopts parallel patch prediction, enabling direct forecasting of multiple future segments within a unified objective and supporting the variable prediction horizons sampled during training. The simplicity of \NAME{} allows us to isolate and evaluate the contribution of training strategies rather than relying on architectural complexity.

Beyond the training distribution, we investigate how representations evolve across Transformer depth. Deep layers often capture richer and more diverse temporal structures, while shallow representations may suffer from limited expressive capacity~\citep{yu2026understanding}. Motivated by this observation, we introduce a training-only \textbf{Rank-Guided Cross-Depth Alignment} objective, which uses late-layer representations as stop-gradient teachers to regularize shallow layers. This alignment improves representation consistency across depth without introducing additional inference cost~\citep{hu2025bridging,jiang2025no}.
We further show that sufficiently small alignment error bounds the perturbation between centered shallow and deep representations and, under an explicit spectral-separation condition, prevents the shallow representation from losing non-negligible modes present in the deep representation.
Moreover, we evaluate \NAME{} trained with \Orbit{} on large-scale heterogeneous forecasting benchmarks, including GIFT-Eval~\citep{aksu2024gift} and fev-bench~\citep{shchur2025fev}. The results demonstrate strong zero-shot forecasting capability across diverse domains and frequencies, validating the effectiveness of the proposed training paradigm and the complementary benefit of cross-depth representation alignment.

The principal contributions of this report are:

\begin{enumerate}[leftmargin=*,itemsep=0pt]




    \item We identify effective pre-training distribution control as a critical but under-explored factor in time series foundation model development, and introduce \Orbit{} (\textbf{O}mni-\textbf{R}ange \textbf{B}ootstrap \textbf{I}ncremental \textbf{T}raining), a unified training paradigm for heterogeneous temporal corpora.

    \item We propose \textbf{Bootstrap Multi-Level Sampling} and \textbf{Omni-Range Incremental Training}, which respectively control heterogeneous data exposure and enable single-stage training over diverse temporal contexts and forecasting horizons.

    \item We develop \NAME{}, a simple encoder-only Transformer trained under \Orbit{}, demonstrating that carefully designed training regimes can unlock strong forecasting capability without relying on excessive architectural complexity.

    \item We introduce \textbf{Rank-Guided Cross-Depth Alignment}, a training-time representation regularization method that transfers information from deeper Transformer layers to shallow layers without additional inference overhead.
\end{enumerate}


\section{Related Work}
\label{sec:related}

\begin{table*}[t]
\centering
\footnotesize
\renewcommand{\arraystretch}{1.15}
\caption{Comparison of representative time series foundation models from the perspective of model design.}
\label{tab:tsfm_design}
\begin{tabularx}{\textwidth}{
l
>{\raggedright\arraybackslash}p{3.75cm}
>{\raggedright\arraybackslash}p{4.75cm}
>{\raggedright\arraybackslash}X
}
\toprule
\textbf{Model} 
& \textbf{Architecture} 
& \textbf{Input Representation} 
& \textbf{Forecasting Formulation} \\
\midrule

Chronos
& Encoder-decoder
& Discretized univariate time series via quantization
& Auto-regressive probabilistic forecasting \\

MOMENT
& Encoder-only
& Patch-based input 
& Masked reconstruction \\

TimesFM
& Decoder-only
& Patch-based input 
& Auto-regressive next-patch prediction \\

Timer
& Decoder-only 
& Patch-based input 
& Auto-regressive next-patch prediction \\

Timer-XL
& Decoder-only 
& Patch-based input 
& Auto-regressive next-patch prediction \\

Timer-S1
& Decoder-only with MoE and STP blocks 
& Patch-based input
& Serial-token prediction with multi-patch quantile forecasting \\

Time-MoE
& Decoder-only with MoE 
& Point-wise tokenization
& Auto-regressive next-patch prediction with multi-resolution forecasting heads \\

Moirai
& Encoder-only 
& Patch-based input with any-variate processing 
& Probabilistic distribution forecasting \\

Moirai2
& Decoder-only
& Patch-based input
& Quantile forecasting with multi-patch prediction \\

TTM
& MLP-Mixer
& Patch-based input with adaptive patching 
& Direct forecasting \\

GTM
& Decoder-only with Fourier attention 
& Patch-based input with frequency-aware 2D positional encoding 
& Unified reconstruction and auto-regressive forecasting \\

Sundial
& Decoder-only 
& Patch-based input 
& Auto-regressive next-patch prediction \\

Chronos-2
& Encoder-only with group attention 
& Patch-based grouped targets and covariates with meta features
& Quantile forecasting \\

TabPFN-TS
& Tabular foundation model
& Time series with tabular features
& Probabilistic forecasting via tabular regression \\

TiRex
& Decoder-only xLSTM 
& Patch-based input 
& Quantile forecasting with contiguous patch masking \\

TiRex-2
& Decoder-only xLSTM with asymmetric grouped attention
& Patch-based multivariate input with past and future-known covariates 
& Quantile forecasting \\

Toto
& Decoder-only with factorized space-time attention 
& Patch-based input with causal scaling 
& Student-\emph{t} mixture robust probabilistic forecasting \\

Toto-2.0
& Decoder-only with factorized space-time attention 
& Patch-based input with causal scaling 
& Quantile forecasting with contiguous patch masking \\

\midrule
\textcolor[HTML]{0369ff}\NAME{}
& Encoder-only
& Patch-based input with meta features
& Quantile forecasting under the \textcolor[HTML]{0369ff}\Orbit \\
\bottomrule
\end{tabularx}
\end{table*}

Time series foundation models (TSFMs) have rapidly evolved along architectural, algorithmic, and methodological dimensions. We organize this section around two core themes: the model architectures themselves, and the data and training regimes that enable their generalization.

\subsection{Time Series Foundation Models}
\label{sec:related_foundation}

Recent progress on time series foundation models (TSFMs) has moved forecasting from task-specific training toward large-scale, cross-domain pre-training with strong zero-shot transferability, while differing mainly in tokenization, architectural inductive bias, probabilistic modeling, and efficiency ~\citep{kottapalli2025foundationmodelstimeseries, Liang_2024}. Table~\ref{tab:tsfm_design} provides a structured comparison of existing models along key design dimensions.

Early representative approaches such as Chronos~\citep{ansari2024chronos} cast time series as a discrete language via scaling and quantization and train language-model-style architectures auto-regressively.  MOMENT~\citep{goswamimoment} learns general-purpose time series representations through masked pretraining, enabling transfer to diverse downstream tasks via task-specific fine-tuning. TimesFM~\citep{das2023timesfm} shows that a decoder-only Transformer trained on patched continuous-valued time series can already serve as a strong general-purpose forecaster. Building on this generative line, Timer~\citep{liu2024timer}, Timer-XL~\citep{liu2025timerxl}, and Timer-S1~\citep{liu2026timers1} progressively scale decoder-based TSFMs toward long-context, multivariate, and sparse mixture-of-experts settings, with Timer-S1 further proposing serial-token prediction to better align model computation with the inherently sequential nature of long-horizon forecasting. Time-MoE~\citep{shi2024timemoe} scales auto-regressive decoder-only TSFMs to 2.4B parameters through sparse mixture-of-experts design. In parallel, Moirai~\citep{woo2024moirai} formulates universal forecasting through a masked encoder with any-variate attention and probabilistic outputs, while Moirai2~\citep{liu2025moirai2} revisits this design and demonstrates that a simpler decoder-only quantile forecasting pipeline can improve both efficiency and performance. Other models explore more specialized directions. TTM~\citep{ekambaram2024ttm} shows that compact mixer-based architectures can also achieve strong transfer. GTM~\citep{woo2024gtm} pushes TSFMs toward generative-task-agnostic modeling with frequency-domain attention and unified adaptation across forecasting, imputation, and anomaly detection. Sundial~\citep{das2024sundial} introduces continuous generative forecasting through flow matching. Chronos-2~\citep{ansari2025chronos2} extends TSFMs from univariate to universal forecasting with multivariate structure and covariates via group attention. TabPFN-TS~\citep{hoo2025tables} reformulates forecasting as a tabular regression problem via temporal featurization. TiRex~\citep{olber2025tirex} leverages xLSTM to combine in-context learning with strong state tracking for long-horizon zero-shot forecasting. Extending this recurrent design, TiRex-2~\citep{podest2026tirex2} generalizes TiRex to multivariate forecasting with past and future-known covariates. Toto~\citep{cohen2026toto} designs an observability-oriented TSFM with causal scaling and factorized time–variate attention. Toto 2.0~\citep{khwaja2026toto2} further designs contiguous patch masking and scales the decoder-only time–variate Transformer from 4M to 2.5B parameters. 

Overall, existing work suggests that TSFMs are evolving along two complementary directions: one focuses on scaling model size, context length, and training corpus diversity to improve universality, while the other emphasizes time-series-native inductive biases and broader task generalization to make foundation models more robust, efficient, and practically useful.

\subsection{Training Regimes in Time Series Foundation Models}

Beyond architecture, TSFMs also differ substantially in how they define the \emph{effective pretraining distribution}. As summarized in Table~\ref{tab:tsfm_training}, we characterize the training regimes of representative TSFMs along five key dimensions: dataset weighting, sequence selection, variable selection, window sampling, and prediction-length assignment. Under this view, early models such as MOMENT~\citep{goswami2024moment}, Time-MoE~\citep{shi2024timemoe}, and Timer-XL~\citep{liu2025timerxl} largely rely on sequential traversal or sliding-window enumeration, together with uniform, sequence-count-aware, and window-count-aware dataset weighting, respectively. When coupled with deterministic enumeration, these pipelines may skew the effective training distribution toward datasets with greater data volume, more constituent sequences, or more candidate training windows, thereby exacerbating cross-dataset imbalance. In contrast, more recent models increasingly adopt stochastic sampling strategies. For instance, Moirai~\citep{woo2024moirai} combines length-aware dataset weighting with probabilistic sequence sampling, variable subsampling, and random cropping. However, the length-aware dataset weighting setting may still favor datasets with greater aggregate sequence length during the sampling process. The Toto family~\citep{cohen2026toto,khwaja2026toto2} traverses sequences sequentially and constructs univariate or multivariate samples through random variable composition, random window sampling. Chronos-2~\citep{ansari2025chronos2} adopts random sampling at both the sequence and window levels and supports grouped multivariate or covariate-informed inputs. By contrast, \NAME{} adopts domain-aware weighting to explicitly control source-level exposure, preventing high-volume datasets from dominating the training distribution while enabling under-represented domains to contribute meaningful supervision. These developments reflect a broader shift from static dataset traversal toward more stochastic, length-diverse, and compositionally flexible sampling schemes.

A further key distinction among TSFMs concerns the \emph{optimization and training pipeline design}. MOMENT~\citep{goswami2024moment} follows the encoder-only masked-reconstruction paradigm, while decoder-style models such as Time-MoE~\citep{shi2024timemoe} and Timer-XL~\citep{liu2025timerxl} adopt auto-regressive next-patch prediction. Recent TSFMs have increasingly moved beyond point forecasting toward probabilistic forecasting. Moirai~\citep{woo2024moirai} optimizes likelihood over parametric probabilistic outputs, whereas latest models such as Chronos-2~\citep{ansari2025chronos2}, TiRex-2~\citep{podest2026tirex2}, and Moirai 2.0~\citep{liu2025moirai2} favor direct quantile regression. In parallel, normalization and training-inference alignment have become increasingly important. In most cases, RevIN~\citep{kim2022revin} is widely used for stabilizing distribution of input data~\citep{goswami2024moment, liu2024timer, shi2024timemoe}. Chronos-2~\citep{ansari2025chronos2} adopts masked instance normalization with optional arcsinh scaling to reduce the influence of outliers. Moirai~\citep{woo2024moirai} uses packed standardization, and Toto~\citep{cohen2026toto} introduces causal patch-wise scaling for nonstationary observability data. Several recent models also move beyond single-stage pretraining, most notably Timer-S1~\citep{liu2026timers1}, which uses pretraining, continued pretraining, and long-context extension, and Chronos-2~\citep{ansari2025chronos2}, which likewise separates base training from later capability refinement. Taken together, these results suggest that progress in TSFMs depends not only on larger models, but also on the joint design of sampling strategy, forecasting objective, normalization scheme, and inference-consistent training procedure.

\begin{table}[t]
\centering
\footnotesize
\renewcommand{\arraystretch}{1.35}
\caption{Training regimes comparison of data construction and sampling strategies in representative TSFMs.}
\label{tab:tsfm_training}
\begin{tabularx}{\textwidth}{
l
>{\raggedright\arraybackslash}p{2.0cm}
>{\raggedright\arraybackslash}p{2.75cm}
>{\raggedright\arraybackslash}p{2.75cm}
>{\raggedright\arraybackslash}p{2.5cm}
>{\raggedright\arraybackslash}X
}
\toprule
\textbf{Model} 
& \textbf{Dataset}\newline\textbf{Weighting} 
& \textbf{Sequence}\newline\textbf{Selection} 
& \textbf{Variable}\newline\textbf{Selection} 
& \textbf{Window}\newline\textbf{Sampling} 
& \textbf{Prediction}\newline\textbf{Length} \\
\midrule
MOMENT$^{1}$
& Uniform
& Sequential traversal 
& All variates retained 
& Sliding window with stride 
& Fixed length \\

Time-MoE$^{2}$
& Sequence-count-aware
& Sequential traversal 
& Univariate samples 
& Non-overlapping sliding window 
& Next-patch horizon \\

Timer-XL$^{3}$
& Window-count-aware
& Sequential traversal 
& All variates retained 
& Sliding window with stride 
& Next-patch horizon \\

Moirai, Moirai2$^{4}$
& Length-aware 
& Probabilistic sampling 
& Random composition 
& Random cropping 
& Random sampling \\

Toto, Toto 2.0$^{5}$
& Unknown
& Sequential traversal 
& Random composition 
& Random sampling 
& Next-patch horizon \\


Chronos-2$^{6}$
& Unknown
& Random sampling 
& All variates retained 
& Random sampling 
& Fixed length \\

\midrule

\textcolor[HTML]{0369ff}\NAME{}
& Domain-aware
& \multicolumn{4}{c}{\textcolor[HTML]{0369ff}\Orbit: Hierarchical sampling with omni-range contexts and horizons} \\
\bottomrule
\end{tabularx}

\vspace{2pt}
\raggedright
\footnotesize
$^{1}$MOMENT github repository: \url{https://github.com/moment-timeseries-foundation-model/moment-research} \\
$^{2}$Time-MoE github repository: \url{https://github.com/Time-MoE/Time-MoE} \\
$^{3}$Timer-XL github repository: \url{https://github.com/thuml/OpenLTM} \\
$^{4}$Moirai family github repository: \url{https://github.com/SalesforceAIResearch/uni2ts} \\
$^{5}$Toto family github repository: \url{https://github.com/DataDog/toto} \\
$^{6}$Chronos-2 github repository: \url{https://github.com/amazon-science/chronos-forecasting}
\end{table}

\medskip

\noindent\textbf{Positioning of \NAME{}.}
Existing time series foundation models have advanced primarily through architectural innovations---including group attention~\citep{ansari2025chronos2}, flow matching~\citep{das2024sundial}, serial-token prediction~\citep{liu2026timers1}, and any-variate attention~\citep{liu2025moirai2}---or through scaling data and model capacity~\citep{liu2025timerxl,shi2024timemoe}. However, the training methodologies governing exposure to heterogeneous temporal corpora remain comparatively under-explored, particularly how time series records, target variables, context windows, and forecasting horizons are sampled and how missing observations are handled during optimization~\citep{syed2026position,yeh2023toward}. \NAME{} targets this gap with a deliberately simple encoder-only Transformer, demonstrating that carefully designed training regimes can unlock strong forecasting capability without relying on excessive architectural complexity. Specifically, \textbf{Bootstrap Multi-Level Sampling} explicitly controls source-level exposure and constructs diverse forecasting tasks, while \textbf{Omni-Range Incremental Training} continuously exposes the model to diverse context lengths and forecasting horizons within a single training stage, with missing observations handled consistently throughout the pipeline.

\section{Model Architecture}
\label{sec:arch}

\subsection{Overview}
\label{sec:arch_overview}

\begin{figure}[t]
    \centering
    \includegraphics[width=\textwidth]{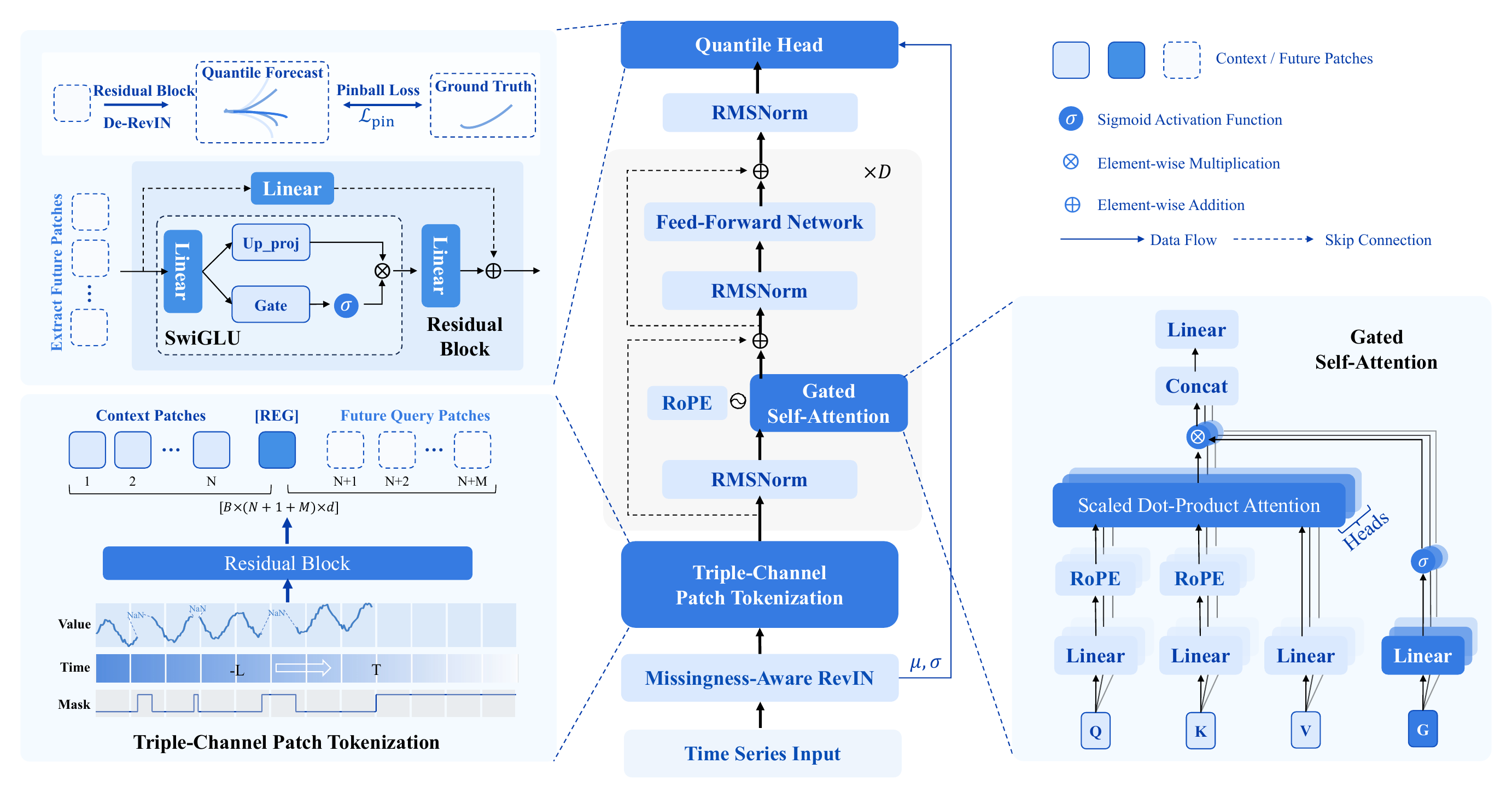}
    \caption{\NAME{} architecture. Following the univariate encoder formulation of Chronos-2~\citep{ansari2025chronos2}, the model represents each context or future query patch through temporal, value, and indicator channels. Context tokens, a learnable \texttt{REG} token, and future query tokens are jointly processed by a bidirectional Transformer encoder. A quantile head decodes all future patches in parallel.}
\label{fig:architecture}
\end{figure}

\NAME{} deliberately adopts a minimally specialized backbone, enabling a controlled assessment of the proposed training regime and representation-level mechanisms. As illustrated in Figure~\ref{fig:architecture}, the model follows the univariate encoder formulation of Chronos-2~\citep{ansari2025chronos2}: missingness-aware reversible instance normalization with an $\mathrm{arcsinh}$ transform, triple-channel patch tokenization using temporal, value, and indicator features, a learnable \texttt{REG} token, future query patches, and direct multi-patch quantile prediction. Unlike Chronos-2, \NAME{} processes target variables independently and does not use group attention. The encoder consists of Pre-RMSNorm Transformer blocks with RoPE, SwiGLU feed-forward layers, and output-gated self-attention.

Our contribution therefore does not lie in proposing an alternative backbone. Rather, this formulation establishes a controlled setting for studying two questions: how \Orbit{} shapes the effective pre-training distribution over heterogeneous time series (Section~\ref{sec:orbit}), and whether the depth-wise rank structure of a time series Transformer can be exploited through cross-depth representation alignment (Section~\ref{sec:arch_rank_alignment}).

\subsection{Input Representation and Backbone}
\label{sec:arch_backbone}

Let $\mathbf{x}=(x_1,\ldots,x_L)\in(\mathbb{R}\cup\{\bot\})^L$ denote a univariate context of length $L$, where $\bot$ represents a missing observation, and let $C$ denote the maximum admissible context length in time steps. The model uses a common patch size $P$ for context and future segments. After left padding, the context contains $N=\left\lceil\frac{L}{P}\right\rceil$ context patches.

\subsubsection{Missingness-Aware Reversible Instance Normalization}
\label{sec:arch_revin}

Let $\mathcal{V}_x=\{i:x_i\ \text{is observed}\}$ denote the observed-index set. When $\mathcal{V}_x$ is nonempty, \NAME{} follows Chronos-2~\citep{ansari2025chronos2} and computes instance statistics only from observed values:
\begin{equation}
\mu
=
\frac{1}{|\mathcal{V}_x|}
\sum_{i\in\mathcal{V}_x}x_i,
\qquad
\sigma
=
\max\!\left(
\sqrt{
\frac{1}{|\mathcal{V}_x|}
\sum_{i\in\mathcal{V}_x}(x_i-\mu)^2
},
\epsilon_{\mathrm{norm}}
\right).
\label{eq:missing_revin}
\end{equation}
For an entirely unobserved context, we define $(\mu,\sigma)=(0,1)$ by convention and set every observation indicator to zero. The transformed value and its observation indicator are
\begin{equation}
\widetilde{x}_i
=
\begin{cases}
\mathrm{arcsinh}\!\left((x_i-\mu)/\sigma\right), & i\in\mathcal{V}_x,\\
0, & i\notin\mathcal{V}_x,
\end{cases}
\qquad
a_i=\mathbb{I}[i\in\mathcal{V}_x].
\label{eq:missing_transform}
\end{equation}
For a prediction $\widehat{\widetilde{y}}$, the original scale is recovered by
$\widehat{y}=\sigma\sinh(\widehat{\widetilde{y}})+\mu$.

\subsubsection{Triple-Channel Patch Tokenization}
\label{sec:arch_triple}

The context is left-padded to $NP$ time steps. We assign each position a relative index $r_i$, with $r_i<0$ for context positions and $r_i\geq0$ for future positions, and use $\tau_i=r_i/C$ as its temporal feature. For context patch $n\in\{1,\ldots,N\}$, let $\mathbf{t}^{\mathrm{ctx}}_n$, $\mathbf{v}^{\mathrm{ctx}}_n$, and $\mathbf{a}^{\mathrm{ctx}}_n$ in $\mathbb{R}^{P}$ denote its temporal, transformed-value, and observation-indicator vectors. Its triple-channel representation is
\begin{equation}
\mathbf{z}^{\mathrm{ctx}}_n
=
\left[
\mathbf{t}^{\mathrm{ctx}}_n;
\mathbf{v}^{\mathrm{ctx}}_n;
\mathbf{a}^{\mathrm{ctx}}_n
\right]
\in\mathbb{R}^{3P}.
\label{eq:triple_concat}
\end{equation}
A shared residual SwiGLU projection
$\phi:\mathbb{R}^{3P}\rightarrow\mathbb{R}^{d}$
maps each patch to the latent dimension:
\begin{equation}
\mathbf{h}^{\mathrm{ctx}}_n=\phi(\mathbf{z}^{\mathrm{ctx}}_n),
\qquad
\mathbf{H}_{\mathrm{ctx}}
\in\mathbb{R}^{N\times d}.
\label{eq:patch_embed}
\end{equation}
Here, $d$ is the latent representation dimension. Since $\phi$ is a standard two-layer residual projection, we omit its internal parameterization.

\subsubsection{Parallel Future Queries}
\label{sec:arch_future}

Let $T$ denote the prediction length represented in a single encoder evaluation. This requires
$M=\lceil T/P\rceil$ future query patches, where $M\leq M_{\max}$ and $M_{\max}$ is the maximum number represented jointly. For a mini-batch $\mathcal{B}=\{1,\ldots,B\}$, let sample $b$ have lengths $(L_b,T_b)$ and patch counts
$N_b=\lceil L_b/P\rceil$ and
$M_b=\lceil T_b/P\rceil$. We write
$N_{\mathcal{B}}=\max_bN_b$,
$M_{\mathcal{B}}=\max_bM_b$, and
$S_{\mathcal{B}}=N_{\mathcal{B}}+1+M_{\mathcal{B}}$
denote the padded token dimensions. Positions outside an individual sample's context or forecast support are excluded from attention and from the empirical objective. The following equations suppress the sample index. For future patch $j\in\{0,\ldots,M-1\}$ and within-patch position $\kappa\in\{0,\ldots,P-1\}$, the temporal feature is
\begin{equation}
\mathbf{t}^{\mathrm{fut}}_j[\kappa]
=
\frac{jP+\kappa}{C}.
\label{eq:future_time}
\end{equation}
Future target values are unavailable, so the value channel is zero. The indicator channel is set to one to distinguish prospective prediction locations:
\begin{equation}
\mathbf{z}^{\mathrm{fut}}_j
=
\left[
\mathbf{t}^{\mathrm{fut}}_j;
\mathbf{0}_{P};
\mathbf{1}_{P}
\right],
\qquad
\mathbf{h}^{\mathrm{fut}}_j
=
\phi(\mathbf{z}^{\mathrm{fut}}_j).
\label{eq:future_query}
\end{equation}
Collecting these representations gives
$\mathbf{H}_{\mathrm{fut}}
=
[\mathbf{h}^{\mathrm{fut}}_0;\ldots;\mathbf{h}^{\mathrm{fut}}_{M-1}]
\in\mathbb{R}^{M\times d}$.
The indicator channel is interpreted jointly with the relative temporal feature: at context positions it records observation status, whereas at future positions it identifies prospective prediction locations. A separate patch-level indicator defines the support of self-attention. A context patch belongs to this support if it contains at least one observation; the \texttt{REG} token and future query patches are included by construction.

The context tokens, a learnable \texttt{REG} token $\mathbf{h}_{\mathrm{reg}}\in\mathbb{R}^{d}$, and the future query tokens are concatenated as
\begin{equation}
\mathbf{H}^{\mathrm{emb}}
=
\left[
\mathbf{H}_{\mathrm{ctx}};
\mathbf{h}_{\mathrm{reg}};
\mathbf{H}_{\mathrm{fut}}
\right]
\in\mathbb{R}^{S\times d},
\qquad
S=N+1+M.
\label{eq:concat_sequence}
\end{equation}

\subsubsection{Transformer Encoder}
\label{sec:arch_encoder}

Let $D$ denote the number of encoder blocks. We adopt the indexing convention $\ell\in\{0,\ldots,D-1\}$, and $\mathbf{H}^{[\ell]}\in\mathbb{R}^{S\times d}$ denotes the output of block $\ell$. Each block uses Pre-RMSNorm, RoPE~\citep{su2024roformer}, output-gated multi-head self-attention~\citep{qiu2026gated}, and a SwiGLU feed-forward layer~\citep{shazeer2020swiglu}. With $n_h$ attention heads and per-head dimension $d_h=d/n_h$, the gated attention operation is summarized as
\begin{equation}
\mathbf{O}
=
\mathrm{softmax}\!\left(
\frac{
\mathbf{Q}_{\mathrm{attn}}
\mathbf{K}_{\mathrm{attn}}^\top
}{
\sqrt{d_h}
}
+
\mathbf{m}
\right)
\mathbf{V}_{\mathrm{attn}},
\qquad
\widetilde{\mathbf{O}}
=
\mathbf{O}\odot\operatorname{sigmoid}(\mathbf{G}).
\label{eq:attention}
\end{equation}
Here, $\mathbf{Q}_{\mathrm{attn}}$, $\mathbf{K}_{\mathrm{attn}}$, $\mathbf{V}_{\mathrm{attn}}$, and $\mathbf{G}$ are the query, key, value, and output-gate projections, respectively. The additive attention mask $\mathbf{m}\in\{0,-\infty\}^{S\times S}$ assigns zero to admissible query--key pairs and $-\infty$ otherwise. The admissible set comprises observed context support, the \texttt{REG} token, and future query tokens, with bidirectional attention within this set. Because no realized future observation is included in the encoder input, the forecasting information set is preserved.

\subsection{Quantile Forecasting Head}
\label{sec:arch_quantile}

Let $\mathbf{H}^{[D-1]}_{\mathrm{fut}}\in\mathbb{R}^{M\times d}$ denote the final-layer states at future query positions, and let
$\mathbf{H}^{\mathrm{out}}_{\mathrm{fut}}=\mathrm{RMSNorm}(\mathbf{H}^{[D-1]}_{\mathrm{fut}})$
be the final normalized representation. A residual quantile head $\psi$ produces forecasts in the normalized $\mathrm{arcsinh}$ space:
\begin{equation}
\widehat{\widetilde{\mathbf{Y}}}
=
\psi\!\left(
\mathbf{H}^{\mathrm{out}}_{\mathrm{fut}}
\right)
\in
\mathbb{R}^{M\times P\times N_q},
\label{eq:quantile_head_block}
\end{equation}
where $N_q=21$ is the number of quantile levels and
\begin{equation}
\mathcal{Q}
=
\{
0.01,0.05,0.10,0.15,0.20,0.25,0.30,0.35,0.40,0.45,
0.50,
0.55,0.60,0.65,0.70,0.75,0.80,0.85,0.90,0.95,0.99
\}.
\label{eq:quantile_set}
\end{equation}
The patch and within-patch dimensions are flattened in chronological order to obtain
$\widehat{\widetilde{\mathbf{Y}}}_{\mathrm{flat}}\in\mathbb{R}^{MP\times N_q}$.
When the requested horizon $T$ is not divisible by $P$, only the first $T$ positions are retained; the remaining positions of the final patch do not participate in training or evaluation. The head therefore produces a complete conditional quantile function at every represented forecast position, while the median channel $q=0.5$ provides the point forecast used by the multi-stage inference procedure in Section~\ref{sec:arch_inference}.

\begin{figure}[t]
    \centering
    \includegraphics[width=\textwidth]{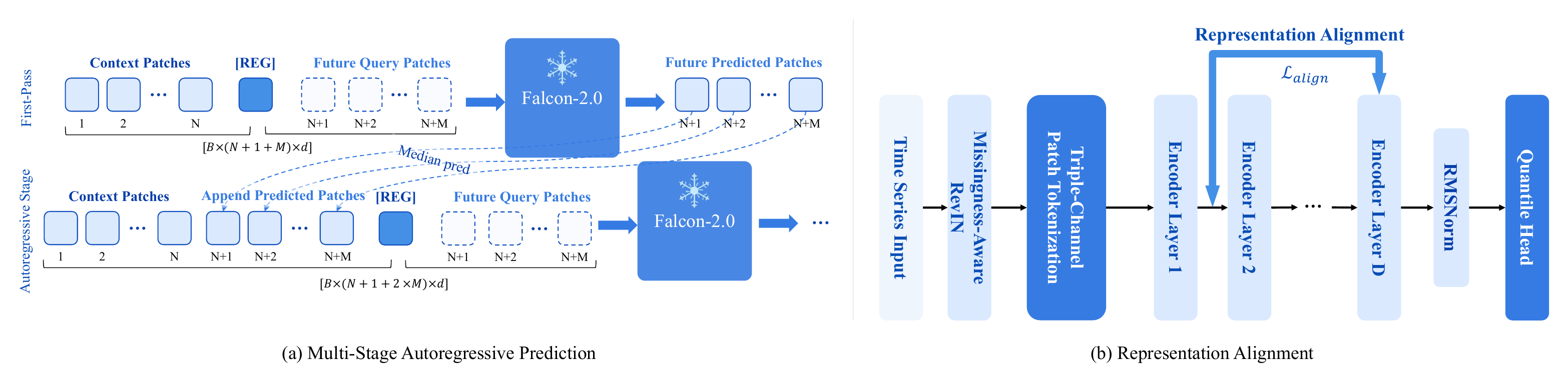}
    \caption{\NAME{} long-horizon inference and auxiliary representation alignment. (a) Multi-Stage Autoregressive Prediction: each stage predicts up to $M_{\max}$ future patches in parallel; its median forecast is appended to the context before the next stage, and stage-wise outputs are concatenated and truncated to the requested horizon. (b) Rank-Guided Cross-Depth Alignment: a deep encoder block provides a stop-gradient reference representation for a shallow block through a token-wise cosine objective. The auxiliary objective is used only during training.}
\label{fig:infer_and_align}
\end{figure}

\subsection{Rank-Guided Cross-Depth Alignment}
\label{sec:arch_rank_alignment}

\paratitle{Motivation.}
Recent analysis shows that time series patch embeddings have sharply decaying singular spectra and that their numerical rank tends to increase across Transformer depth, a phenomenon termed the \emph{flow of ranks}~\citep{yu2026understanding}. That work uses the low-rank structure of early layers primarily for compression. We investigate a complementary training hypothesis: when a late-layer representation exhibits a broader non-negligible spectrum, can its geometry guide a shallow layer without adding an external teacher or changing inference?

\paratitle{Asymmetric cross-depth objective.}
For a batch of size $B$, let
$\mathbf{h}^{[\ell]}_{b,u}\in\mathbb{R}^{d}$
denote token $u$ of sample $b$ after encoder block $\ell$. We align a shallow block $\ell_{\mathrm{sh}}$ with a deeper block $\ell_{\mathrm{dp}}$, where
$0\leq\ell_{\mathrm{sh}}<\ell_{\mathrm{dp}}\leq D-1$.
Under this indexing convention, the default 32-layer model uses
$(\ell_{\mathrm{sh}},\ell_{\mathrm{dp}})=(1,31)$.

Inspired by the asymmetric optimization used by TimeAlign~\citep{hu2025bridging}, gradients are stopped through the deep representation. Unlike TimeAlign, however, both representations here come from different depths of the same encoder rather than from a separate branch that observes future targets. Define
\begin{equation}
\mathbf{z}^{\mathrm{sh}}_{b,u}
=
\frac{
\mathbf{h}^{[\ell_{\mathrm{sh}}]}_{b,u}
}{
\left\|
\mathbf{h}^{[\ell_{\mathrm{sh}}]}_{b,u}
\right\|_2
},
\qquad
\mathbf{z}^{\mathrm{dp}}_{b,u}
=
\operatorname{sg}\!\left(
\frac{
\mathbf{h}^{[\ell_{\mathrm{dp}}]}_{b,u}
}{
\left\|
\mathbf{h}^{[\ell_{\mathrm{dp}}]}_{b,u}
\right\|_2
}
\right),
\label{eq:rank_align_normalize}
\end{equation}
where $\operatorname{sg}(\cdot)$ denotes stop-gradient. For numerical stability, each denominator is lower-bounded by a small positive constant $\epsilon_{\mathrm{align}}$.

Let $\omega_{b,u}\in\{0,1\}$ denote the alignment-support indicator. It equals one for valid context patches and future query patches within the sampled horizon, and zero for padding-induced positions and the \texttt{REG} token. The token-wise cosine objective is
\begin{equation}
\mathcal{L}_{\mathrm{align}}
=
\frac{
1
}{
n_v
}
\sum_{b=1}^{B}
\sum_{u=1}^{S_{\mathcal{B}}}
\omega_{b,u}
\left(
1-
\left\langle
\mathbf{z}^{\mathrm{sh}}_{b,u},
\mathbf{z}^{\mathrm{dp}}_{b,u}
\right\rangle
\right),
\qquad
n_v
=
\sum_{b,u}\omega_{b,u}.
\label{eq:rank_align_loss}
\end{equation}
The objective updates the shallow representation using the current deep representation as an asymmetric reference signal. It introduces no additional forecasting function and is not evaluated during inference.

\paratitle{Spectral diagnostics.}
Stack the $n_v$ valid, row-normalized shallow and deep token vectors into
$\mathbf{Z}_{\mathrm{sh}},\mathbf{Z}_{\mathrm{dp}}\in\mathbb{R}^{n_v\times d}$.
To remove a shared mean direction before measuring spectral breadth, define the centering matrix
$\mathbf{C}_v=\mathbf{I}_{n_v}-n_v^{-1}\mathbf{1}\mathbf{1}^{\top}$ and
\begin{equation}
\overline{\mathbf{Z}}_{\mathrm{sh}}
=
\mathbf{C}_v\mathbf{Z}_{\mathrm{sh}},
\qquad
\overline{\mathbf{Z}}_{\mathrm{dp}}
=
\mathbf{C}_v\mathbf{Z}_{\mathrm{dp}}.
\label{eq:rank_align_center}
\end{equation}
For a centered representation matrix $\overline{\mathbf{Z}}$ with singular values
$s_1(\overline{\mathbf{Z}})\geq s_2(\overline{\mathbf{Z}})\geq\cdots$, we consider the $\varepsilon$-numerical rank and stable rank
\begin{equation}
r_{\varepsilon}(\overline{\mathbf{Z}})
=
\left|
\left\{
j:
s_j(\overline{\mathbf{Z}})
>
\varepsilon s_1(\overline{\mathbf{Z}})
\right\}
\right|,
\qquad
r_{\mathrm{stable}}(\overline{\mathbf{Z}})
=
\frac{
\|\overline{\mathbf{Z}}\|_F^2
}{
\|\overline{\mathbf{Z}}\|_2^2
}.
\label{eq:rank_metrics}
\end{equation}
The first counts singular directions that remain non-negligible relative to the dominant mode, whereas the second measures how broadly representation energy is distributed. Both are diagnostics rather than optimization targets: a broader spectrum does not by itself imply more predictive information.

\paratitle{From cosine alignment to spectral transfer.}
Because every aligned row has unit norm, the token-wise cosine objective is exactly equivalent to a matrix discrepancy:
\begin{equation}
\left\|
\mathbf{Z}_{\mathrm{sh}}
-
\mathbf{Z}_{\mathrm{dp}}
\right\|_F^2
=
2n_v\mathcal{L}_{\mathrm{align}}.
\label{eq:rank_align_cos_fro}
\end{equation}
Since $\|\mathbf{C}_v\|_2\leq1$, centering cannot amplify this discrepancy. Hence
\begin{equation}
\delta_F
:=
\left\|
\overline{\mathbf{Z}}_{\mathrm{sh}}
-
\overline{\mathbf{Z}}_{\mathrm{dp}}
\right\|_F
\leq
\sqrt{2n_v\mathcal{L}_{\mathrm{align}}},
\qquad
\eta
:=
\left\|
\overline{\mathbf{Z}}_{\mathrm{sh}}
-
\overline{\mathbf{Z}}_{\mathrm{dp}}
\right\|_2
\leq
\delta_F.
\label{eq:rank_align_centered_bound}
\end{equation}
Standard singular-value perturbation bounds~\citep{stewart1990matrix} then give, for every $j$,
\begin{equation}
\left|
s_j(\overline{\mathbf{Z}}_{\mathrm{sh}})
-
s_j(\overline{\mathbf{Z}}_{\mathrm{dp}})
\right|
\leq
\eta.
\label{eq:rank_align_weyl}
\end{equation}
Consequently, if the $r$-th deep singular mode is separated from the numerical-rank threshold by more than the alignment perturbation, namely
\begin{equation}
s_r(\overline{\mathbf{Z}}_{\mathrm{dp}})
-
\eta
>
\varepsilon
\left(
s_1(\overline{\mathbf{Z}}_{\mathrm{dp}})
+
\eta
\right),
\label{eq:rank_align_tail_condition}
\end{equation}
then
$r_{\varepsilon}(\overline{\mathbf{Z}}_{\mathrm{sh}})\geq r$.
This establishes a conditional transfer result: cosine alignment does not create rank unconditionally, but sufficiently accurate alignment prevents the shallow representation from retaining fewer than $r$ non-negligible modes when the deep spectrum satisfies the stated separation condition.

The same perturbation also controls stable rank. By the triangle and reverse-triangle inequalities, whenever
$\delta_F<\|\overline{\mathbf{Z}}_{\mathrm{dp}}\|_F$,
\begin{equation}
r_{\mathrm{stable}}(\overline{\mathbf{Z}}_{\mathrm{sh}})
\geq
\frac{
\left(
\|\overline{\mathbf{Z}}_{\mathrm{dp}}\|_F-\delta_F
\right)^2
}{
\left(
\|\overline{\mathbf{Z}}_{\mathrm{dp}}\|_2+\eta
\right)^2
}.
\label{eq:rank_align_stable_bound}
\end{equation}
Thus, a small alignment loss also limits how far the shallow representation's energy distribution can deviate from that of the deep representation, although the bound may be loose when the deep spectrum is highly concentrated.

\paratitle{Subspace interpretation and directionality.}
Let $\mathbf{V}^{\mathrm{sh}}_r$ and $\mathbf{V}^{\mathrm{dp}}_r$ contain the top-$r$ right singular vectors of the centered shallow and deep representations, and define the corresponding projectors
$\mathbf{P}^{\mathrm{sh}}_r=\mathbf{V}^{\mathrm{sh}}_r(\mathbf{V}^{\mathrm{sh}}_r)^\top$
and
$\mathbf{P}^{\mathrm{dp}}_r=\mathbf{V}^{\mathrm{dp}}_r(\mathbf{V}^{\mathrm{dp}}_r)^\top$.
Their normalized chordal discrepancy is
\begin{equation}
d_{\mathrm{sub}}^2(r)
=
\frac{1}{2r}
\left\|
\mathbf{P}^{\mathrm{sh}}_r
-
\mathbf{P}^{\mathrm{dp}}_r
\right\|_F^2
=
\frac{1}{r}
\sum_{j=1}^{r}
\sin^2\theta_j,
\label{eq:rank_align_subspace}
\end{equation}
where $\theta_j$ are the principal angles between the two feature subspaces. When the deep representation has a nonzero spectral gap around its $r$-th singular value, matrix perturbation theory provides an upper bound on this discrepancy that vanishes with $\eta$. We use Equation~\eqref{eq:rank_align_subspace} only as an analysis diagnostic; the training objective remains the less expensive token-wise cosine loss in Equation~\eqref{eq:rank_align_loss}.

Finally, stop-gradient determines the direction of transfer. Without it, the auxiliary loss could decrease by moving the deep representation toward the narrower shallow representation. Treating the deep state as fixed within each update instead directs the alignment gradient to the shallow branch and preceding blocks. The forecasting objective continues to update the full network, so alignment regularizes representation development without freezing the deep encoder during training.

\subsection{Training Loss}
\label{sec:arch_loss}

Let sample $b$ have a sampled forecast horizon $T_b$ and targets
$\mathbf{y}_b=(y_{b,1},\ldots,y_{b,T_b})$.
Its observed target set is
\begin{equation}
\mathcal{V}_b
=
\left\{
t\in\{1,\ldots,T_b\}:
y_{b,t}\ \text{is observed}
\right\}.
\label{eq:target_support}
\end{equation}
The context statistics $(\mu_b,\sigma_b)$ from Equation~\eqref{eq:missing_revin} are reused to transform every observed target:
\begin{equation}
\widetilde{y}_{b,t}
=
\mathrm{arcsinh}\!\left(
\frac{y_{b,t}-\mu_b}{\sigma_b}
\right),
\qquad
t\in\mathcal{V}_b.
\label{eq:target_transform}
\end{equation}
Let $\widehat{\widetilde{y}}_{b,t}^{(q)}$ denote the corresponding quantile-head output for $q\in\mathcal{Q}$ after flattening the patch dimension. With
$N_{\mathrm{tgt}}=\sum_{b=1}^{B}|\mathcal{V}_b|$, the forecasting loss is the observed-target pinball objective~\citep{koenker2001quantile}
\begin{equation}
\mathcal{L}_{\mathrm{pin}}
=
\frac{1}{
N_qN_{\mathrm{tgt}}
}
\sum_{b=1}^{B}
\sum_{t\in\mathcal{V}_b}
\sum_{q\in\mathcal{Q}}
\rho_q\!\left(
\widetilde{y}_{b,t}
-
\widehat{\widetilde{y}}_{b,t}^{(q)}
\right),
\qquad
\rho_q(e)
=
\max\!\left(
qe,(q-1)e
\right).
\label{eq:pinball}
\end{equation}
The residual convention is $e=\widetilde{y}-\widehat{\widetilde{y}}$, under which minimizing $\rho_q$ estimates the conditional $q$-quantile. Targets that are unobserved, introduced solely by batch padding, or located beyond the sampled horizon have zero contribution. Samples with $\mathcal{V}_b=\varnothing$ are excluded before the batch objective is formed, ensuring that $N_{\mathrm{tgt}}>0$.

Training in the normalized $\mathrm{arcsinh}$ space prevents high-amplitude series from dominating the batch objective while retaining an invertible map to the original scale. The median prediction, $q=0.5$, minimizes the absolute-error component and is used as the recursive input during long-horizon inference; the remaining quantiles characterize predictive uncertainty and are never collapsed into the median during training.

The complete pre-training objective combines forecasting accuracy with cross-depth representation alignment:
\begin{equation}
\mathcal{L}_{\mathrm{total}}
=
\mathcal{L}_{\mathrm{pin}}
+
\lambda_{\mathrm{align}}
\mathcal{L}_{\mathrm{align}},
\label{eq:rank_align_total}
\end{equation}
where the default run uses $\lambda_{\mathrm{align}}=10.0$. The forecasting term updates the patch-embedding map, all encoder blocks, and the quantile head. Because the deep representation is stop-gradient in Equation~\eqref{eq:rank_align_normalize}, the auxiliary term updates only the shallow branch and its preceding computation. At evaluation time, $\mathcal{L}_{\mathrm{align}}$ is absent and the forecasting architecture is unchanged.

\subsection{Multi-Stage Autoregressive Prediction}
\label{sec:arch_inference}

\NAME{} represents at most $M_{\max}$ future patches in one encoder evaluation, giving a per-stage forecasting capacity
\begin{equation}
T_{\max}
=
M_{\max}P.
\label{eq:stage_capacity}
\end{equation}
For a requested horizon $T_{\mathrm{req}}$, the number of stages is
\begin{equation}
K
=
\left\lceil
\frac{T_{\mathrm{req}}}{T_{\max}}
\right\rceil.
\label{eq:num_inference_stages}
\end{equation}
At stage $k\in\{1,\ldots,K\}$, let
\begin{equation}
T_k
=
\min\!\left(
T_{\max},
T_{\mathrm{req}}-(k-1)T_{\max}
\right),
\qquad
M_k
=
\left\lceil
\frac{T_k}{P}
\right\rceil.
\label{eq:stage_horizon}
\end{equation}
The model constructs $M_k$ future query patches and predicts all $M_kP$ positions in parallel. Only the first $T_k$ positions are retained when the final stage ends inside a patch.

Normalization is performed once. Specifically, $(\mu,\sigma)$ are computed from the original observed context, and the resulting transformed context is denoted by $\widetilde{\mathbf{x}}^{(0)}$. The same statistics are held fixed at every stage so that observed context values and recursively generated values remain in a common coordinate system. Given the current context $\widetilde{\mathbf{x}}^{(k-1)}$, stage $k$ produces
$\widehat{\widetilde{\mathbf{y}}}^{(q,k)}\in\mathbb{R}^{T_k}$
for every $q\in\mathcal{Q}$. Its median is incorporated into the next context as
\begin{equation}
\widetilde{\mathbf{x}}^{(k)}
=
\operatorname{Tail}_{C}\!\left(
\left[
\widetilde{\mathbf{x}}^{(k-1)};
\widehat{\widetilde{\mathbf{y}}}^{(0.5,k)}
\right]
\right),
\label{eq:stage_context_update}
\end{equation}
where $\operatorname{Tail}_{C}$ retains the most recent $C$ time steps when the accumulated context exceeds the admissible context length. Recursively generated positions receive observation indicator one because they are available as conditioning values in the subsequent stage.

For each quantile level, the final transformed forecast concatenates the stage outputs in temporal order:
\begin{equation}
\widehat{\widetilde{\mathbf{y}}}^{(q)}
=
\operatorname{Trunc}_{T_{\mathrm{req}}}\!\left(
\left[
\widehat{\widetilde{\mathbf{y}}}^{(q,1)};
\cdots;
\widehat{\widetilde{\mathbf{y}}}^{(q,K)}
\right]
\right).
\label{eq:stage_output_concat}
\end{equation}
The original scale is recovered only after all stages:
\begin{equation}
\widehat{y}^{(q)}_t
=
\sigma
\sinh\!\left(
\widehat{\widetilde{y}}^{(q)}_t
\right)
+
\mu,
\qquad
t=1,\ldots,T_{\mathrm{req}}.
\label{eq:stage_inverse_transform}
\end{equation}
Therefore, prediction is parallel within each stage and autoregressive only across stages. When $T_{\mathrm{req}}\leq T_{\max}$, $K=1$ and the procedure reduces to direct parallel forecasting without recursive feedback. For longer horizons, only the median trajectory is fed back; all quantile trajectories are nevertheless retained as outputs.

\subsection{Model Configuration}
\label{sec:arch_config}

The default \NAME{} configuration contains $D=32$ encoder blocks and 585M trainable parameters. Its latent dimension is $d=1024$, partitioned across $n_h=16$ attention heads with $d_h=64$, satisfying $d=n_hd_h$. Each block expands the representation to $d_{\mathrm{ff}}=4096$ in its SwiGLU feed-forward sublayer. The stack uses Pre-RMSNorm, RoPE with base $10{,}000$, output-gated self-attention, and bias-free linear maps.

The common patch size is $P=16$ for both context and future segments. A maximum context of $C=8192$ time steps therefore contains
$N_{\max}=C/P=512$ context patches. Together with one \texttt{REG} token and $M_{\max}=6$ future query patches, the largest encoder sequence contains
$S_{\max}=N_{\max}+1+M_{\max}=519$ tokens. The corresponding per-stage forecasting capacity is
$T_{\max}=M_{\max}P=96$ time steps; longer horizons use the procedure in Section~\ref{sec:arch_inference}. These values specify the model's representational capacity, whereas the per-example context and horizon are sampled by \Orbit{} as described in Section~\ref{sec:orbit}.

Table~\ref{tab:model_config} summarizes the configuration. We report it for reproducibility and do not treat these conventional architectural choices as the principal contribution.

\begin{table}[t]
\centering
\caption{Model configuration of \NAME{}. Encoder blocks follow the indexing convention $\{0,\ldots,D-1\}$.}
\label{tab:model_config}
\tablestyle{6pt}{1.15}
\begin{tabular}{ll}
\toprule
\textbf{Parameter} & \textbf{Value} \\
\midrule
Encoder blocks ($D$) & 32 \\
Trainable parameters & 585M \\
Latent representation dimension ($d$) & 1024 \\
FFN hidden size ($d_{\mathrm{ff}}$) & 4096 \\
Attention heads ($n_h$) & 16 \\
Per-head dimension ($d_h$) & 64 \\
Patch size ($P$) & 16 \\
Maximum future patches ($M_{\max}$) & 6 \\
Maximum per-stage horizon ($T_{\max}$) & 96 \\
Maximum context length ($C$) & 8192 \\
Maximum context patches ($N_{\max}$) & 512 \\
Maximum encoder tokens ($S_{\max}$) & 519 \\
Positional encoding & RoPE (base $10{,}000$) \\
Normalization & RMSNorm ($\epsilon_{\mathrm{rms}}=10^{-5}$) \\
Attention & Output-gated self-attention \\
FFN activation & SwiGLU \\
Bias terms in linear maps & Absent \\
Quantile levels ($N_q$) & 21 \\
Alignment blocks $(\ell_{\mathrm{sh}},\ell_{\mathrm{dp}})$ & $(1,31)$ \\
\bottomrule
\end{tabular}
\end{table}

\section{\Orbit{}}
\label{sec:orbit}

\begin{figure}[t]
    \centering
    \includegraphics[width=\textwidth]{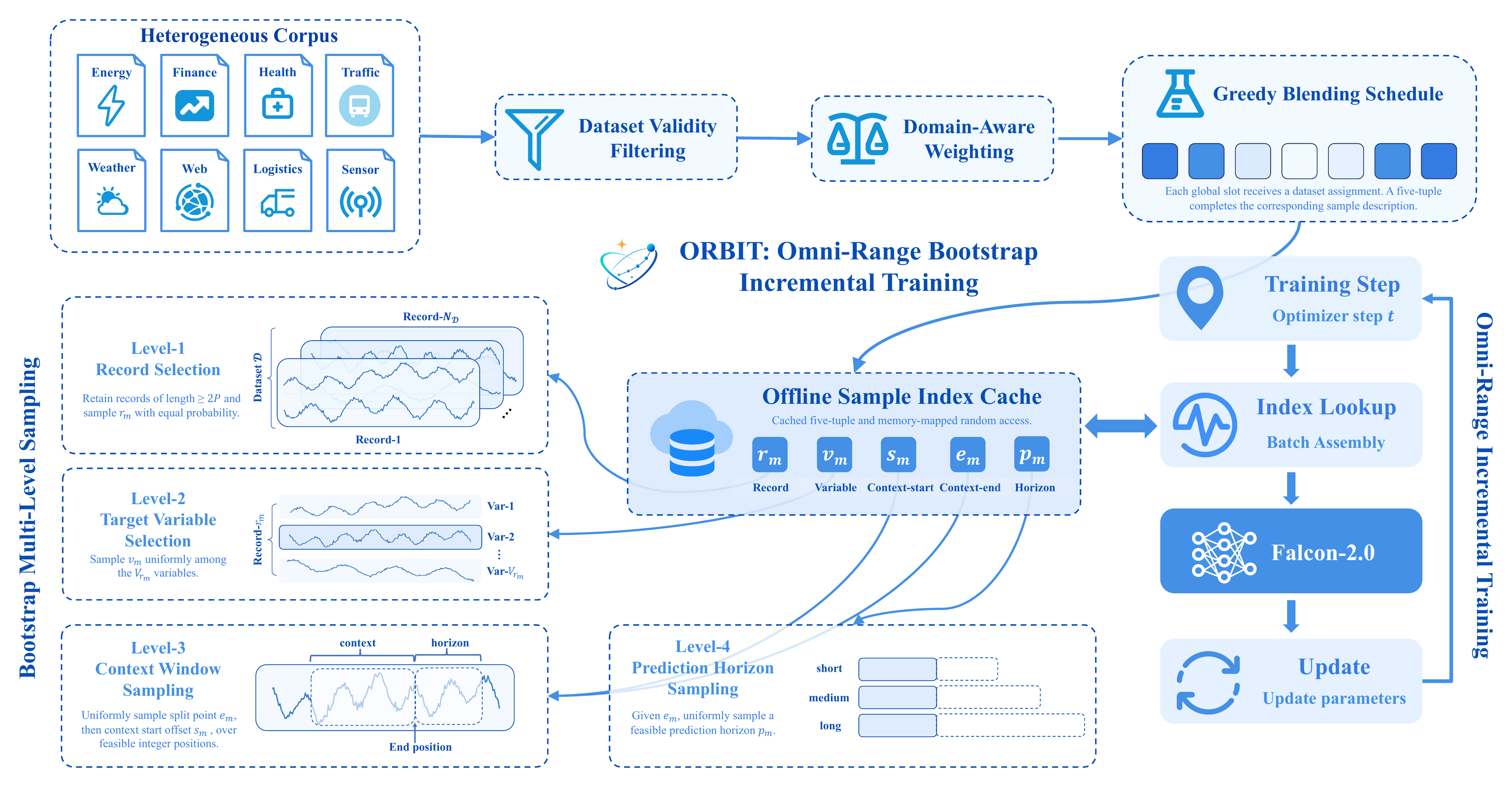}
    \caption{Overview of \Orbit{}. At the corpus level, dataset validity filtering and prescribed weights are translated by greedy blending into dataset assignments for the global training stream. Within each retained dataset, record filtering and four-level stochastic sampling construct a sample index of five-tuples over records, target variables, context windows, and prediction horizons. The resulting offline cache supports index lookup and batch assembly for \textbf{Omni-Range Incremental Training}.}
\label{fig:orbit}
\end{figure}

Training a time series foundation model on a heterogeneous corpus implicitly defines an \emph{effective pre-training distribution}: the probability with which each dataset, record, target variable, context window, and prediction horizon contributes to the optimization objective. This distribution is particularly consequential for time series because data sources differ jointly in domain, sampling frequency, record length, variable count, and missingness~\citep{aksu2024gifteval,shchur2025fev}. When training examples are obtained through sequential traversal or deterministic window enumeration, their exposure is determined largely by corpus layout and the number of enumerable windows, rather than by an explicitly specified training objective.

As illustrated in Figure~\ref{fig:orbit}, \Orbit{} separates the construction and consumption of this distribution into two complementary components. \textbf{Bootstrap Multi-Level Sampling} controls source exposure and constructs forecasting examples by sampling records, target variables, temporal split points, and prediction horizons. \textbf{Omni-Range Incremental Training} interleaves the resulting context and horizon ranges throughout a single step-based training run, rather than assigning different ranges to separate training stages. Operationally, the split-point and horizon variables are sampled once during index construction and are subsequently consumed by Omni-Range training; they are not independently resampled by the two components. The channel-independent backbone, triple-channel tokenization, and parallel patch prediction of \NAME{} provide an interface for processing this variable-length and missingness-aware stream, but do not themselves define the pre-training distribution.

\subsection{Problem Formulation and Motivation}
\label{sec:bootstrap_motivation}

Given a prescribed weighting over the pre-training datasets, consider a dataset $\mathcal{D}$ containing $N_{\mathcal{D}}$ time series records. Record $i$ is represented as $\mathbf{X}_i \in \mathbb{R}^{L_i \times V_i}$, where $L_i$ denotes its temporal length and $V_i \geq 1$ its number of target variables; the $v$-th target variable is denoted by $\mathbf{x}_{i,v} \in \mathbb{R}^{L_i}$. The objective is to make the aggregate dataset composition of the global training stream follow these weights while sampling forecasting instances over time series records, target variables, temporal windows, and prediction horizons. At the same time, the data pipeline must support efficient random access so that data loading does not become a bottleneck in large-scale distributed training.

A common baseline is \emph{fixed sliding-window sampling}, which sequentially enumerates training examples from each record~\citep{goswami2024moment,shi2024timemoe,liu2025timerxl}. Although straightforward, fixed window enumeration makes training exposure depend on the number of eligible windows. Adjacent windows often overlap substantially, producing highly redundant training examples. Datasets that yield more windows receive greater exposure. Reusing a fixed set of window boundaries also limits the diversity of context--target configurations. Together, these effects increase the nominal sample count without a corresponding gain in training diversity, potentially skewing data exposure during optimization.

Such distributional biases motivate a sampling strategy that randomizes across multiple dimensions instead of relying on deterministic window traversal. To specify the forecasting examples produced by such a strategy, we introduce the following representation.

\begin{definition}[Sample Index]
\label{def:sample_index}
For the $N_{\mathcal{D}}$ records in dataset $\mathcal{D}$, a \emph{sample index} $\mathcal{I}_{\mathcal{D}}$ with $M_{\mathcal{D}}$ entries maps each sample identifier $m \in \{1,\ldots,M_{\mathcal{D}}\}$ to a valid extraction tuple
\[
\mathcal{I}_{\mathcal{D}}(m) = (r_m, v_m, s_m, e_m, p_m),
\]
where $r_m \in \{1,\ldots,N_{\mathcal{D}}\}$ denotes the index of a time series record with temporal length $L_{r_m}$ and $V_{r_m}$ target variables, and $v_m \in \{1,\ldots,V_{r_m}\}$ denotes the index of a target variable within that record. $s_m,e_m \in \mathbb{Z}$ denote the starting offset and exclusive endpoint, respectively, of the context window within record $r_m$, with $0 \leq s_m < e_m < L_{r_m}$ and $e_m-s_m \geq P$, while $p_m \in \mathbb{Z}_{\geq P}$ denotes an admissible prediction horizon satisfying $p_m \leq T_{\max}$ and $e_m + p_m \leq L_{r_m}$. Accordingly, the context is the segment of $\mathbf{x}_{r_m,v_m}$ spanning offsets $s_m$ through $e_m-1$, and the target is the immediately following segment spanning offsets $e_m$ through $e_m+p_m-1$.
\end{definition}

Beyond the validity of individual extraction tuples, the sampling scheme should satisfy requirements at two levels. At the corpus level, aggregate dataset exposure over the complete global training stream should follow the prescribed weights. Within each dataset, the construction of the sample index should ensure that every valid record remains eligible for selection and that time series records, target variables, temporal window positions, and prediction horizons are selected stochastically.

\subsection{Bootstrap Multi-Level Sampling}
\label{sec:bootstrap}

Bootstrap Multi-Level Sampling addresses these requirements hierarchically. At the corpus level, dataset weighting and blending control how frequently each dataset contributes to the global training stream. Within each dataset, a four-level stochastic procedure constructs the corresponding sample index by selecting the time series record, target variable, context window, and prediction horizon of each indexed example.

\subsubsection{Dataset Weighting and Blending}
\label{sec:bootstrap_weighting}

We first apply dataset-level validity filtering to exclude datasets with excessive overall missingness or no record long enough to form a valid context--target pair. Across the retained datasets, \Orbit{} uses \emph{domain-aware weighting} to balance exposure across domains and prevent high-volume datasets from dominating the training distribution. These weights define the desired dataset composition of an ordered global training stream. The stream length is fixed in advance to match the total sample budget of a single pre-training run.

With the desired composition and stream length specified, we use a low-discrepancy greedy blending rule to materialize the dataset assignments. For each slot, the rule selects the dataset whose cumulative assigned count has the largest deficit relative to its target count at that point. Unlike independent categorical sampling, which matches the prescribed proportions only in expectation, this rule keeps the cumulative dataset composition close to its target after every assignment. Once a dataset has been assigned to each slot, a local sample identifier is selected for that dataset. As described next, the corresponding per-dataset sample index maps this identifier to the five-tuple \((r_m,v_m,s_m,e_m,p_m)\). Together with the dataset assignment, this five-tuple completes the indexed extraction description associated with the corresponding slot.

\subsubsection{Bootstrap Stochastic Sampling}
\label{sec:bootstrap_algorithm}

Complementing the corpus-level dataset assignments, Bootstrap Stochastic Sampling operates separately within each retained dataset \(\mathcal{D}\) to construct its sample index \(\mathcal{I}_{\mathcal{D}}\). For every sample identifier \(m\in\{1,\ldots,M_{\mathcal{D}}\}\), a four-level stochastic procedure generates the extraction tuple \(\mathcal{I}_{\mathcal{D}}(m)=(r_m,v_m,s_m,e_m,p_m)\) by sampling a valid record, a target variable within that record, a context window, and a compatible prediction horizon. This construction avoids deterministic record traversal and fixed window boundaries, helping reduce sequential correlations and increase the diversity of context--target configurations.

\paratitle{Level-1: Record Selection.} We first exclude records shorter than $2P$ time steps, where $P$ is the common patch size introduced in Section~\ref{sec:arch_backbone}, so that every retained record can provide at least one full patch for both the context and target segments. For each sample identifier $m$, the record index $r_m$ is sampled with equal probability from the retained records, without weighting by record length or variable count.

\paratitle{Level-2: Target Variable Selection.} Conditioned on the selected record $r_m$, the procedure draws a target-variable index $v_m$ with equal probability from $\{1,\ldots,V_{r_m}\}$. Together with the equal-probability record sampling in Level-1, this prevents target-variable exposure from scaling with the number of extractable temporal windows.

\paratitle{Level-3: Context Window Sampling.} Context requirements vary substantially across records and sampling frequencies, motivating coverage of both short local histories and longer temporal extents. After selecting record $r_m$ and target variable $v_m$ in Levels 1 and 2, respectively, the procedure operates on the resulting univariate series $\mathbf{x}_{r_m,v_m}$ of length $L_{r_m}$. It first samples the exclusive endpoint of the context window, which also serves as the context--target split point, uniformly from the feasible integer positions:
\begin{equation}
e_m \mid r_m \sim \operatorname{Unif}\!\left\{P,\ldots,L_{r_m}-P\right\}.
\end{equation}
Here $P$ is the common patch size introduced in Section~\ref{sec:arch_backbone}; the two bounds reserve at least one full patch on each side of the split point. Conditioned on $e_m$, the context starting offset is then sampled uniformly from its feasible integer range:
\begin{equation}
s_m \mid e_m \sim \operatorname{Unif}\!\left\{\max(0,e_m-C),\ldots,e_m-P\right\},
\end{equation}
where $C$ is the maximum admissible context length defined in Section~\ref{sec:arch_backbone}. Consequently, the sampled context length $e_m-s_m$ ranges from $P$ to $\min(C,e_m)$, exposing the model to different temporal extents without exceeding its context capacity.

\paratitle{Level-4: Prediction Horizon Sampling.} Forecasting applications likewise require prediction horizons ranging from short to long, making a single fixed horizon unnecessarily restrictive. Given the sampled split point $e_m$, the prediction horizon length is sampled uniformly from the feasible integer set
\begin{equation}
p_m \mid e_m,r_m \sim \operatorname{Unif}\!\left\{P,\ldots,\min\!\left(T_{\max},L_{r_m}-e_m\right)\right\}.
\end{equation}
Here $T_{\max}$ is the maximum per-stage forecasting capacity defined in Section~\ref{sec:arch_inference}. The lower bound supplies at least one full target patch, while the upper bound respects both the model's per-stage capacity and the number of future observations remaining after $e_m$. Equivalently, the exclusive endpoint of the target lies between $e_m+P$ and $\min(e_m+T_{\max},L_{r_m})$. The sampling is therefore uniform over the feasible horizon range determined jointly by the selected record and split point. Earlier split points can admit horizons up to $T_{\max}$, whereas later positions are limited by the shorter remaining suffix. Across the entries of the sample index, this construction allows short- to long-range targets to coexist whenever the selected records permit, rather than binding the sample index to a single prediction length.

Repeating the four-level procedure for all $M_{\mathcal{D}}$ sample identifiers yields the sample index $\mathcal{I}_{\mathcal{D}}\in\mathbb{Z}^{M_{\mathcal{D}}\times 5}$. Constructed offline and cached for reproducibility and reuse across training runs, the sample index separates the specification of context--horizon configurations from their subsequent consumption during training. Algorithm~\ref{alg:bootstrap_uni} summarizes the complete construction.

\begin{algorithm}[H]
\caption{Bootstrap Sample Index Construction}
\label{alg:bootstrap_uni}
\begin{algorithmic}[1]
\Require Dataset $\mathcal{D}$ with $N_{\mathcal{D}}$ records, record lengths $\{L_i\}$, and variable counts $\{V_i\}$; patch size $P$; maximum context length $C \geq P$; maximum per-stage horizon $T_{\max} \geq P$; sample-index size $M_{\mathcal{D}}$
\Ensure Sample index $\mathcal{I}_{\mathcal{D}} \in \mathbb{Z}^{M_{\mathcal{D}} \times 5}$
\State Initialize a random generator $G$
\State $\mathcal{R}_{\mathcal{D}} \leftarrow \{i \in \{1,\ldots,N_{\mathcal{D}}\}: L_i \geq 2P\}$
\For{$m = 1$ to $M_{\mathcal{D}}$}
    \State \Comment{Level-1: Record Selection}
    \State $r_m \leftarrow G.\mathrm{choice}(\mathcal{R}_{\mathcal{D}})$
    \State \Comment{Level-2: Target Variable Selection}
    \State $v_m \leftarrow G.\mathrm{choice}(\{1,\ldots,V_{r_m}\})$
    \State \Comment{Level-3: Context Window Sampling}
    \State $e_m \leftarrow G.\mathrm{choice}(\{P,\ldots,L_{r_m}-P\})$
    \State $s_m \leftarrow G.\mathrm{choice}(\{\max(0,e_m-C),\ldots,e_m-P\})$
    \State \Comment{Level-4: Prediction Horizon Sampling}
    \State $p_m \leftarrow G.\mathrm{choice}(\{P,\ldots,\min(T_{\max},L_{r_m}-e_m)\})$
    \State $\mathcal{I}_{\mathcal{D}}(m) \leftarrow (r_m,\, v_m,\, s_m,\, e_m,\, p_m)$
\EndFor
\State \Return $\mathcal{I}_{\mathcal{D}}$
\end{algorithmic}
\end{algorithm}

\subsection{Omni-Range Incremental Training}
\label{sec:omnirange}

A time series foundation model must generalize across temporal resolutions and forecasting requirements that call for widely different context lengths and prediction horizons. Training with a fixed context--horizon configuration covers only a narrow operating regime, whereas multi-stage context extension or horizon-specific optimization treats different temporal ranges through separate schedules or objectives and increases training complexity~\citep{ansari2025chronos2,liu2026timers1,shi2024timemoe}. Building on the sample indices defined in Section~\ref{sec:bootstrap_algorithm}, Omni-Range training stochastically mixes diverse context--horizon configurations in a single training run, enabling the model to learn across a broad range of temporal scales. Operationally, Omni-Range Incremental Training comprises two complementary components: assembling samples with different context lengths and prediction horizons into mini-batches and incrementally consuming cached sample index entries during optimization.

\paratitle{Omni-Range Batch Assembly.} The five-tuples selected for a mini-batch generally encode different context lengths $e_m-s_m$ and prediction horizons $p_m$ and therefore cannot be stacked directly. During batch assembly, context windows are left-padded to the longest context in the mini-batch, aligning their valid endpoints at the context--target split, while target windows are right-padded to the longest prediction horizon, aligning the beginnings of their forecast ranges. Padded context positions are marked invalid in the observation-indicator channel, with fully padded context patches excluded by the attention mask; padded target positions are excluded by the loss mask and therefore make no direct contribution to the training objective. The model can therefore train jointly on examples spanning different context and prediction ranges within a single mini-batch, without any additional runtime cropping or resampling.

\paratitle{Incremental Sample Consumption.} Before optimization, the global training stream constructed in Section~\ref{sec:bootstrap_weighting} is globally shuffled, while the sample identifiers within each dataset are shuffled independently. Together, these shuffling steps establish the sequence in which sample index entries are accessed during the training run. Each entry is a five-tuple of extraction metadata, $(r_m,v_m,s_m,e_m,p_m)$, rather than a time series sample that can be consumed directly. At training time, the corresponding context and target segments are loaded on demand from memory-mapped storage as specified by the five-tuple. Loading only the required segments avoids materializing all sampled windows in memory, reducing the memory footprint while supporting efficient batch construction. Overall, the globally interleaved training stream supports incremental consumption of stochastically constructed samples, promoting sample diversity and reducing redundant exposure compared with conventional epoch-based training, which repeatedly traverses a fixed sample collection.

\section{Pre-training}
\label{sec:pretrain}

The preceding section introduced the sampling and horizon-control mechanisms that define the training distribution of \NAME{}. This section specifies how those mechanisms are instantiated in the pre-training run, including the corpus, optimization schedule, alignment objective, and distributed execution setup.

\subsection{Training Data Corpus}
\label{sec:pretrain_data}

\NAME{} is pre-trained on a large-scale heterogeneous corpus spanning seven domains---Energy, Finance, Healthcare, Nature, Sales, Transport, and Cloud/IT---with each domain contributing multiple datasets of varying temporal lengths, sampling frequencies, and variate counts. Following the data leakage prevention principles established by GIFT-Eval~\citep{aksu2024gifteval}, we rigorously separate pre-training data from all evaluation benchmarks, ensuring that zero-shot performance reflects genuine generalization rather than memorization. Detailed information on the corpus composition and dataset statistics can be found in Appendix~\ref{sec:appendix_pre-training_corpus}.

\subsection{Training Configuration}
\label{sec:pretrain_config}

The full set of training hyperparameters is summarized in Table~\ref{tab:training_config}. \NAME{} is trained with 21-quantile regression using the pinball loss (Section~\ref{sec:arch_quantile}), with quantile levels
$\mathcal{Q} = \{0.01, 0.05, \ldots, 0.99\}$,
providing calibrated coverage from the extreme tails to the median. Optimization is performed with AdamW~($\beta_1 = 0.9$, $\beta_2 = 0.95$)~\citep{loshchilov2017decoupled} for $1{,}000{,}000$ optimizer steps. The learning rate peaks at $6 \times 10^{-5}$ and follows cosine decay for $999{,}000$ steps to a minimum of $6 \times 10^{-6}$, with a warmup fraction of $0.001$. Weight decay is set to $0.1$ and gradients are clipped to norm $1.0$. Training uses BF16 mixed precision with a per-GPU batch size of $64$ on NVIDIA B200-180GB GPU clusters. The Omni-Range parameters are set to $p_{\min} = 16$ and $p_{\max} = 96$, matching the single-pass horizon $H_1 = M \times P_{\text{out}} = 6 \times 16 = 96$ from Section~\ref{sec:arch_future}. We additionally enable the alignment auxiliary loss between encoder blocks $\ell_{\mathrm{sh}}=1$ and $\ell_{\mathrm{dp}}=31$ with weight $\lambda_{\mathrm{align}}=10.0$.

\begin{table}[t]
\centering
\caption{Pre-training configuration.}
\label{tab:training_config}
\tablestyle{6pt}{1.15}
\begin{tabular}{ll}
\toprule
\textbf{Configuration} & \textbf{Setting used in pre-training} \\
\midrule
Loss function & 21-quantile pinball regression \\
Quantile levels $\mathcal{Q}$ & $\{0.01, 0.05, 0.10, \ldots, 0.90, 0.95, 0.99\}$ \\
Optimizer & AdamW ($\beta_1{=}0.9,\; \beta_2{=}0.95$) \\
Peak learning rate & $6 \times 10^{-5}$ \\
LR schedule & Cosine annealing \\
Minimum learning rate & $6 \times 10^{-6}$ \\
Training iterations & $1{,}000{,}000$ \\
LR decay iterations & $999{,}000$ \\
Weight decay & 0.1 \\
Gradient clipping & 1.0 \\
Batch size & 64 \\
Precision & BF16 mixed precision \\
Min prediction length $p_{\min}$ & 16 \\
Max prediction length $p_{\max}$ & 96 \\
Alignment blocks $(\ell_{\mathrm{sh}},\ell_{\mathrm{dp}})$ & $(1,31)$, with $\lambda_{\mathrm{align}}=10.0$ \\
Training schedule & Step-based \\
\bottomrule
\end{tabular}
\end{table}

\subsection{Distributed Training}
\label{sec:pretrain_dist}

\NAME{} is trained with Megatron-LM~\citep{shoeybi2019megatron} on an NVIDIA B200-180GB GPU cluster using data parallelism and the distributed optimizer; tensor and pipeline parallel sizes are set to $1$ in the reported run. Each optimizer update aggregates per-rank micro-batches across the data-parallel group, while optimizer states are sharded to reduce memory pressure. To keep distributed data access deterministic and inexpensive, sample indices are constructed once on rank~0, cached under a configuration-dependent key, and then loaded by the remaining ranks after synchronization. This preserves identical sample ordering across ranks while avoiding redundant index construction during large-scale pre-training.

\section{Experiments}
\label{sec:experiments}

\subsection{Evaluation Benchmarks}
\label{sec:exp_benchmarks}

We evaluate forecasting performance on two complementary benchmarks. \emph{GIFT-Eval}~\citep{aksu2024gifteval} contains 23 datasets spanning seven domains and ten sampling frequencies. Its short-, medium-, and long-horizon settings form 97 dataset--frequency--horizon configurations, and its leakage-aware construction makes it a focused test of out-of-distribution generalization. We report Seasonal-Naive-normalized MASE for median point forecasts and Continuous Ranked Probability Score (CRPS) for probabilistic forecasts. Full dataset statistics and horizon definitions are provided in Appendix~\ref{sec:appendix_gift-eval}.

\emph{fev-bench}~\citep{shchur2025fev} broadens the evaluation to 100 tasks across seven domains, including 46 tasks with known-future covariates. We report normalized MASE and Weighted Quantile Loss (WQL), using the geometric mean across tasks as in the released leaderboard. The complete task list, frequency--horizon mapping, and evaluation-window construction are deferred to Appendix~\ref{sec:appendix_fev-eval}.

\subsection{Main Results}
\label{sec:exp_main}

\begin{figure}[!t]
    \centering
    \includegraphics[width=\textwidth]{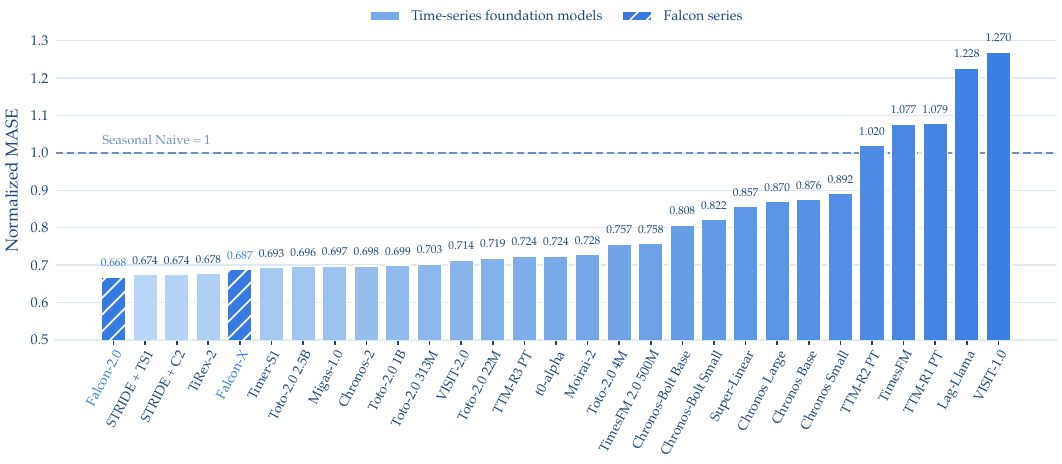}
    \caption{\textbf{GIFT-Eval pretrained-model comparison.} Seasonal-Naive-normalized MASE for the 29 pretrained models included in the GIFT-Eval leaderboard as of July 2026. Scores are geometrically aggregated over all 97 dataset--frequency--horizon configurations; lower is better.}
    \label{fig:gift_results}
\end{figure}

\begin{figure}[!t]
    \centering
    \includegraphics[width=\textwidth]{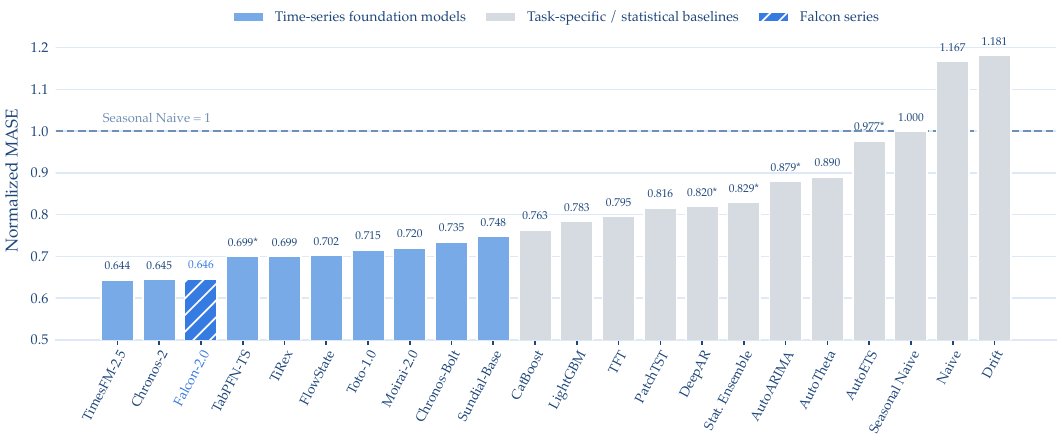}
    \caption{\textbf{fev-bench model comparison.} Seasonal-Naive-normalized MASE for all 22 models included in the fev-bench leaderboard as of July 2026. Scores are geometrically aggregated over the benchmark tasks; lower is better.}
    \label{fig:fev_results}
\end{figure}

Figures~\ref{fig:gift_results} and~\ref{fig:fev_results} compare \NAME{} with existing methods using leaderboard results available as of July 2026. For GIFT-Eval, we report all models categorized as \emph{pretrained}, together with \NAME{}; for fev-bench, we include the complete leaderboard. Both comparisons use Seasonal-Naive-normalized MASE, where lower values indicate better point-forecast accuracy. 

\textbf{Overall comparison.}\quad
\NAME{} establishes the strongest point-forecasting result in the GIFT-Eval comparison. Among the 29  evaluated pretrained models, it achieves both the lowest normalized MASE (\textbf{0.6684}) and the best mean MASE rank (\textbf{7.81}), improving over STRIDE + Timer-S1 (0.6744) by 0.9\%. The consistent agreement between the aggregate MASE and mean rank highlights that \NAME{}'s lead is not driven by isolated outsized wins, but rather reflects uniform strength across the 97 configurations. In terms of probabilistic performance, \NAME{} remains highly competitive, securing the seventh-lowest CRPS (0.4843) with a mean CRPS rank of 9.62, though STRIDE + Chronos-2 retains an edge on this specific dimension (0.4544; rank 6.84). 

The fev-bench evaluation (Figure~\ref{fig:fev_results}) further validates the robustness of \NAME{} under a more heterogeneous task suite. Its aggregate normalized MASE of 0.6459 is within 0.3\% of the top-performing TimesFM-2.5 (0.6438) and virtually tied with Chronos-2 (0.645), while establishing a superior mean MASE rank (5.15 vs. 5.63). Crucially, \NAME{} achieves the best aggregate WQL (\textbf{0.4842}) among all models while successfully completing all 100 tasks. No single baseline outperforms \NAME{} on both aggregate MASE and WQL simultaneously: TimesFM-2.5 yields slightly better point-forecasts but higher WQL, whereas Chronos-2 offers competitive mean ranks but inferior aggregate WQL. Consequently, \NAME{} occupies a highly desirable operating Pareto-frontier, successfully marrying near-optimal point accuracy with state-of-the-art probabilistic calibration.

Taken together, these cross-benchmark results confirm that \NAME{}'s capabilities generalize well beyond a narrow subset of tasks. This cross-benchmark strength provides the central empirical support for \NAME{}: explicitly controlling source exposure and interleaving context and horizon ranges during training translates into a model that is competitive across distinct forecasting regimes. The remaining GIFT-Eval CRPS gap and the task-level advantages of Chronos-2 define concrete directions for improving probabilistic consistency without diminishing \NAME{}'s established point-forecasting strength.

\begin{figure}[!t]
    \centering
    \includegraphics[width=0.96\textwidth]{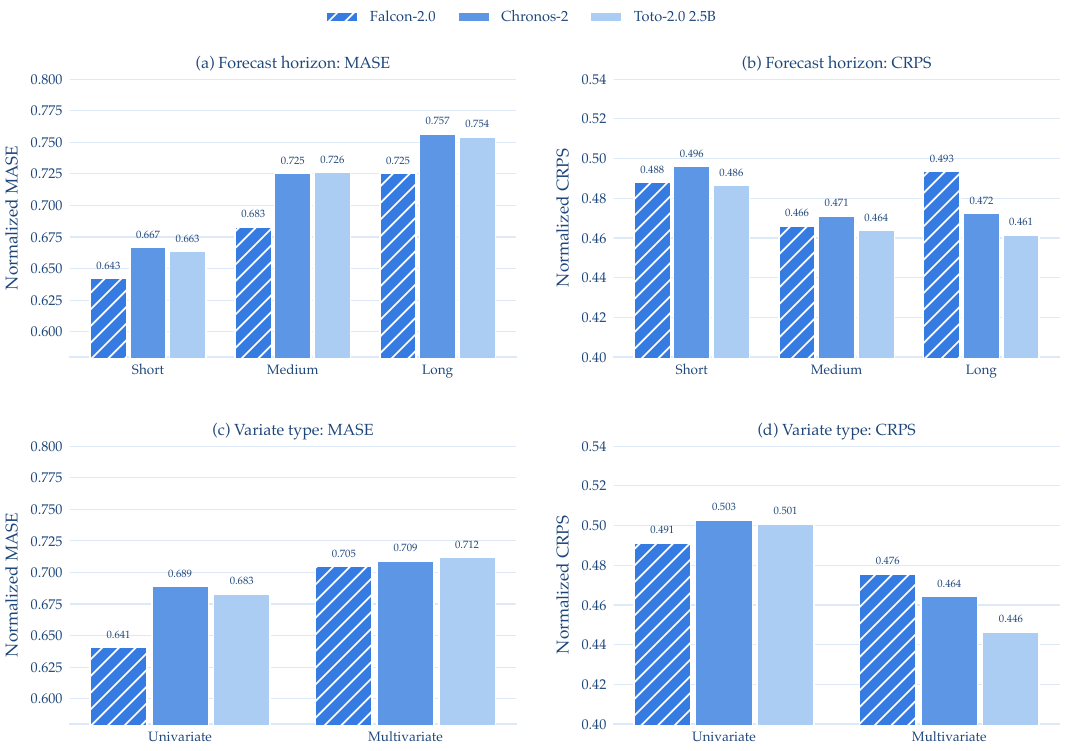}
    \caption{\textbf{GIFT-Eval fine-grained comparison.} Seasonal-Naive-normalized MASE for \NAME{}, Chronos-2, and Toto-2.0-2.5B; lower is better. Panels (a)--(b) report MASE and CRPS by short, medium, and long forecast horizons, while panels (c)--(d) report the same metrics for univariate and multivariate tasks. Bars show subgroup aggregates results.}
    \label{fig:gift_breakdown}
\end{figure}

\begin{figure}[!t]
    \centering
    \includegraphics[height=0.62\textheight,keepaspectratio]{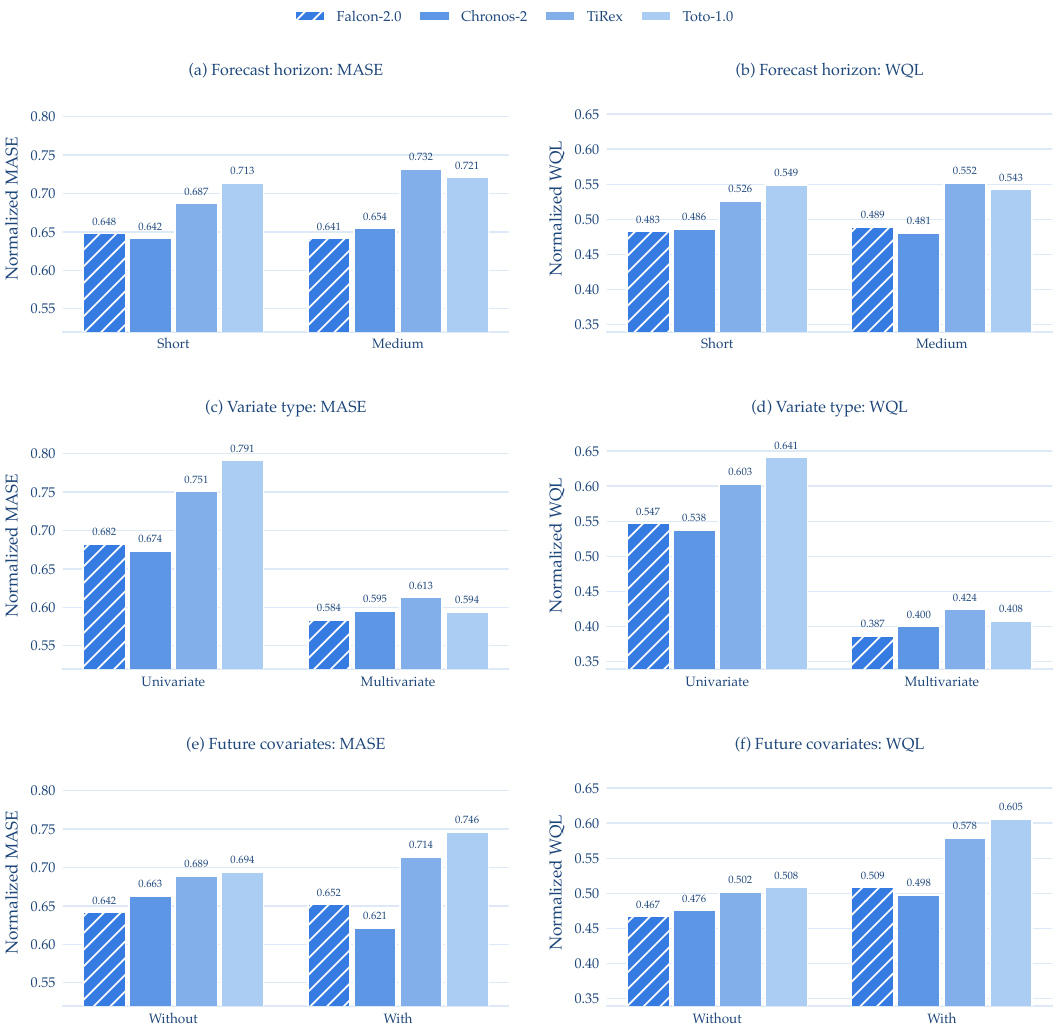}
    \caption{\textbf{fev-bench fine-grained comparison.} Seasonal-Naive-normalized MASE (left column) and WQL (right column) for \NAME{}, Chronos-2, TiRex, and Toto-1.0; lower is better. Panels (a)--(b), (c)--(d), and (e)--(f) decompose performance by forecast horizon, variate type, and availability of known-future covariates, respectively. Bars show subgroup aggregates results.}
    \label{fig:fev_breakdown}
\end{figure}

\textbf{Fine-grained behavior.}\quad
Figures~\ref{fig:gift_breakdown} and~\ref{fig:fev_breakdown} dissect the point and probabilistic accuracy of \NAME{} across forecast horizons, variate configurations, and the availability of known-future covariates.

On GIFT-Eval (Figure~\ref{fig:gift_breakdown}), \NAME{}'s normalized MASE changes smoothly from 0.643 on short horizons to 0.683 and 0.725 on medium and long horizons, outperforming both Chronos-2 and Toto-2.0-2.5B at every horizon. A similar advantage is observed across variate settings, where \NAME{} leads on both univariate (0.641) and multivariate (0.705) tasks. While its probabilistic predictions (CRPS) remain stable across horizons (ranging within 0.466--0.493), the fine-grained split exposes a multivariate bottleneck: \NAME{} leads on univariate probabilistic tasks (0.491 vs. 0.503 and 0.501) but trails Toto-2.0-2.5B on multivariate tasks (0.476 vs. 0.446) and at long horizons (0.493 vs. 0.461).

The fev-bench decomposition (Figure~\ref{fig:fev_breakdown}) isolates a critical architectural boundary. On the 54 tasks \emph{without} known-future covariates, \NAME{} significantly outperforms Chronos-2 in both MASE (0.642 vs. 0.663) and WQL (0.467 vs. 0.476). However, on the 46 tasks \emph{with} covariates, where \NAME{}'s autoregressive interface does not ingest future features, this performance ordering reverses (MASE of 0.652 vs. 0.621; WQL of 0.509 vs. 0.498). Crucially, horizon length itself does not degrade performance significantly—MASE/WQL shift minimally from 0.648/0.483 (short) to 0.641/0.489 (medium). These findings suggest that the residual performance gap on fev-bench stems primarily from covariate conditioning limitations rather than sensitivity to the forecast window.

\begin{figure}[!t]
    \centering
    \includegraphics[width=0.96\textwidth]{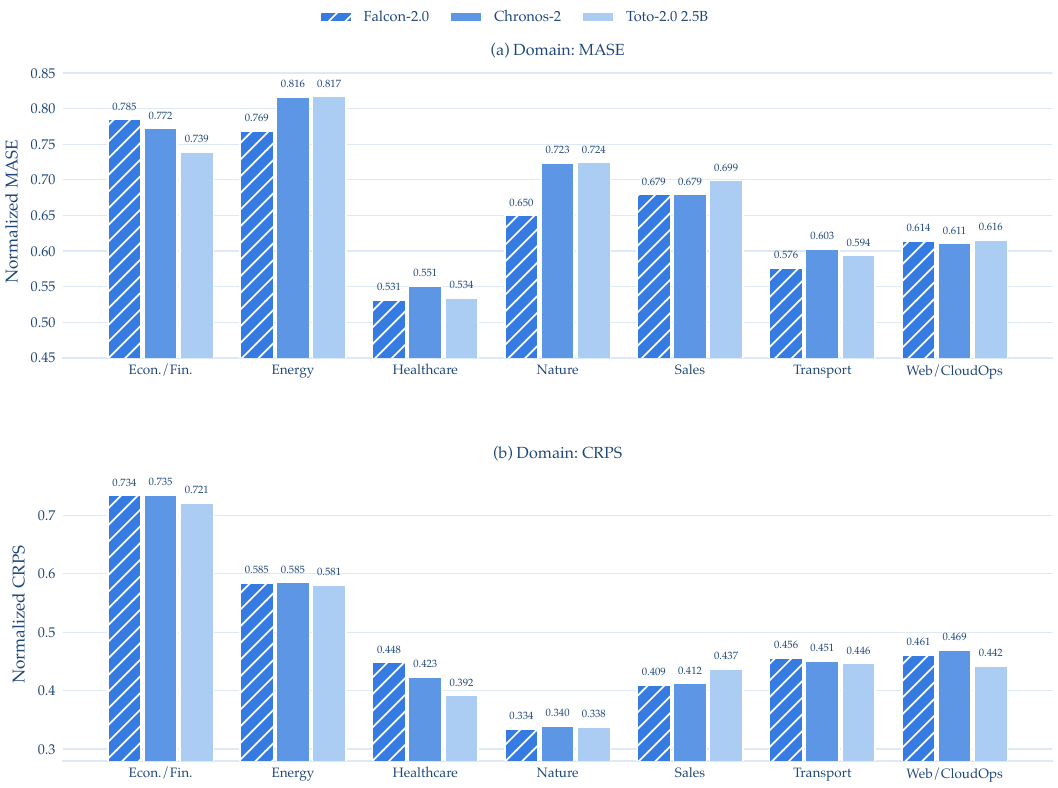}
    \caption{\textbf{GIFT-Eval domain-level comparison.} Seasonal-Naive-normalized MASE (a) and CRPS (b) across the seven domains for \NAME{}, Chronos-2, and Toto-2.0-2.5B; lower is better.}
    \label{fig:gift_domain}
\end{figure}

\begin{figure}[!t]
    \centering
    \includegraphics[width=0.913\textwidth]{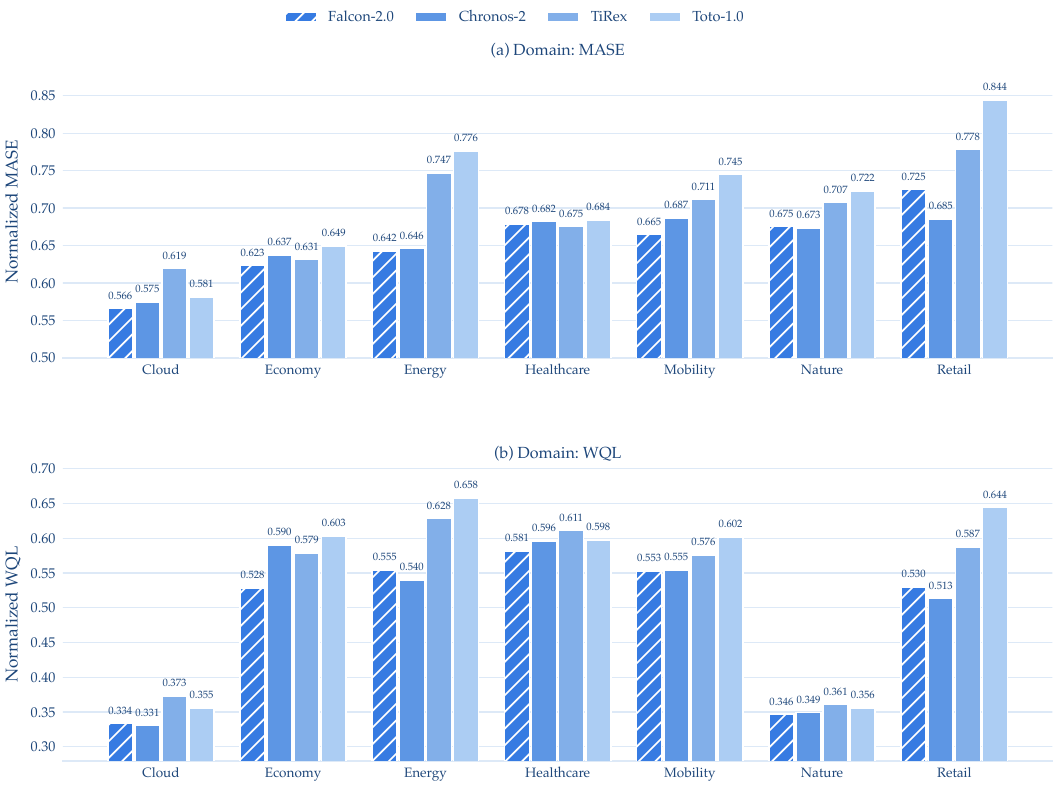}
    \caption{\textbf{fev-bench domain-level comparison.} Seasonal-Naive-normalized MASE (a) and WQL (b) across the seven domains for \NAME{}, Chronos-2, TiRex, and Toto-1.0; lower is better.}
    \label{fig:fev_domain}
\end{figure}

\textbf{Domain-level analysis.}\quad
We further analyze domain-specific performance in Figures~\ref{fig:gift_domain} and~\ref{fig:fev_domain} to identify where \NAME{}'s aggregate gains originate.

On GIFT-Eval (Figure~\ref{fig:gift_domain}), \NAME{} achieves the lowest normalized MASE in four of seven domains: Energy (0.769), Healthcare (0.531), Nature (0.650), and Transport (0.576). The largest margin occurs in Nature (0.650 versus 0.723 for the next-best model, a 10.2\% reduction), while Sales and Web/CloudOps are effectively tied across models. In contrast, Econ/Fin remains a point-forecasting weakness, where Toto-2.0-2.5B dominates (0.739 vs. 0.785). The domain-level CRPS aligns with this trend: \NAME{} leads only in Nature (0.334) and Sales (0.409), whereas Toto-2.0-2.5B leads in five domains, confirming that \NAME{}’s global superiority on GIFT-Eval is largely anchored by its highly robust point predictions.

On fev-bench (Figure~\ref{fig:fev_domain}), \NAME{} demonstrates broad domain-level coverage, leading MASE in Cloud (0.566), Economy (0.623), Energy (0.642), and Mobility (0.665), while leading WQL in Economy (0.528), Healthcare (0.581), Mobility (0.553), and Nature (0.346). Its probabilistic performance in the Economy domain is exceptional, providing an 8.8\% relative WQL reduction over the next-best model (0.528 vs. 0.579). Nevertheless, Chronos-2 maintains dominance in the Retail domain across both metrics (0.685/0.513 vs. \NAME{}'s 0.725/0.530) and secures the top WQL in Cloud and Energy. These findings pinpoint covariate-rich domains and specialized Retail regimes as key frontiers for future refinement of \NAME{}'s probabilistic consistency.

\subsection{Scaling Behavior}
\label{sec:exp_scaling}
We investigate the scaling behavior of \NAME{} along data exposure and model capacity.

\subsubsection{Data Scaling}
\label{sec:exp_data_scaling}

\begin{figure}[!t]
    \centering
    \includegraphics[width=0.96\textwidth]{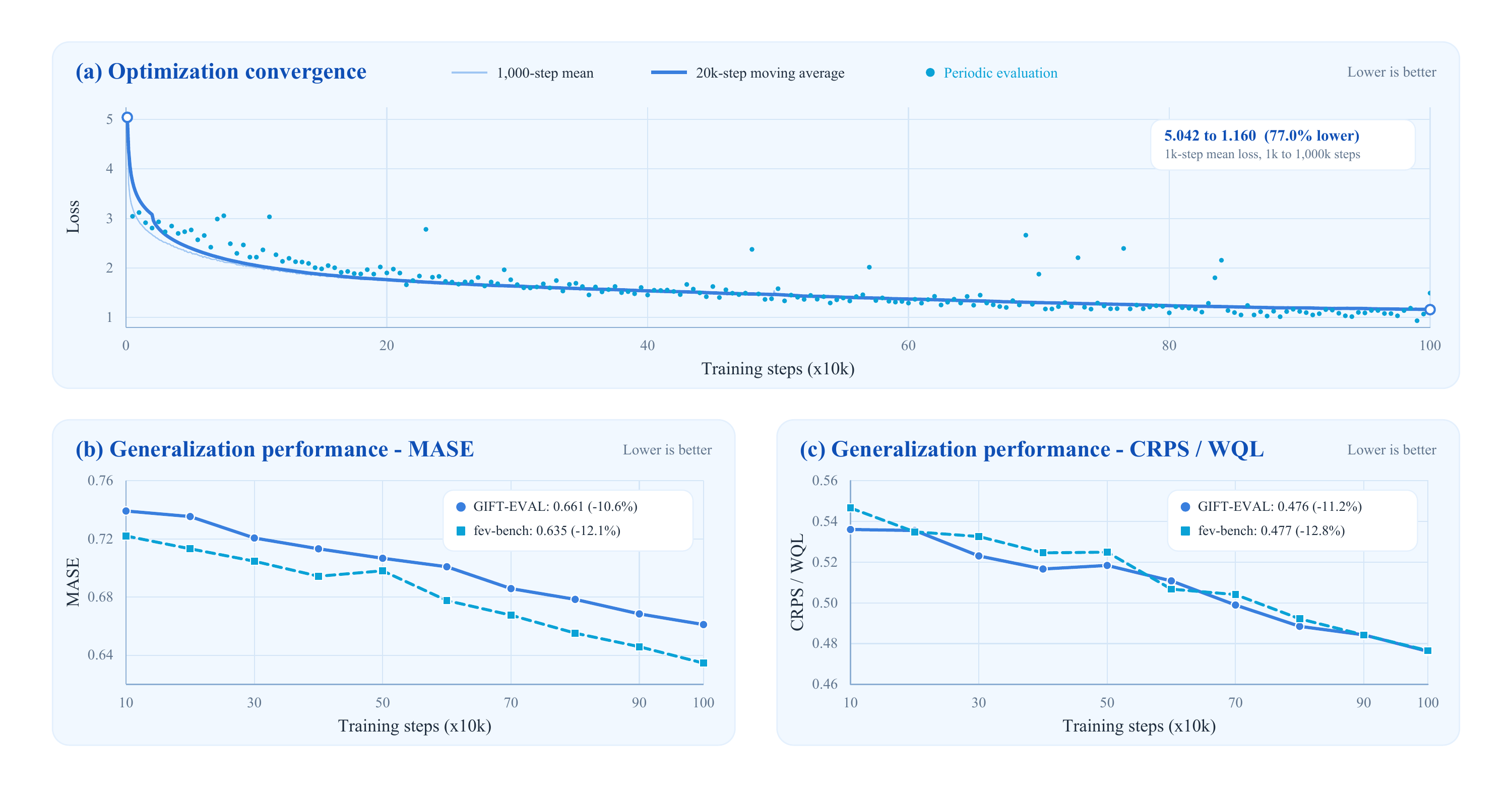}
    \caption{\textbf{Training convergence and benchmark performance.} (a) The 1,000-step mean training loss decreases from 5.042 at the beginning of logging to 1.160 at one million training steps (77.0\% reduction); the dark-blue curve denotes a 20k-step moving average and cyan markers denote periodic evaluation loss. (b)--(c) Checkpoint evaluations from 100k to one million training steps show consistent overall improvements on both GIFT-Eval and fev-bench. At the final checkpoint, MASE reaches 0.661 and 0.635 on GIFT-Eval and fev-bench, respectively; CRPS on GIFT-Eval reaches 0.476, while WQL on fev-bench reaches 0.477.}
    \label{fig:training_scaling}
\end{figure}

Because \NAME{} reconstructs batches from its bootstrap distribution throughout step-based training, increasing the training budget increases cumulative exposure to heterogeneous records, variables, contexts, and horizons. Holding the 585M-parameter architecture fixed, Figure~\ref{fig:training_scaling} measures how this additional data exposure affects optimization and forecasting performance.

The smoothed loss decreases throughout training despite occasional evaluation spikes, indicating stable optimization under the heterogeneous sampling regime. Generalization improves in parallel: from 100k to one million steps, MASE falls by 10.6\% on GIFT-Eval and 12.1\% on fev-bench, while CRPS on GIFT-Eval and WQL on fev-bench fall by 11.2\% and 12.8\%, respectively. Both benchmarks attain their best checkpoint-level scores at the end of training, with only minor intermediate fluctuations. Thus, the additional sampled-data exposure continues to transfer across benchmarks rather than producing a late-stage generalization reversal, although the flattening loss curve suggests diminishing marginal returns.

\subsubsection{Model Scaling}
\label{sec:exp_model_scaling}

\begin{figure}[!t]
    \centering
    \includegraphics[width=0.96\textwidth]{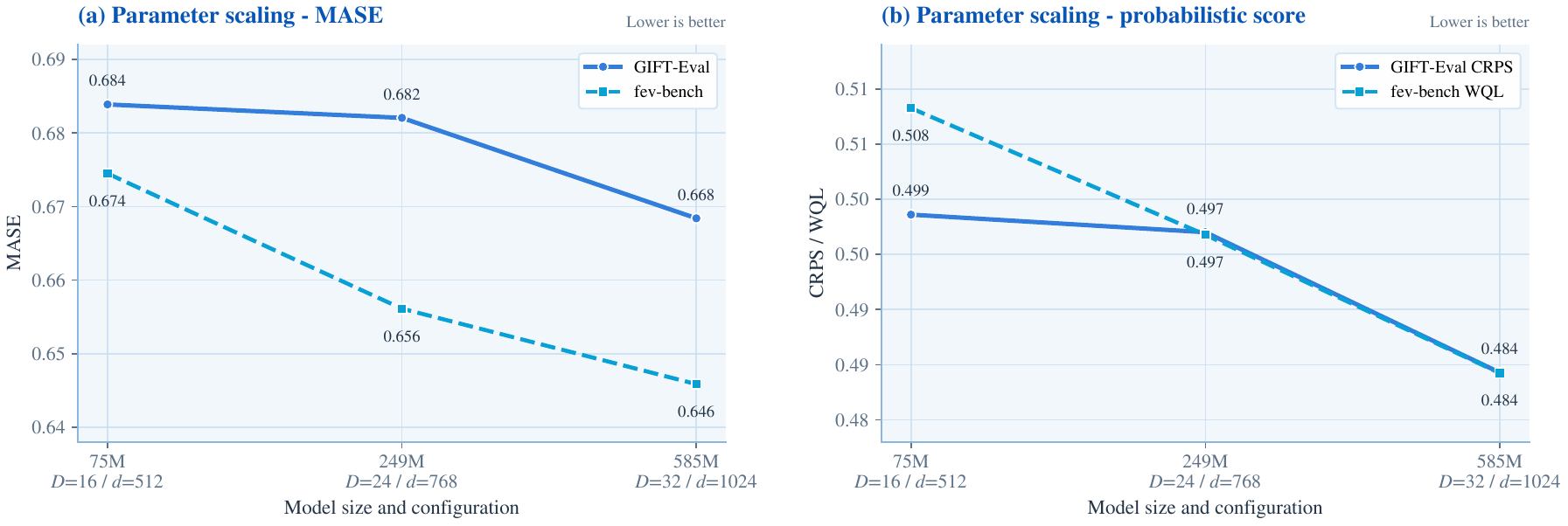}
    \caption{\textbf{Parameter scaling at a fixed one-million-iteration budget.} Performance of the 75M, 249M, and 585M variants after the same number of training iterations. (a) Seasonal-Naive-normalized MASE on GIFT-Eval and fev-bench. (b) CRPS on GIFT-Eval and WQL on fev-bench. Lower is better.}
    \label{fig:parameter_scaling}
\end{figure}

We next investigate the effect of model scaling under a fixed training budget of one million iterations. Specifically, we compare the default 585M configuration ($D=32$, $d=1024$; Table~\ref{tab:model_config}) against two progressively smaller variants: a 249M model with $D=24$ and $d=768$, and a 75M model with $D=16$ and $d=512$. All three variants are trained and evaluated using the same protocol, ensuring that the comparison primarily reflects differences in model capacity rather than training conditions. Figure~\ref{fig:parameter_scaling} summarizes their performance at the final checkpoint.

Increasing capacity improves all four metrics without a reversal. On GIFT-Eval, scaling from 75M to 585M reduces MASE from 0.6839 to 0.6684 and CRPS from 0.4986 to 0.4843, corresponding to relative reductions of 2.3\% and 2.9\%. On fev-bench, MASE falls from 0.6745 to 0.6459 and WQL from 0.5083 to 0.4842, corresponding to 4.2\% and 4.7\% reductions. The 249M model already captures much of the fev-bench improvement, whereas the largest GIFT-Eval gain appears between 249M and 585M. Across these three capacities, the consistent direction of change supports a stable capacity--performance relationship.

\subsection{Ablation Study}
\label{sec:exp_ablation}

\subsubsection{Architecture}
\label{sec:exp_ablation_arch}

\begin{figure}[!t]
    \centering
    \includegraphics[width=0.92\textwidth]{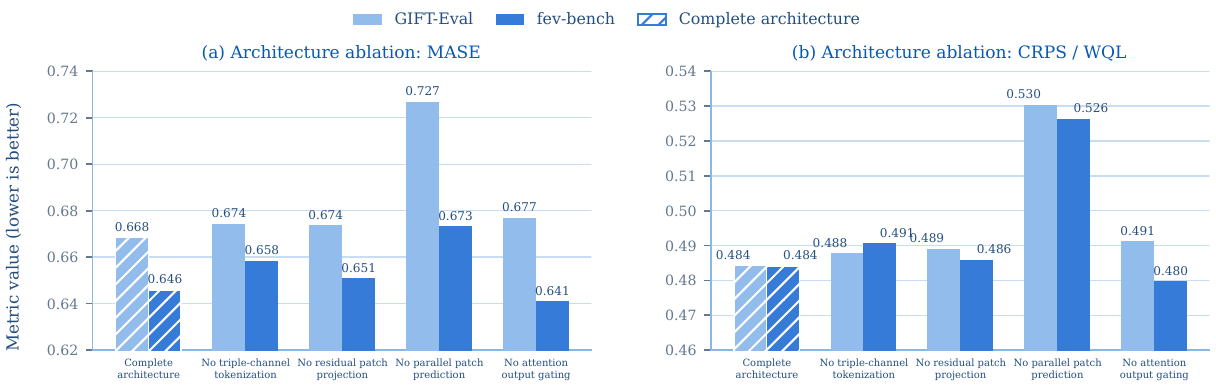}
    \caption{\textbf{Architecture ablation of \NAME{}.} The complete 585M architecture is compared with four variants, each removing one component under the same training and evaluation protocol. Panel (a) reports Seasonal-Naive-normalized MASE on GIFT-Eval and fev-bench. Panel (b) reports CRPS on GIFT-Eval and WQL on fev-bench. Hatched bars denote the complete architecture. Scores are geometrically aggregated over configurations or tasks, and lower values indicate better performance.}
    \label{fig:architecture_ablation}
\end{figure}

We evaluate four architectural choices in \NAME{}: Triple-Channel Patch Tokenization, the shared residual SwiGLU patch projection $\phi$, Parallel Patch Prediction, and output gating in self-attention. Each ablated variant removes one component from the full model while following the same training and evaluation protocol. Figure~\ref{fig:architecture_ablation} reports the geometrically aggregated scores on both benchmarks; lower values indicate better performance.

Parallel Patch Prediction has the largest and most consistent effect. Relative to the variant without this component, the full architecture reduces MASE and CRPS on GIFT-Eval by 8.0\% and 8.7\%, respectively, and MASE and WQL on fev-bench by 4.1\% and 8.0\%, respectively. The degradation across both point and probabilistic metrics indicates that direct multi-patch forecasting makes the largest observed contribution among the architectural choices evaluated here.

Triple-Channel Patch Tokenization and the residual SwiGLU patch projection provide smaller but consistent improvements. Restoring the triple-channel representation lowers the four error metrics by 0.7--1.9\% relative to its ablation, while restoring the residual SwiGLU patch projection $\phi$ lowers them by 0.4--1.0\%. Output gating in self-attention has a comparatively modest effect. It lowers GIFT-Eval MASE and CRPS from 0.677 and 0.491 to 0.668 and 0.484, respectively. On fev-bench, the corresponding differences remain below 1\% for both metrics. Overall, these results identify Parallel Patch Prediction as the principal architectural contributor, while the remaining components have more modest effects.

\subsubsection{Sampling}
\label{sec:exp_ablation_sampling}

\begin{figure}[!t]
    \centering
    \includegraphics[width=0.93\textwidth]{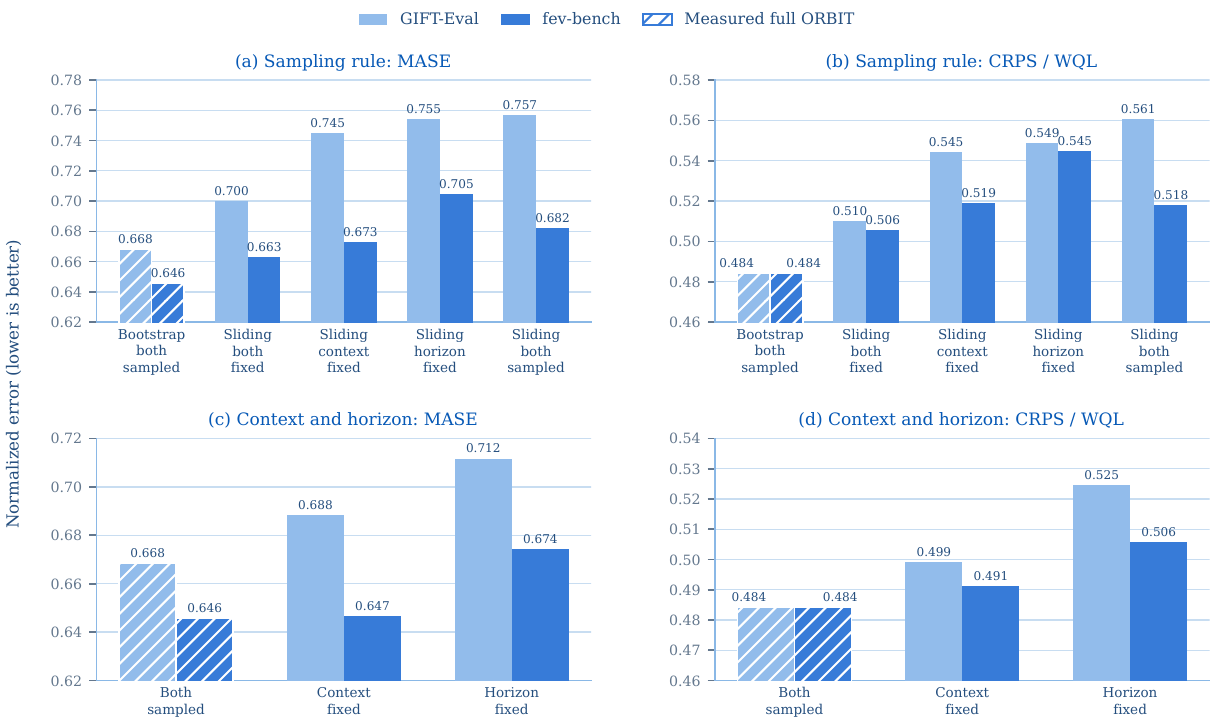}
    \caption{\textbf{Sampling ablation of \NAME{}.} All configurations use the 585M \NAME{} model. Panels (a)--(b) compare Bootstrap Stochastic Sampling with four sliding-window variants covering all combinations of fixed and sampled context lengths and prediction horizons. Panels (c)--(d) assess the individual contributions of context and horizon sampling by comparing joint sampling with variants that fix one length while sampling the other. Panels (a) and (c) report normalized MASE, while panels (b) and (d) report CRPS on GIFT-Eval and WQL on fev-bench. Hatched bars denote the full \Orbit{} configuration. Lower values indicate better performance.}
    \label{fig:sampling_ablation}
\end{figure}

We conduct a controlled ablation to assess how different strategies for constructing samples within each dataset affect forecasting performance. All comparisons use the 585M \NAME{} architecture and the same corpus, training budget, optimization settings, and evaluation protocol. Specifically, we compare Bootstrap Stochastic Sampling with sliding-window enumeration and independently control whether the context length and prediction horizon are fixed or sampled from their feasible ranges. For the sliding-window variants, eligible cutoff positions are enumerated rather than sampled, while the context length and prediction horizon are fixed or sampled as specified by each configuration. Following the evaluation protocol in Section~\ref{sec:exp_benchmarks}, metric values are geometrically aggregated over configurations or tasks and reported in Figure~\ref{fig:sampling_ablation}.

\textbf{Sampling rule.}\quad
We first isolate the effect of the sampling rule by comparing Bootstrap Stochastic Sampling with the sliding-window variant that samples context lengths and prediction horizons from the same feasible ranges. Relative to this sliding-window variant, Bootstrap Stochastic Sampling reduces MASE and CRPS on GIFT-Eval by 11.7\% and 13.6\%, respectively, and MASE and WQL on fev-bench by 5.4\% and 6.5\%. The gains are consistent across point and probabilistic metrics on both benchmarks. Among the sliding-window variants, fixing both lengths outperforms sampling either or both, yet remains worse than Bootstrap Stochastic Sampling on all four metrics. This pattern suggests that varying context lengths and prediction horizons alone cannot offset the redundancy introduced by enumerating adjacent cutoff positions. Stochastic selection across records, variables, and temporal positions also contributes to sample diversity.

\textbf{Context and horizon sampling.}\quad
We next isolate the effects of sampling context lengths and prediction horizons within Bootstrap Stochastic Sampling. Compared with the fixed-horizon variant, joint sampling reduces MASE and CRPS on GIFT-Eval by 6.1\% and 7.7\%, respectively, and MASE and WQL on fev-bench by 4.2\% each, showing that horizon sampling has the larger effect. Compared with the fixed-context variant, joint sampling reduces GIFT-Eval MASE and CRPS by 2.9\% and 3.0\%, respectively. On fev-bench, the MASE difference is small (0.6459 versus 0.6465), although joint sampling also improves WQL. Joint sampling is the only configuration to achieve the lowest error on all four metrics, supporting the simultaneous sampling of diverse context lengths and prediction horizons within a single training stage.

\FloatBarrier

\section{Conclusion}
\label{sec:conclusion}

We introduce \Orbit{}, a training paradigm that explicitly controls the effective pre-training distribution of heterogeneous time series corpora. Bootstrap Multi-Level Sampling regulates source exposure and constructs diverse examples across records, variables, temporal windows, and prediction horizons, while Omni-Range Incremental Training jointly covers varying context lengths and horizons within a single stage. Under \Orbit{}, we train \NAME{}, an encoder-only Transformer with missingness-aware triple-channel tokenization and parallel patch prediction, and explore Rank-Guided Cross-Depth Alignment to regularize shallow representations using deeper layers. Evaluations on GIFT-Eval and fev-bench show strong zero-shot forecasting performance, while ablation and scaling studies validate the benefits of controlled sampling, joint context-horizon coverage, and increased training exposure. These results highlight training-distribution design as a key factor in building scalable and generalizable time series foundation models.

\newpage
\bibliographystyle{assets/plainnat}
\bibliography{main}

\begin{thebibliography}{82}
\providecommand{\natexlab}[1]{#1}
\providecommand{\url}[1]{\texttt{#1}}
\expandafter\ifx\csname urlstyle\endcsname\relax
  \providecommand{\doi}[1]{doi: #1}\else
  \providecommand{\doi}{doi: \begingroup \urlstyle{rm}\Url}\fi

\bibitem[Admin and Cukierski(2014)]{Kaggle2014Walmart}
Walmart~Competition Admin and Will Cukierski.
\newblock Walmart recruiting - store sales forecasting.
\newblock \url{https://kaggle.com/competitions/walmart-recruiting-store-sales-forecasting}, 2014.
\newblock Kaggle.

\bibitem[Aksu et~al.(2024{\natexlab{a}})Aksu, Woo, Liu, Liu, Liu, Savarese, Xiong, and Sahoo]{aksu2024gift}
Taha Aksu, Gerald Woo, Juncheng Liu, Xu~Liu, Chenghao Liu, Silvio Savarese, Caiming Xiong, and Doyen Sahoo.
\newblock Gift-{EVAL}: A benchmark for general time series forecasting model evaluation.
\newblock \emph{arXiv preprint arXiv:2410.10393}, 2024{\natexlab{a}}.

\bibitem[Aksu et~al.(2024{\natexlab{b}})Aksu, Woo, Liu, Liu, Liu, Savarese, Xiong, and Sahoo]{aksu2024gifteval}
Taha Aksu, Gerald Woo, Juncheng Liu, Xu~Liu, Chenghao Liu, Silvio Savarese, Caiming Xiong, and Doyen Sahoo.
\newblock Gift-eval: A benchmark for general time series forecasting model evaluation.
\newblock \emph{arXiv preprint arXiv:2410.10393}, 2024{\natexlab{b}}.

\bibitem[Ansari et~al.(2024)Ansari, Stella, Turkmen, Zhang, Mercado, Shen, Shchur, Rangapuram, Arango, Kapoor, et~al.]{ansari2024chronos}
Abdul~Fatir Ansari, Lorenzo Stella, Caner Turkmen, Xiyuan Zhang, Pedro Mercado, Huibin Shen, Oleksandr Shchur, Syama~Sundar Rangapuram, Sebastian~Pineda Arango, Shubham Kapoor, et~al.
\newblock Chronos: Learning the language of time series.
\newblock \emph{Transactions on Machine Learning Research}, 2024.

\bibitem[Ansari et~al.(2025)Ansari, Shchur, K{\"u}ken, Auer, Han, Mercado, Rangapuram, Shen, Stella, Zhang, et~al.]{ansari2025chronos2}
Abdul~Fatir Ansari, Oleksandr Shchur, Jaris K{\"u}ken, Andreas Auer, Boran Han, Pedro Mercado, Syama~Sundar Rangapuram, Huibin Shen, Lorenzo Stella, Xiyuan Zhang, et~al.
\newblock Chronos-2: From univariate to universal forecasting.
\newblock \emph{arXiv preprint arXiv:2510.15821}, 2025.

\bibitem[Auer et~al.(2026)Auer, Podest, Klotz, B{\"o}ck, Klambauer, and Hochreiter]{olber2025tirex}
Andreas Auer, Patrick Podest, Daniel Klotz, Sebastian B{\"o}ck, G{\"u}nter Klambauer, and Sepp Hochreiter.
\newblock Tirex: Zero-shot forecasting across long and short horizons with enhanced in-context learning.
\newblock In \emph{The Thirty-ninth Annual Conference on Neural Information Processing Systems}, 2026.
\newblock \url{https://openreview.net/forum?id=v7UqniC9pF}.

\bibitem[Che et~al.(2018)Che, Purushotham, Cho, Sontag, and Liu]{che2018recurrent}
Zhengping Che, Sanjay Purushotham, Kyunghyun Cho, David Sontag, and Yan Liu.
\newblock Recurrent neural networks for multivariate time series with missing values.
\newblock \emph{Scientific reports}, 8\penalty0 (1):\penalty0 6085, 2018.

\bibitem[Chung et~al.(2024)Chung, Jo, Kwon, and Choi]{chung2024time}
Hyunseung Chung, Sumin Jo, Yeonsu Kwon, and Edward Choi.
\newblock Time is not enough: Time-frequency based explanation for time-series black-box models.
\newblock In \emph{Proceedings of the 33rd ACM International Conference on Information and Knowledge Management}, pages 394--403, 2024.

\bibitem[Cohen et~al.(2026)Cohen, Khwaja, Doubli, Lemaachi, Lettieri, Masson, Miccinilli, Ram{\'e}, Ren, Rostamizadeh, et~al.]{cohen2026toto}
Ben Cohen, Emaad Khwaja, Youssef Doubli, Salahidine Lemaachi, Chris Lettieri, Charles Masson, Hugo Miccinilli, Elise Ram{\'e}, Qiqi Ren, Afshin Rostamizadeh, et~al.
\newblock This time is different: An observability perspective on time series foundation models.
\newblock \emph{Advances in neural information processing systems}, 38:\penalty0 50907--50951, 2026.

\bibitem[Das et~al.(2024)Das, Kong, Sen, and Zhou]{das2023timesfm}
Abhimanyu Das, Weihao Kong, Rajat Sen, and Yichen Zhou.
\newblock A decoder-only foundation model for time-series forecasting.
\newblock In \emph{Proceedings of the 41st International Conference on Machine Learning}, ICML'24. JMLR.org, 2024.

\bibitem[Data(2020)]{OPSD2020}
Open Power~System Data.
\newblock Data package time series. version 2020-10-06, 2020.
\newblock \url{https://doi.org/10.25832/time_series/2020-10-06}.

\bibitem[data from official {UK}~government sources(2022)]{Kaggle2022UKCovidDashboard}
{UK COVID-19} data from official {UK}~government sources.
\newblock {UK COVID-19} dashboard data.
\newblock \url{https://www.kaggle.com/datasets/happyadam73/uk-covid19-dashboard-data-sqlite-compressed}, 2022.
\newblock Kaggle.

\bibitem[David et~al.(2022)David, Bellot, and Corff]{david2022hermes}
Etienne David, Jean Bellot, and Sylvain~Le Corff.
\newblock H{ERMES}: Hybrid error-corrector model with inclusion of external signals for nonstationary fashion time series.
\newblock \emph{arXiv preprint arXiv:2202.03224}, 2022.

\bibitem[{De Vito} et~al.(2008){De Vito}, Massera, Piga, Martinotto, and {Di Francia}]{DeVito2008ElectronicNoseCalibration}
S.~{De Vito}, E.~Massera, M.~Piga, L.~Martinotto, and G.~{Di Francia}.
\newblock On field calibration of an electronic nose for benzene estimation in an urban pollution monitoring scenario.
\newblock \emph{Sensors and Actuators B: Chemical}, 129\penalty0 (2):\penalty0 750--757, 2008.
\newblock ISSN 0925-4005.
\newblock \doi{https://doi.org/10.1016/j.snb.2007.09.060}.
\newblock \url{https://www.sciencedirect.com/science/article/pii/S0925400507007691}.

\bibitem[{ECDC}(2025)]{ECDC2025RespiratoryViruses}
{ECDC}.
\newblock Respiratory viruses weekly data.
\newblock \url{https://github.com/EU-ECDC/Respiratory_viruses_weekly_data/tree/main}, 2025.
\newblock Open data repository; weekly respiratory virus surveillance in the {EU/EEA}.

\bibitem[Ekambaram et~al.(2024)Ekambaram, Jati, Dayama, Mukherjee, Nguyen, Gifford, Reddy, and Kalagnanam]{ekambaram2024ttm}
Vijay Ekambaram, Arindam Jati, Pankaj Dayama, Sumanta Mukherjee, Nam~H Nguyen, Wesley~M. Gifford, Chandra Reddy, and Jayant Kalagnanam.
\newblock Tiny time mixers ({TTM}s): Fast pre-trained models for enhanced zero/few-shot forecasting of multivariate time series.
\newblock In \emph{The Thirty-eighth Annual Conference on Neural Information Processing Systems}, 2024.
\newblock \url{https://openreview.net/forum?id=3O5YCEWETq}.

\bibitem[Fleming and Wallace(1986)]{lago2021forecasting}
Philip~J Fleming and John~J Wallace.
\newblock How not to lie with statistics: the correct way to summarize benchmark results.
\newblock \emph{Communications of the ACM}, 29\penalty0 (3):\penalty0 218--221, 1986.

\bibitem[FlorianKnauer and Cukierski(2015)]{Kaggle2015Rossmann}
FlorianKnauer and Will Cukierski.
\newblock Rossmann store sales.
\newblock \url{https://kaggle.com/competitions/rossmann-store-sales}, 2015.
\newblock Kaggle.

\bibitem[Godahewa et~al.(2021)Godahewa, Bergmeir, Webb, Hyndman, and Montero-Manso]{godahewa2021monash}
Rakshitha~Wathsadini Godahewa, Christoph Bergmeir, Geoffrey~I. Webb, Rob Hyndman, and Pablo Montero-Manso.
\newblock Monash time series forecasting archive.
\newblock In \emph{The Conference on Neural Information Processing Systems Datasets and Benchmarks Track}, 2021.
\newblock \url{https://openreview.net/forum?id=wEc1mgAjU-}.

\bibitem[Goswami et~al.(2024{\natexlab{a}})Goswami, Szafer, Choudhry, Cai, Li, and Dubrawski]{goswami2024moment}
Mononito Goswami, Konrad Szafer, Arjun Choudhry, Yifu Cai, Shuo Li, and Artur Dubrawski.
\newblock Moment: a family of open time-series foundation models.
\newblock In \emph{Proceedings of the 41st International Conference on Machine Learning}, ICML'24. JMLR.org, 2024{\natexlab{a}}.

\bibitem[Goswami et~al.(2024{\natexlab{b}})Goswami, Szafer, Choudhry, Cai, Li, and Dubrawski]{goswamimoment}
Mononito Goswami, Konrad Szafer, Arjun Choudhry, Yifu Cai, Shuo Li, and Artur Dubrawski.
\newblock M{OMENT}: A family of open time-series foundation models.
\newblock In \emph{International Conference on Machine Learning}, 2024{\natexlab{b}}.

\bibitem[Grinsztajn et~al.(2026)Grinsztajn, Fl{\"o}ge, Key, Birkel, Jund, Roof, Manium, Hoo, B{\"u}hler, Garg, et~al.]{grinsztajn2026tabpfn3}
L{\'e}o Grinsztajn, Klemens Fl{\"o}ge, Oscar Key, Felix Birkel, Philipp Jund, Brendan Roof, Mihir Manium, Shi~Bin Hoo, Magnus B{\"u}hler, Anurag Garg, et~al.
\newblock Tabpfn-3: Technical report.
\newblock \emph{arXiv preprint arXiv:2605.13986}, 2026.

\bibitem[HE et~al.(2026)HE, Huang, Jiang, Li, Lian, Xie, Chen, xijie liang, Zhengzengrong, and Lee]{woo2024gtm}
Cheng HE, Xu~Huang, Gangwei Jiang, Zhaoyi Li, Defu Lian, Hong Xie, Enhong Chen, xijie liang, Zhengzengrong, and Patrick Lee.
\newblock {GTM}: A general time-series model for enhanced representation learning of time-series data.
\newblock In \emph{The Fourteenth International Conference on Learning Representations}, 2026.
\newblock \url{https://openreview.net/forum?id=PWM6FERWz9}.

\bibitem[Hong et~al.(2014)Hong, Pinson, and Fan]{hong2014global}
Tao Hong, Pierre Pinson, and Shu Fan.
\newblock Global energy forecasting competition 2012.
\newblock \emph{International Journal of Forecasting}, 30\penalty0 (2):\penalty0 357--363, 2014.

\bibitem[Hoo et~al.(2025)Hoo, M{\"u}ller, Salinas, and Hutter]{hoo2025tables}
Shi~Bin Hoo, Samuel M{\"u}ller, David Salinas, and Frank Hutter.
\newblock From tables to time: Extending tabpfn-v2 to time series forecasting.
\newblock \emph{arXiv preprint arXiv:2501.02945}, 2025.

\bibitem[Howard et~al.(2017)Howard, Yui, McDonald, and Cukierski]{howard2017recruit}
Addison Howard, Haruka Yui, Mark McDonald, and Will Cukierski.
\newblock Recruit restaurant visitor forecasting.
\newblock \url{https://kaggle.com/competitions/recruit-restaurant-visitor-forecasting}, 2017.
\newblock Kaggle.

\bibitem[Hu et~al.()Hu, Yang, Dai, Cai, Ding, Li, Liu, Ma, Qu, Wang, et~al.]{hulandscape}
Yifan Hu, Jie Yang, Xilin Dai, Wanxu Cai, Kuiye Ding, Yuante Li, Qinghua Liu, Enze Ma, Zhiyuan Qu, Yixin Wang, et~al.
\newblock The landscape of agentic time series systems: Architectures, reliability, and frontiers.

\bibitem[Hu et~al.(2025{\natexlab{a}})Hu, Liu, Zhu, Cheng, and Dai]{hu2025adaptive}
Yifan Hu, Peiyuan Liu, Peng Zhu, Dawei Cheng, and Tao Dai.
\newblock Adaptive multi-scale decomposition framework for time series forecasting.
\newblock In \emph{Proceedings of the AAAI Conference on Artificial Intelligence}, volume~39, pages 17359--17367, 2025{\natexlab{a}}.

\bibitem[Hu et~al.(2025{\natexlab{b}})Hu, Yang, Zhou, Liu, Tang, Jin, and Sun]{hu2025bridging}
Yifan Hu, Jie Yang, Tian Zhou, Peiyuan Liu, Yujin Tang, Rong Jin, and Liang Sun.
\newblock Bridging past and future: Distribution-aware alignment for time series forecasting.
\newblock \emph{arXiv preprint arXiv:2509.14181}, 2025{\natexlab{b}}.

\bibitem[Hu et~al.(2025{\natexlab{c}})Hu, Zhang, Liu, Lan, Li, Cheng, Dai, Xia, and Pan]{hu2025timefilter}
Yifan Hu, Guibin Zhang, Peiyuan Liu, Disen Lan, Naiqi Li, Dawei Cheng, Tao Dai, Shu-Tao Xia, and Shirui Pan.
\newblock Timefilter: Patch-specific spatial-temporal graph filtration for time series forecasting.
\newblock In \emph{International Conference on Machine Learning}, pages 24893--24911. PMLR, 2025{\natexlab{c}}.

\bibitem[Hu et~al.(2026)Hu, Chen, Liu, Liu, Dong, and Yang]{hu2026existence}
Yifan Hu, Hongzhou Chen, Peiyuan Liu, Yiding Liu, Zewei Dong, and Jiang-Ming Yang.
\newblock Existence precedes value: Joint modeling of observational existence and evolving states in time series forecasting.
\newblock \emph{arXiv preprint arXiv:2606.13571}, 2026.

\bibitem[Jiang et~al.(2025)Jiang, Wang, Li, Zhang, Wang, Wei, Dai, Zhang, and Wang]{jiang2025no}
Dengyang Jiang, Mengmeng Wang, Liuzhuozheng Li, Lei Zhang, Haoyu Wang, Wei Wei, Guang Dai, Yanning Zhang, and Jingdong Wang.
\newblock No other representation component is needed: Diffusion transformers can provide representation guidance by themselves.
\newblock \emph{arXiv preprint arXiv:2505.02831}, 2025.

\bibitem[Khwaja et~al.(2026)Khwaja, Lettieri, Woo, Belouadah, Cenac, Jarry, Paquin, Zhao, Zhukov, Abou-Amal, et~al.]{khwaja2026toto2}
Emaad Khwaja, Chris Lettieri, Gerald Woo, Eden Belouadah, Marc Cenac, Guillaume Jarry, Enguerrand Paquin, Xunyi Zhao, Viktoriya Zhukov, Othmane Abou-Amal, et~al.
\newblock Toto 2.0: Time series forecasting enters the scaling era.
\newblock \emph{arXiv preprint arXiv:2605.20119}, 2026.

\bibitem[Kim et~al.(2022)Kim, Kim, Tae, Park, Choi, and Choo]{kim2022revin}
Taesung Kim, Jinhee Kim, Yunwon Tae, Cheonbok Park, Jang-Ho Choi, and Jaegul Choo.
\newblock Reversible instance normalization for accurate time-series forecasting against distribution shift.
\newblock In \emph{International Conference on Learning Representations}, 2022.
\newblock \url{https://openreview.net/forum?id=cGDAkQo1C0p}.

\bibitem[Koenker and Hallock(2001)]{koenker2001quantile}
Roger Koenker and Kevin~F Hallock.
\newblock Quantile regression.
\newblock \emph{Journal of economic perspectives}, 15\penalty0 (4):\penalty0 143--156, 2001.

\bibitem[Kottapalli et~al.(2025)Kottapalli, Hubli, Chandrashekhara, Jain, Hubli, Botla, and Doddaiah]{kottapalli2025foundationmodelstimeseries}
Siva Rama~Krishna Kottapalli, Karthik Hubli, Sandeep Chandrashekhara, Garima Jain, Sunayana Hubli, Gayathri Botla, and Ramesh Doddaiah.
\newblock Foundation models for time series: A survey.
\newblock \emph{arXiv preprint arXiv:2504.04011}, 2025.

\bibitem[Lai et~al.(2017)Lai, Chang, Yang, and Liu]{Lai2017ModelingLA}
Guokun Lai, Wei-Cheng Chang, Yiming Yang, and Hanxiao Liu.
\newblock Modeling long- and short-term temporal patterns with deep neural networks.
\newblock In \emph{The International ACM SIGIR Conference on Research \& Development in Information Retrieval}, 2017.
\newblock \url{https://api.semanticscholar.org/CorpusID:4922476}.

\bibitem[lexis Cook et~al.(2020)lexis Cook, DanB, inversion, and Holbrook]{Kaggle2020StoreSales}
lexis Cook, DanB, inversion, and Ryan Holbrook.
\newblock Store sales -- time series forecasting.
\newblock \url{https://www.kaggle.com/competitions/store-sales-time-series-forecasting}, 2020.
\newblock Kaggle.

\bibitem[Liang et~al.(2024)Liang, Wen, Nie, Jiang, Jin, Song, Pan, and Wen]{Liang_2024}
Yuxuan Liang, Haomin Wen, Yuqi Nie, Yushan Jiang, Ming Jin, Dongjin Song, Shirui Pan, and Qingsong Wen.
\newblock Foundation models for time series analysis: A tutorial and survey.
\newblock In \emph{Proceedings of the 30th ACM SIGKDD Conference on Knowledge Discovery and Data Mining}, KDD ’24, page 6555–6565. ACM, August 2024.
\newblock \doi{10.1145/3637528.3671451}.
\newblock \url{http://dx.doi.org/10.1145/3637528.3671451}.

\bibitem[Lim et~al.(2021)Lim, Ar{\i}k, Loeff, and Pfister]{lim2021temporal}
Bryan Lim, Sercan~{\"O} Ar{\i}k, Nicolas Loeff, and Tomas Pfister.
\newblock Temporal fusion transformers for interpretable multi-horizon time series forecasting.
\newblock \emph{International Journal of Forecasting}, 37\penalty0 (4):\penalty0 1748--1764, 2021.

\bibitem[Liu et~al.(2025{\natexlab{a}})Liu, Aksu, Liu, Liu, Yan, Pham, Savarese, Sahoo, Xiong, and Li]{liu2025moirai2}
Chenghao Liu, Taha Aksu, Juncheng Liu, Xu~Liu, Hanshu Yan, Quang Pham, Silvio Savarese, Doyen Sahoo, Caiming Xiong, and Junnan Li.
\newblock Moirai 2.0: When less is more for time series forecasting.
\newblock \emph{arXiv preprint arXiv:2511.11698}, 2025{\natexlab{a}}.

\bibitem[Liu et~al.(2026{\natexlab{a}})Liu, Hu, Xia, Liu, Chen, Dai, Dong, and Yang]{liu2026falconx}
Yiding Liu, Yifan Hu, Hongjie Xia, Peiyuan Liu, Hongzhou Chen, Xilin Dai, Zewei Dong, and Jiang-Ming Yang.
\newblock Falcon-x: A time series foundation model for heterogeneous multivariate modeling.
\newblock \emph{arXiv preprint arXiv:2605.27286}, 2026{\natexlab{a}}.

\bibitem[Liu et~al.(2024{\natexlab{a}})Liu, Hu, Zhang, Wu, Wang, Ma, and Long]{liu2024itransformer}
Yong Liu, Tengge Hu, Haoran Zhang, Chenyu Wu, Shiyu Wang, Lintao Ma, and Mingsheng Long.
\newblock itransformer: Inverted transformers are effective for time series forecasting.
\newblock \emph{International Conference on Learning Representations (ICLR)}, 2024{\natexlab{a}}.

\bibitem[Liu et~al.(2024{\natexlab{b}})Liu, Zhang, Li, Huang, Wang, and Long]{liu2024timer}
Yong Liu, Haoran Zhang, Chenyu Li, Xiangdong Huang, Jianmin Wang, and Mingsheng Long.
\newblock Timer: generative pre-trained transformers are large time series models.
\newblock In \emph{Proceedings of the 41st International Conference on Machine Learning}, ICML'24. JMLR.org, 2024{\natexlab{b}}.

\bibitem[Liu et~al.(2025{\natexlab{b}})Liu, Qin, Huang, Wang, and Long]{liu2025timerxl}
Yong Liu, Guo Qin, Xiangdong Huang, Jianmin Wang, and Mingsheng Long.
\newblock Timer-{XL}: Long-context transformers for unified time series forecasting.
\newblock In \emph{The Thirteenth International Conference on Learning Representations}, 2025{\natexlab{b}}.
\newblock \url{https://openreview.net/forum?id=KMCJXjlDDr}.

\bibitem[Liu et~al.(2025{\natexlab{c}})Liu, Qin, Shi, Chen, Yang, Huang, Wang, and Long]{das2024sundial}
Yong Liu, Guo Qin, Zhiyuan Shi, Zhi Chen, Caiyin Yang, Xiangdong Huang, Jianmin Wang, and Mingsheng Long.
\newblock Sundial: A family of highly capable time series foundation models.
\newblock In \emph{Forty-second International Conference on Machine Learning}, 2025{\natexlab{c}}.
\newblock \url{https://openreview.net/forum?id=LO7ciRpjI5}.

\bibitem[Liu et~al.(2026{\natexlab{b}})Liu, Su, Wang, Zhang, Liu, Wang, Ye, Xiang, Wang, and Long]{liu2026timers1}
Yong Liu, Xingjian Su, Shiyu Wang, Haoran Zhang, Haixuan Liu, Yuxuan Wang, Zhou Ye, Yang Xiang, Jianmin Wang, and Mingsheng Long.
\newblock Timer-s1: A billion-scale time series foundation model with serial scaling.
\newblock \emph{arXiv preprint arXiv:2603.04791}, 2026{\natexlab{b}}.

\bibitem[Loshchilov and Hutter(2017)]{loshchilov2017decoupled}
Ilya Loshchilov and Frank Hutter.
\newblock Decoupled weight decay regularization.
\newblock \emph{arXiv preprint arXiv:1711.05101}, 2017.

\bibitem[Makridakis et~al.(2018)Makridakis, Spiliotis, and Assimakopoulos]{Makridakis2018TheMC}
Spyros Makridakis, Evangelos Spiliotis, and Vassilios Assimakopoulos.
\newblock The {M}4 competition: Results, findings, conclusion and way forward.
\newblock \emph{International Journal of Forecasting}, 2018.

\bibitem[Makridakis et~al.(2022)Makridakis, Spiliotis, and Assimakopoulos]{makridakis2022m5}
Spyros Makridakis, Evangelos Spiliotis, and Vassilios Assimakopoulos.
\newblock M5 accuracy competition: Results, findings, and conclusions.
\newblock \emph{International Journal of Forecasting}, 38\penalty0 (4):\penalty0 1346--1364, 2022.
\newblock ISSN 0169-2070.
\newblock \doi{https://doi.org/10.1016/j.ijforecast.2021.11.013}.
\newblock \url{https://www.sciencedirect.com/science/article/pii/S0169207021001874}.
\newblock Special Issue: M5 competition.

\bibitem[Mancuso et~al.(2021)Mancuso, Piccialli, and Sudoso]{Mancuso2020AML}
Paolo Mancuso, Veronica Piccialli, and Antonio~M Sudoso.
\newblock A machine learning approach for forecasting hierarchical time series.
\newblock \emph{Expert Systems with Applications}, 182:\penalty0 115102, 2021.

\bibitem[McCracken and Ng(2016)]{McCracken01102016}
Michael~W. McCracken and Serena Ng.
\newblock F{RED-MD}: A monthly database for macroeconomic research.
\newblock \emph{Journal of Business \& Economic Statistics}, 34\penalty0 (4):\penalty0 574--589, 2016.
\newblock \doi{10.1080/07350015.2015.1086655}.
\newblock \url{https://doi.org/10.1080/07350015.2015.1086655}.

\bibitem[McCracken and Ng(2021)]{mccracken2020fred}
Michael~W. McCracken and Serena Ng.
\newblock F{RED-QD}: A quarterly database for macroeconomic research.
\newblock \emph{Review}, 103\penalty0 (1):\penalty0 1--44, January 2021.
\newblock \doi{10.20955/r.103.1-44}.
\newblock \url{https://ideas.repec.org/a/fip/fedlrv/90588.html}.

\bibitem[MichalKecera(2024)]{RohlikSalesForecasting2024}
MichalKecera.
\newblock Rohlik sales forecasting challenge.
\newblock \url{https://kaggle.com/competitions/rohlik-sales-forecasting-challenge-v2}, 2024.
\newblock Kaggle.

\bibitem[Mohaddes and Raissi(2024)]{mohaddes2024gvar}
Kamiar Mohaddes and Mehdi Raissi.
\newblock Compilation, revision and updating of the global var (gvar) database.
\newblock Mendeley Data, Version 1, 2024.
\newblock \url{https://doi.org/10.17632/kfp5fhgkvf.1}.

\bibitem[Nie et~al.(2023)Nie, H.~Nguyen, Sinthong, and Kalagnanam]{nie2023patchtst}
Yuqi Nie, Nam H.~Nguyen, Phanwadee Sinthong, and Jayant Kalagnanam.
\newblock A time series is worth 64 words: Long-term forecasting with transformers.
\newblock In \emph{International Conference on Learning Representations}, 2023.

\bibitem[of~Health~Affairs and Ministry~of Health(2024)]{Kaggle2025RiyadhHospitalAdmissions}
General~Directorate of~Health~Affairs and Saudi~Arabia Ministry~of Health.
\newblock Riyadh hospital admissions dataset (2020–2024).
\newblock \url{https://www.kaggle.com/dsv/9992619}, 2024.

\bibitem[Palaskar et~al.(2024)Palaskar, Ekambaram, Jati, Gantayat, Saha, Nagar, Nguyen, Dayama, Sindhgatta, Mohapatra, Kumar, Kalagnanam, Hemachandra, and Rangaraj]{automixer}
Santosh Palaskar, Vijay Ekambaram, Arindam Jati, Neelamadhav Gantayat, Avirup Saha, Seema Nagar, Nam Nguyen, Pankaj Dayama, Renuka Sindhgatta, Prateeti Mohapatra, Harshit Kumar, Jayant Kalagnanam, Nandyala Hemachandra, and Narayan Rangaraj.
\newblock Automixer for improved multivariate time-series forecasting on business and it observability data.
\newblock \emph{Proceedings of the AAAI Conference on Artificial Intelligence}, 38:\penalty0 22962--22968, 2024.

\bibitem[Podest et~al.(2026)Podest, Pichler, B{\"u}rger, Z{\'o}lyomi, Voggenberger, Berghammer, Klotz, B{\"o}ck, Klambauer, and Hochreiter]{podest2026tirex2}
Patrick Podest, Marco Pichler, Elias B{\"u}rger, Levente Z{\'o}lyomi, Bernhard Voggenberger, Wilhelm Berghammer, Daniel Klotz, Sebastian B{\"o}ck, G{\"u}nter Klambauer, and Sepp Hochreiter.
\newblock Tirex-2: Generalizing tirex to multivariate data and streaming.
\newblock \emph{arXiv preprint arXiv:2607.01204}, 2026.

\bibitem[Qiu et~al.(2026)Qiu, Wang, Zheng, Huang, Wen, Yang, Men, Yu, Huang, Huang, Liu, Zhou, and Lin]{qiu2026gated}
Zihan Qiu, Zekun Wang, Bo~Zheng, Zeyu Huang, Kaiyue Wen, Songlin Yang, Rui Men, Le~Yu, Fei Huang, Suozhi Huang, Dayiheng Liu, Jingren Zhou, and Junyang Lin.
\newblock Gated attention for large language models: Non-linearity, sparsity, and attention-sink-free.
\newblock In \emph{The Thirty-ninth Annual Conference on Neural Information Processing Systems}, 2026.
\newblock \url{https://openreview.net/forum?id=1b7whO4SfY}.

\bibitem[Shao et~al.(2024)Shao, Wang, Xu, Wei, Yu, Zhang, Yao, Sun, Jin, Cao, et~al.]{shao2024exploring}
Zezhi Shao, Fei Wang, Yongjun Xu, Wei Wei, Chengqing Yu, Zhao Zhang, Di~Yao, Tao Sun, Guangyin Jin, Xin Cao, et~al.
\newblock Exploring progress in multivariate time series forecasting: Comprehensive benchmarking and heterogeneity analysis.
\newblock \emph{IEEE Transactions on Knowledge and Data Engineering}, 37\penalty0 (1):\penalty0 291--305, 2024.

\bibitem[Shao et~al.(2025)Shao, Yu, and Wang]{shao2025heterogeneity}
Zezhi Shao, Chengqing Yu, and Fei Wang.
\newblock Heterogeneity in multivariate time series: Comprehensive analysis and adaptive modeling.
\newblock In \emph{Proceedings of the 19th International Symposium on Spatial and Temporal Data}, pages 76--79, 2025.

\bibitem[Shazeer(2020)]{shazeer2020swiglu}
Noam Shazeer.
\newblock Glu variants improve transformer.
\newblock \emph{arXiv preprint arXiv:2002.05202}, 2020.

\bibitem[Shchur et~al.(2025)Shchur, Ansari, Turkmen, Stella, Erickson, Guerron, Bohlke-Schneider, and Wang]{shchur2025fev}
Oleksandr Shchur, Abdul~Fatir Ansari, Caner Turkmen, Lorenzo Stella, Nick Erickson, Pablo Guerron, Michael Bohlke-Schneider, and Yuyang Wang.
\newblock fev-bench: A realistic benchmark for time series forecasting.
\newblock \emph{arXiv preprint arXiv:2509.26468}, 2025.

\bibitem[Shen et~al.(2015)Shen, Van~Beek, and Iosup]{grid_workloads_archive}
Siqi Shen, Vincent Van~Beek, and Alexandru Iosup.
\newblock Statistical characterization of business-critical workloads hosted in cloud datacenters.
\newblock In \emph{IEEE/ACM International Symposium on Cluster, Cloud and Grid Computing}, pages 465--474. IEEE, 2015.

\bibitem[Shi et~al.(2025)Shi, Wang, Nie, Li, Ye, Wen, and Jin]{shi2024timemoe}
Xiaoming Shi, Shiyu Wang, Yuqi Nie, Dianqi Li, Zhou Ye, Qingsong Wen, and Ming Jin.
\newblock Time-moe: Billion-scale time series foundation models with mixture of experts.
\newblock In \emph{The Thirteenth International Conference on Learning Representations}, 2025.
\newblock \url{https://openreview.net/forum?id=e1wDDFmlVu}.

\bibitem[Shoeybi et~al.(2019)Shoeybi, Patwary, Puri, LeGresley, Casper, and Catanzaro]{shoeybi2019megatron}
Mohammad Shoeybi, Mostofa Patwary, Raul Puri, Patrick LeGresley, Jared Casper, and Bryan Catanzaro.
\newblock Megatron-lm: Training multi-billion parameter language models using model parallelism.
\newblock \emph{arXiv preprint arXiv:1909.08053}, 2019.

\bibitem[Stewart and Sun(1990)]{stewart1990matrix}
Gilbert~W Stewart and Ji-guang Sun.
\newblock Matrix perturbation theory.
\newblock 1990.

\bibitem[Su et~al.(2024)Su, Ahmed, Lu, Pan, Bo, and Liu]{su2024roformer}
Jianlin Su, Murtadha Ahmed, Yu~Lu, Shengfeng Pan, Wen Bo, and Yunfeng Liu.
\newblock Roformer: Enhanced transformer with rotary position embedding.
\newblock \emph{Neurocomputing}, 568:\penalty0 127063, 2024.

\bibitem[Syed et~al.(2026)Syed, Ahamed, and Wasi]{syed2026position}
Md~Asif~Bin Syed, Md~Younus Ahamed, and Azmine~Toushik Wasi.
\newblock Position: Time-series foundation models require explicit domain-level benchmarks.
\newblock In \emph{Forty-third International Conference on Machine Learning Position Paper Track}, 2026.
\newblock \url{https://openreview.net/forum?id=W2eEMPjzIQ}.

\bibitem[Trindade(2015)]{electricityloaddiagrams20112014_321}
Artur Trindade.
\newblock {ElectricityLoadDiagrams20112014}.
\newblock UCI Machine Learning Repository, 2015.
\newblock {DOI}: https://doi.org/10.24432/C58C86.

\bibitem[van Renen et~al.(2024)van Renen, Horn, Pfeil, Vaidya, Dong, Narayanaswamy, Liu, Saxena, Kipf, and Kraska]{renen2024redset}
Alexander van Renen, Dominik Horn, Pascal Pfeil, Kapil Vaidya, Wenjian Dong, Murali Narayanaswamy, Zhengchun Liu, Gaurav Saxena, Andreas Kipf, and Tim Kraska.
\newblock Why {TPC} is not enough: An analysis of the amazon redshift fleet.
\newblock \emph{Proc. VLDB Endow.}, 17\penalty0 (11):\penalty0 3694–3706, July 2024.
\newblock ISSN 2150-8097.
\newblock \doi{10.14778/3681954.3682031}.
\newblock \url{https://doi.org/10.14778/3681954.3682031}.

\bibitem[Wang et~al.(2023)Wang, Jiang, Jiang, Han, and Zhao]{wang2023libcity}
Jingyuan Wang, Jiawei Jiang, Wenjun Jiang, Chengkai Han, and Wayne~Xin Zhao.
\newblock Towards efficient and comprehensive urban spatial-temporal prediction: A unified library and performance benchmark.
\newblock \emph{arXiv preprint arXiv:2304.14343}, 2023.

\bibitem[Woo et~al.(2024)Woo, Liu, Kumar, Xiong, Savarese, and Sahoo]{woo2024moirai}
Gerald Woo, Chenghao Liu, Akshat Kumar, Caiming Xiong, Silvio Savarese, and Doyen Sahoo.
\newblock Unified training of universal time series forecasting transformers.
\newblock In \emph{Proceedings of the 41st International Conference on Machine Learning}, ICML'24. JMLR.org, 2024.

\bibitem[Wu et~al.(2021)Wu, Xu, Wang, and Long]{Wu2021AutoformerDT}
Haixu Wu, Jiehui Xu, Jianmin Wang, and Mingsheng Long.
\newblock Autoformer: Decomposition transformers with auto-correlation for long-term series forecasting.
\newblock In \emph{Neural Information Processing Systems}, 2021.
\newblock \url{https://api.semanticscholar.org/CorpusID:235623791}.

\bibitem[Xue et~al.(2026)Xue, Zhu, Zhang, Cai, Wang, Mu, Zhou, Li, Di, and Yu]{xue2026quitobench}
Siqiao Xue, Zhaoyang Zhu, Wei Zhang, Rongyao Cai, Rui Wang, Yixiang Mu, Fan Zhou, Jianguo Li, Peng Di, and Hang Yu.
\newblock Quito{B}ench: A high-quality open time series forecasting benchmark.
\newblock \emph{arXiv preprint arXiv:2603.26017}, 2026.

\bibitem[Yeh et~al.(2023)Yeh, Dai, Chen, Zheng, Fan, Der, Lai, Zhuang, Wang, Wang, et~al.]{yeh2023toward}
Chin-Chia~Michael Yeh, Xin Dai, Huiyuan Chen, Yan Zheng, Yujie Fan, Audrey Der, Vivian Lai, Zhongfang Zhuang, Junpeng Wang, Liang Wang, et~al.
\newblock Toward a foundation model for time series data.
\newblock In \emph{Proceedings of the 32nd ACM International Conference on Information and Knowledge Management}, pages 4400--4404, 2023.

\bibitem[Yu et~al.(2026)Yu, Maddix, Han, Zhang, Ansari, Shchur, Faloutsos, Wilson, Mahoney, and Wang]{yu2026understanding}
Annan Yu, Danielle~C. Maddix, Boran Han, Xiyuan Zhang, Abdul~Fatir Ansari, Oleksandr Shchur, Christos Faloutsos, Andrew~Gordon Wilson, Michael~W. Mahoney, and Bernie Wang.
\newblock Understanding transformers for time series: Rank structure, flow-of-ranks, and compressibility.
\newblock In \emph{The Fourteenth International Conference on Learning Representations}, 2026.
\newblock \url{https://openreview.net/forum?id=axR2KZwaD3}.

\bibitem[Zhang et~al.(2022)Zhang, Zhao, Tsiligkaridis, and Zitnik]{zhang2022self}
Xiang Zhang, Ziyuan Zhao, Theodoros Tsiligkaridis, and Marinka Zitnik.
\newblock Self-supervised contrastive pre-training for time series via time-frequency consistency.
\newblock \emph{Advances in neural information processing systems}, 35:\penalty0 3988--4003, 2022.

\bibitem[Zhou et~al.(2021)Zhou, Zhang, Peng, Zhang, Li, Xiong, and Zhang]{zhou2021informer}
Haoyi Zhou, Shanghang Zhang, Jieqi Peng, Shuai Zhang, Jianxin Li, Hui Xiong, and Wancai Zhang.
\newblock Informer: Beyond efficient transformer for long sequence time-series forecasting.
\newblock \emph{AAAI}, 2021.

\bibitem[Zhou et~al.(2024)Zhou, Lu, Xiao, Tang, Su, Li, Liu, Lyu, Ma, and Dou]{zhou2022sdwpf}
Jingbo Zhou, Xinjiang Lu, Yixiong Xiao, Jian Tang, Jiantao Su, Yu~Li, Ji~Liu, Junfu Lyu, Yanjun Ma, and Dejing Dou.
\newblock S{DWPF}: A dataset for spatial dynamic wind power forecasting over a large turbine array.
\newblock \emph{Scientific Data}, 11\penalty0 (1):\penalty0 649, 2024.
\newblock \doi{10.1038/s41597-024-03427-5}.
\newblock \url{https://doi.org/10.1038/s41597-024-03427-5}.

\bibitem[Zhou et~al.(2022)Zhou, Ma, Wen, Wang, Sun, and Jin]{zhou2022fedformer}
Tian Zhou, Ziqing Ma, Qingsong Wen, Xue Wang, Liang Sun, and Rong Jin.
\newblock Fedformer: Frequency enhanced decomposed transformer for long-term series forecasting.
\newblock In \emph{International conference on machine learning}, pages 27268--27286. PMLR, 2022.

\end{thebibliography}

\newpage
\appendix

\section{Data Sources and Benchmark Specifications}

This appendix documents the data interfaces behind the experiments. For pre-training, the relevant unit is a source collection and its contribution to the overall corpus. For evaluation, the relevant unit is a forecasting task specified by a dataset, sampling frequency, prediction horizon, and set of evaluation windows. The following sections therefore separate corpus composition from benchmark construction.

\subsection{Pre-training Data Inventory}
\label{sec:appendix_pre-training_corpus}

The observed portion of the pre-training corpus is assembled from three complementary releases. The GIFT-Eval pre-training split~\citep{aksu2024gifteval}\footnote{\url{https://huggingface.co/datasets/Salesforce/GiftEvalPretrain}} supplies broad coverage across application domains and temporal resolutions; the Chronos training collection~\citep{ansari2024chronos}\footnote{\url{https://huggingface.co/datasets/autogluon/chronos_datasets}} contributes several high-volume public forecasting datasets; and the Quito corpus~\citep{xue2026quitobench}\footnote{\url{https://huggingface.co/datasets/hq-bench/quito-corpus}} adds production application-traffic traces at two granularities. Table~\ref{tab:pretrain_inventory} keeps these source boundaries explicit while reporting the frequency, number of series, variable count, and total observations of every entry.

Two Chronos procedures further augment this pool~\citep{ansari2024chronos}. TSMixup first rescales sampled series and then combines them, allowing temporal motifs drawn from different datasets to appear in new compositions. KernelSynth expands the corpus in a different direction by sampling functions from randomly composed Gaussian-process kernels. The former recombines patterns already present in the public data, whereas the latter introduces controlled trend, smoothness, and periodic structures without requiring an additional observed dataset.

\begin{table}[p]
\centering
\captionsetup{skip=4pt}
\caption{Inventory of the 70 real-world datasets used for pre-training. Rows are organized by the public collection from which each dataset was obtained; frequencies and counts follow the released metadata.}
\label{tab:pretrain_inventory}
\begin{threeparttable}
\fontsize{7.3}{7.8}\selectfont
\setlength{\tabcolsep}{4pt}
\renewcommand{\arraystretch}{0.98}
\begin{tabular*}{\textwidth}{@{\extracolsep{\fill}} l l r r r l @{} }
\toprule
\textbf{Dataset} & \textbf{Freq.} & \textbf{Series} & \textbf{Variables} & \textbf{Time points} & \textbf{Domain} \\
\midrule
\rowcolor{blue!8}\multicolumn{6}{@{}l}{\textbf{GIFT-Eval pre-training collection}} \\
\texttt{BDG-2} & H & 611 & 1 & 9,454,968 & Energy \\
\texttt{BEIJING\_SUBWAY\_30MIN} & 30T & 276 & 2 & 433,872 & Transport \\
\texttt{CIF 2016} & M & 72 & 1 & 6,334 & Finance \\
\texttt{CMIP6} & 6H & 270,336 & 53 & 1,973,452,800 & Nature \\
\texttt{ERA5} & H & 245,760 & 45 & 2,146,959,360 & Nature \\
\texttt{HZMETRO} & 15T & 80 & 2 & 190,160 & Transport \\
\texttt{LOS\_LOOP} & 5T & 207 & 1 & 7,094,304 & Transport \\
\texttt{LargeST} & 5T & 42,333 & 1 & 4,452,510,528 & Transport \\
\texttt{M1} & A, M, Q & 921 & 1 & 57,882 & Finance \\
\texttt{M3} & A, M, Q & 3,003 & 1 & 209,114 & Finance \\
\texttt{NN5} & D, W & 222 & 1 & 93,240 & Finance \\
\texttt{PEMS03} & 5T & 358 & 1 & 9,382,464 & Transport \\
\texttt{PEMS04} & 5T & 307 & 3 & 5,216,544 & Transport \\
\texttt{PEMS07} & 5T & 883 & 1 & 24,921,792 & Transport \\
\texttt{PEMS08} & 5T & 170 & 3 & 3,035,520 & Transport \\
\texttt{PEMS\_BAY} & 5T & 325 & 1 & 16,941,600 & Transport \\
\texttt{Q-TRAFFIC} & 15T & 45,148 & 1 & 264,386,688 & Transport \\
\texttt{Residential Power} & T & 504 & 3 & 271,333,509 & Energy \\
\texttt{SHMETRO} & 15T & 288 & 2 & 2,536,992 & Transport \\
\texttt{Tourism} & A, M, Q & 1,212 & 1 & 150,822 & Finance \\
\texttt{Traffic} & H, W & 1,724 & 1 & 15,060,864 & Transport \\
\texttt{Uber TLC} & D, H & 524 & 1 & 1,176,531 & Transport \\
\texttt{alibaba\_cluster\_trace\_2018} & 5T & 58,409 & 2 & 95,192,530 & Web \\
\texttt{australian\_electricity\_demand} & 30T & 5 & 1 & 1,153,584 & Energy \\
\texttt{azure\_vm\_traces\_2017} & 5T & 159,472 & 1 & 885,522,908 & Web \\
\texttt{beijing\_air\_quality} & H & 12 & 11 & 420,768 & Nature \\
\texttt{bitcoin\_with\_missing} & D & 18 & 1 & 81,918 & Finance \\
\texttt{borealis} & H & 15 & 1 & 83,269 & Energy \\
\texttt{borg\_cluster\_data\_2011} & 5T & 143,386 & 2 & 537,552,854 & Web \\
\texttt{buildings\_900k} & H & 1,792,328 & 1 & 15,702,585,608 & Energy \\
\texttt{bull} & H & 41 & 1 & 719,304 & Energy \\
\texttt{cdc\_fluview\_ilinet} & W & 75 & 5 & 63,903 & Healthcare \\
\texttt{cdc\_fluview\_who\_nrevss} & W & 74 & 4 & 41,760 & Healthcare \\
\texttt{china\_air\_quality} & H & 437 & 6 & 5,739,234 & Nature \\
\texttt{cockatoo} & H & 1 & 1 & 17,544 & Energy \\
\texttt{covid19\_energy} & H & 1 & 1 & 31,912 & Energy \\
\texttt{covid\_mobility} & D & 362 & 1 & 148,602 & Transport \\
\texttt{elecdemand} & 30T & 1 & 1 & 17,520 & Energy \\
\texttt{elf} & H & 1 & 1 & 21,792 & Energy \\
\texttt{extended\_web\_traffic\_with\_missing} & D & 145,063 & 1 & 370,926,091 & Web \\
\texttt{godaddy} & M & 3,135 & 2 & 128,535 & Finance \\
\texttt{hog} & H & 24 & 1 & 421,056 & Energy \\
\texttt{ideal} & H & 217 & 1 & 1,255,253 & Energy \\
\texttt{kaggle\_web\_traffic\_weekly} & W & 145,063 & 1 & 16,537,182 & Web \\
\texttt{lcl} & H & 713 & 1 & 9,543,553 & Energy \\
\texttt{london\_smart\_meters\_with\_missing} & 30T & 5,520 & 1 & 166,238,880 & Energy \\
\texttt{oikolab\_weather} & H & 8 & 1 & 800,456 & Nature \\
\texttt{pdb} & H & 1 & 1 & 17,520 & Energy \\
\texttt{pedestrian\_counts} & H & 66 & 1 & 3,130,762 & Transport \\
\texttt{project\_tycho} & W & 1,258 & 1 & 1,377,707 & Healthcare \\
\texttt{rideshare\_with\_missing} & H & 2,304 & 1 & 859,392 & Transport \\
\texttt{sceaux} & H & 1 & 1 & 34,223 & Energy \\
\texttt{smart} & H & 5 & 1 & 95,709 & Energy \\
\texttt{solar\_power} & 4S & 1 & 1 & 7,397,222 & Energy \\
\texttt{spain} & H & 1 & 1 & 35,064 & Energy \\
\texttt{subseasonal} & D & 862 & 4 & 14,197,140 & Nature \\
\texttt{subseasonal\_precip} & D & 862 & 1 & 9,760,426 & Nature \\
\texttt{sunspot\_with\_missing} & D & 1 & 1 & 73,894 & Nature \\
\texttt{vehicle\_trips\_with\_missing} & D & 329 & 1 & 32,512 & Transport \\
\texttt{weather} & D & 3,010 & 1 & 42,941,700 & Nature \\
\texttt{wiki-rolling\_nips} & D & 47,675 & 1 & 40,619,100 & Web \\
\texttt{wind\_power} & 4S & 1 & 1 & 7,397,147 & Energy \\
\rowcolor{blue!8}\multicolumn{6}{@{}l}{\textbf{Chronos training collection}} \\
\texttt{Solar} & 5T & 5,166 & 1 & 543,049,920 & Energy \\
\texttt{Taxi} & 30T, H & 70,412 & 1 & 56,793,348 & Transport \\
\texttt{Weatherbench} & D, H, W & 675,840 & 1 & 82,753,646,592 & Nature \\
\texttt{Wind Farms} & D, H, T & 1,011 & 1 & 175,154,333 & Energy \\
\texttt{dominick} & W & 100,014 & 1 & 29,652,492 & Sales \\
\texttt{exchange\_rate} & D & 8 & 1 & 84,976 & Finance \\
\texttt{mexico\_city\_bikes} & H & 494 & 1 & 38,687,004 & Transport \\
\texttt{ushcn\_daily} & D & 1,218 & 5 & 47,080,115 & Nature \\
\texttt{wiki\_daily\_100k} & D & 100,000 & 1 & 274,100,000 & Web \\
\rowcolor{blue!8}\multicolumn{6}{@{}l}{\textbf{Quito corpus}} \\
\texttt{Quito} & 10T, H & 33,806 & 5 & 313,269,828 & Various \\
\bottomrule
\end{tabular*}
\begin{tablenotes}[flushleft]
\fontsize{7.3}{7.8}\selectfont
\item Frequency aliases: S = second, T = minute, H = hourly, D = daily, W = weekly, M = monthly, Q = quarterly, and A = annual.
\end{tablenotes}
\end{threeparttable}
\end{table}

\subsection{GIFT-Eval Task Composition}
\label{sec:appendix_gift-eval}

GIFT-Eval is organized as a grid of forecasting conditions rather than as a single pooled test set~\citep{aksu2024gifteval}. Its 23 datasets are evaluated at the available sampling resolutions and, where applicable, at short-, medium-, and long-range horizons. This expansion produces 97 dataset--frequency--horizon configurations spanning seven domains and ten frequencies. Table~\ref{tab:gift-eval} exposes the construction in two layers: the data-profile columns describe each dataset--frequency instance, while the evaluation columns give the horizon and number of forecast windows attached to it.

The protocol evaluates the final 10\% of each series through non-overlapping rolling windows. Standard competition horizons are retained for collections such as M4~\citep{Makridakis2018TheMC}; other datasets receive horizon scales chosen according to their frequency and application setting. Results are then combined across configurations, so comparison is performed at the level of forecasting conditions rather than by pooling observations from datasets of very different sizes.

The resulting evaluation unit is therefore more specific than a dataset name. A series resampled at two frequencies produces two data profiles, and a short, medium, or long horizon attached to either profile defines a separate forecasting condition. Table~\ref{tab:gift-eval} makes this distinction visible: the main body records one profile for each available frequency, while the bracketed values report how many valid rolling origins are available at the corresponding horizon. This representation avoids making a high-resolution dataset with many observations automatically more influential than a smaller source. It also shows where the benchmark retains a single application-specific horizon and where it probes the same data over multiple forecast ranges.

\renewcommand{\arraystretch}{1.02}
\newcommand{\pw}[2]{#1\,[#2]}
\begin{table}[p]
\caption{GIFT-Eval task specifications grouped by application domain. Short, Medium, and Long report forecast horizon [number of rolling windows]; a dash indicates that the range is not evaluated.}
\label{tab:gift-eval}
\centering
\fontsize{8.2}{9.0}\selectfont
\setlength{\tabcolsep}{3pt}
\begin{tabular*}{\textwidth}{@{\extracolsep{\fill}}lcrrcrrr@{}}
\toprule
& & \multicolumn{3}{c}{\textbf{Data profile}} & \multicolumn{3}{c}{\textbf{Evaluation schedule}} \\
\cmidrule(lr){3-5}\cmidrule(l){6-8}
\textbf{Dataset} & \textbf{Freq.} & \textbf{Series} & \textbf{Avg. length} & \textbf{Variables} & \textbf{Short} & \textbf{Medium} & \textbf{Long} \\
\midrule

\rowcolor{blue!8}\multicolumn{8}{@{}l}{\textbf{Nature}} \\
Jena Weather & 10T & 1 & 52,704 & 21 & \pw{48}{20} & \pw{480}{11} & \pw{720}{8} \\
 & H & 1 & 8,784 & 21 & \pw{48}{19} & \pw{480}{2} & \pw{720}{2} \\
 & D & 1 & 366 & 21 & \pw{30}{2} & -- & -- \\

Saugeen & D & 1 & 23,741 & 1 & \pw{30}{20} & -- & -- \\
 & W-THU & 1 & 3,391 & 1 & \pw{8}{20} & -- & -- \\
 & M & 1 & 780 & 1 & \pw{12}{7} & -- & -- \\
Temperature Rain & D & 32,072 & 725 & 1 & \pw{30}{3} & -- & -- \\
KDD Cup 2018 & H & 270 & 10,898 & 1 & \pw{48}{20} & \pw{480}{2} & \pw{720}{2} \\
 & D & 270 & 455 & 1 & \pw{30}{2} & -- & -- \\

\rowcolor{blue!8}\multicolumn{8}{@{}l}{\textbf{Web/CloudOps}} \\
BizITObs - Application & 10S & 1 & 8,834 & 2 & \pw{60}{15} & \pw{600}{2} & \pw{900}{1} \\
BizITObs - Service & 10S & 21 & 8,835 & 2 & \pw{60}{15} & \pw{600}{2} & \pw{900}{1} \\
BizITObs - L2C & 5T & 1 & 31,968 & 7 & \pw{48}{20} & \pw{480}{7} & \pw{720}{5} \\
 & H & 1 & 2,664 & 7 & \pw{48}{6} & \pw{480}{1} & \pw{720}{1} \\
Bitbrains - Fast Storage & 5T & 1,250 & 8,640 & 2 & \pw{48}{18} & \pw{480}{2} & \pw{720}{2} \\
 & H & 1,250 & 721 & 2 & \pw{48}{2} & -- & -- \\
Bitbrains - rnd & 5T & 500 & 8,640 & 2 & \pw{48}{18} & \pw{480}{2} & \pw{720}{2} \\
 & H & 500 & 720 & 2 & \pw{48}{2} & -- & -- \\

\rowcolor{blue!8}\multicolumn{8}{@{}l}{\textbf{Energy}} \\
ETT1 & 15T & 1 & 69,680 & 7 & \pw{48}{20} & \pw{480}{15} & \pw{720}{10} \\
 & H & 1 & 17,420 & 7 & \pw{48}{20} & \pw{480}{4} & \pw{720}{3} \\
 & D & 1 & 725 & 7 & \pw{30}{3} & -- & -- \\
 & W-THU & 1 & 103 & 7 & \pw{8}{2} & -- & -- \\
ETT2 & 15T & 1 & 69,680 & 7 & \pw{48}{20} & \pw{480}{15} & \pw{720}{10} \\
 & H & 1 & 17,420 & 7 & \pw{48}{20} & \pw{480}{4} & \pw{720}{3} \\
 & D & 1 & 725 & 7 & \pw{30}{3} & -- & -- \\
 & W-THU & 1 & 103 & 7 & \pw{8}{2} & -- & -- \\
Solar & 10T & 137 & 52,560 & 1 & \pw{48}{20} & \pw{480}{11} & \pw{720}{8} \\
 & H & 137 & 8,760 & 1 & \pw{48}{19} & \pw{480}{2} & \pw{720}{2} \\
 & D & 137 & 365 & 1 & \pw{30}{2} & -- & -- \\
 & W-FRI & 137 & 52 & 1 & \pw{8}{1} & -- & -- \\
Electricity & 15T & 370 & 140,256 & 1 & \pw{48}{20} & \pw{480}{20} & \pw{720}{20} \\
 & H & 370 & 35,064 & 1 & \pw{48}{20} & \pw{480}{8} & \pw{720}{5} \\
 & D & 370 & 1,461 & 1 & \pw{30}{5} & -- & -- \\
 & W-FRI & 370 & 208 & 1 & \pw{8}{3} & -- & -- \\

\rowcolor{blue!8}\multicolumn{8}{@{}l}{\textbf{Transport}} \\
Loop Seattle & 5T & 323 & 105,120 & 1 & \pw{48}{20} & \pw{480}{20} & \pw{720}{15} \\
 & H & 323 & 8,760 & 1 & \pw{48}{19} & \pw{480}{2} & \pw{720}{2} \\
 & D & 323 & 365 & 1 & \pw{30}{2} & -- & -- \\
SZ-Taxi & 15T & 156 & 2,976 & 1 & \pw{48}{7} & \pw{480}{1} & \pw{720}{1} \\
 & H & 156 & 744 & 1 & \pw{48}{2} & -- & -- \\
M\_DENSE & H & 30 & 17,520 & 1 & \pw{48}{20} & \pw{480}{4} & \pw{720}{3} \\
 & D & 30 & 730 & 1 & \pw{30}{3} & -- & -- \\

\rowcolor{blue!8}\multicolumn{8}{@{}l}{\textbf{Sales}} \\
Restaurant & D & 807 & 358 & 1 & \pw{30}{1} & -- & -- \\
Hierarchical Sales & D & 118 & 1,825 & 1 & \pw{30}{7} & -- & -- \\
 & W-WED & 118 & 260 & 1 & \pw{8}{4} & -- & -- \\
Car Parts & M & 2,674 & 51 & 1 & \pw{12}{1} & -- & -- \\

\rowcolor{blue!8}\multicolumn{8}{@{}l}{\textbf{Econ/Fin}} \\
M4 Yearly & A & 22,974 & 37 & 1 & \pw{6}{1} & -- & -- \\
M4 Quarterly & Q & 24,000 & 100 & 1 & \pw{8}{1} & -- & -- \\
M4 Monthly & M & 48,000 & 234 & 1 & \pw{18}{1} & -- & -- \\
M4 Weekly & W & 359 & 1,035 & 1 & \pw{13}{1} & -- & -- \\
M4 Daily & D & 4,227 & 2,371 & 1 & \pw{14}{1} & -- & -- \\
M4 Hourly & H & 414 & 902 & 1 & \pw{48}{2} & -- & -- \\

\rowcolor{blue!8}\multicolumn{8}{@{}l}{\textbf{Healthcare}} \\
Hospital & M & 767 & 84 & 1 & \pw{12}{1} & -- & -- \\
COVID Deaths & D & 266 & 212 & 1 & \pw{30}{1} & -- & -- \\
US Births & D & 1 & 7,305 & 1 & \pw{30}{20} & -- & -- \\
 & W-TUE & 1 & 1,043 & 1 & \pw{8}{14} & -- & -- \\
 & M & 1 & 240 & 1 & \pw{12}{2} & -- & -- \\

\bottomrule
\end{tabular*}

\end{table}
\FloatBarrier

The evaluation data originate from ten public sources. Environmental and infrastructure measurements include Jena Weather, ETT, Electricity, and Solar~\citep{Wu2021AutoformerDT,zhou2021informer,electricityloaddiagrams20112014_321,Lai2017ModelingLA}; operational workloads are represented by BizITObs and Bitbrains~\citep{automixer,grid_workloads_archive}; and urban mobility data are obtained through LibCity~\citep{wang2023libcity}. Sales, economic, financial, and healthcare tasks are drawn from the Recruit competition, hierarchical-sales data, and the Monash archive~\citep{howard2017recruit,Mancuso2020AML,godahewa2021monash}. The benchmark curation keeps its evaluation sources separate from the pre-training split.

\subsection{fev-bench Task Composition}
\label{sec:appendix_fev-eval}

fev-bench treats a forecasting task as a complete evaluation specification: it fixes the data source, target selection, forecast horizon, and rolling evaluation cutoffs~\citep{shchur2025fev}. The benchmark derives 100 tasks from 96 datasets and emphasizes breadth across applications rather than repeatedly evaluating every dataset at multiple alternative horizons. Competition tasks preserve their published forecast lengths; the remaining horizons follow frequency-aware choices, including week-ahead forecasts for selected hourly datasets. Each task is evaluated at $W$ rolling origins, with $W$ adapted to dataset scale and the amount of history available before the first forecast. Table~\ref{tab:fev-bench} records the resulting task definitions using standard \texttt{pandas} frequency aliases.

Each row in Table~\ref{tab:fev-bench} should consequently be interpreted as an evaluation contract. The frequency fixes the temporal grid, $T$ translates the intended forecast interval into a number of future observations, and $W$ states how many rolling origins contribute to the task-level score. A dataset name may appear in several rows when the source supports distinct sampling resolutions; these rows represent different forecasting settings rather than duplicated measurements. The median-length, series, and target columns further expose the amount of usable history and the panel structure presented to the model.

Rolling origins prevent a single terminal cutoff from determining a task result; $W$ is reduced for shorter records, large panels, or fixed competition splits to keep the evaluation feasible. The source-family and application-domain fields serve different purposes as well: the former preserves dataset provenance, while the shaded groups organize tasks by their operational interpretation. Keeping these two axes separate makes it possible to inspect domain coverage without losing traceability to the benchmark or repository from which a task was constructed.

The source-family column in Table~\ref{tab:fev-bench} reflects three complementary acquisition routes. First, established forecasting resources contribute tasks from GIFT-Eval, Monash, and BOOMLET~\citep{aksu2024gifteval,godahewa2021monash,cohen2026toto}. Second, domain repositories supply macroeconomic and energy measurements, including GVAR, FRED, EPF, ERCOT, and ENTSO-e~\citep{mohaddes2024gvar,McCracken01102016,mccracken2020fred,lago2021forecasting,ansari2024chronos,OPSD2020}. Third, forecasting competitions add fixed-horizon retail and energy problems from Favorita, M5, Rossmann, Walmart, Rohlik, KDD Cup 2022, and the Global Energy Forecasting Competitions~\citep{Kaggle2020StoreSales,makridakis2022m5,Kaggle2015Rossmann,Kaggle2014Walmart,RohlikSalesForecasting2024,zhou2022sdwpf,hong2014global}.

Additional tasks cover public health, environmental monitoring, fashion, and cloud systems through ECDC influenza surveillance, UK COVID-19 statistics, UCI air quality, Hermes, hospital admissions, and Redset~\citep{ECDC2025RespiratoryViruses,Kaggle2022UKCovidDashboard,DeVito2008ElectronicNoseCalibration,david2022hermes,Kaggle2025RiyadhHospitalAdmissions,renen2024redset}. Together, they test robustness across domains, temporal scales, and history lengths.

\clearpage
{
\fontsize{8.1}{9.1}\selectfont
\setlength{\tabcolsep}{7.5pt}
\renewcommand{\arraystretch}{1.22}

\begin{longtable}{@{}p{4.5cm}p{2.4cm}lccrrr@{}}
\caption{fev-bench task specifications arranged by application domain. The source-family column records how each task enters the benchmark; $T$ and $W$ denote the forecast horizon and number of rolling evaluation windows, respectively.}
\label{tab:fev-bench} \\
\toprule
\textbf{Task} & \textbf{Source family} & \textbf{Freq.} & $T$ & $W$ & \textbf{Median length} & \textbf{Series} & \textbf{Targets} \\
\midrule
\endfirsthead

\caption[]{fev-bench task specifications (continued).} \\
\toprule
\textbf{Task} & \textbf{Source family} & \textbf{Freq.} & $T$ & $W$ & \textbf{Median length} & \textbf{Series} & \textbf{Targets} \\
\midrule
\endhead

\midrule
\multicolumn{8}{r}{\textit{Continued on next page}} \\
\endfoot

\bottomrule
\endlastfoot

\rowcolor{blue!8}\multicolumn{8}{@{}l}{\textbf{Cloud}} \\
BizITObs-L2C & GIFT-Eval & 5T & 288 & 20 & 31,968 & 1 & 7 \\
BizITObs-L2C & GIFT-Eval & H & 24 & 20 & 2,664 & 1 & 7 \\
BOOMLET-1062 & BOOMLET & 5T & 288 & 20 & 16,384 & 1 & 21 \\
BOOMLET-1209 & BOOMLET & 5T & 288 & 20 & 16,384 & 1 & 53 \\
BOOMLET-1225 & BOOMLET & T & 60 & 20 & 16,384 & 1 & 49 \\
BOOMLET-1230 & BOOMLET & 5T & 288 & 20 & 16,384 & 1 & 23 \\
BOOMLET-1282 & BOOMLET & T & 60 & 20 & 16,384 & 1 & 35 \\
BOOMLET-1487 & BOOMLET & 5T & 288 & 20 & 16,384 & 1 & 54 \\
BOOMLET-1631 & BOOMLET & 30T & 96 & 20 & 10,463 & 1 & 40 \\
BOOMLET-1676 & BOOMLET & 30T & 96 & 20 & 10,463 & 1 & 100 \\
BOOMLET-1855 & BOOMLET & H & 24 & 20 & 5,231 & 1 & 52 \\
BOOMLET-1975 & BOOMLET & H & 24 & 20 & 5,231 & 1 & 75 \\
BOOMLET-2187 & BOOMLET & H & 24 & 20 & 5,231 & 1 & 100 \\
BOOMLET-285 & BOOMLET & T & 60 & 20 & 16,384 & 1 & 75 \\
BOOMLET-619 & BOOMLET & T & 60 & 20 & 16,384 & 1 & 52 \\
BOOMLET-772 & BOOMLET & T & 60 & 20 & 16,384 & 1 & 67 \\
BOOMLET-963 & BOOMLET & T & 60 & 20 & 16,384 & 1 & 28 \\
Redset & Other & 5T & 288 & 10 & 25,920 & 118 & 1 \\
Redset & Other & 15T & 96 & 10 & 8,640 & 126 & 1 \\
Redset & Other & H & 24 & 10 & 2,160 & 138 & 1 \\

\rowcolor{blue!8}\multicolumn{8}{@{}l}{\textbf{Economy}} \\
Australian Tourism & Macro & Q & 8 & 2 & 36 & 89 & 1 \\
FRED-MD-CEE & Macro & M & 12 & 20 & 798 & 1 & 3 \\
FRED-MD-Macro & Macro & M & 12 & 20 & 798 & 1 & 51 \\
FRED-QD-CEE & Macro & Q & 8 & 20 & 266 & 1 & 3 \\
FRED-QD-Macro & Macro & Q & 8 & 20 & 266 & 1 & 51 \\
GVAR & Macro & Q & 8 & 10 & 178 & 33 & 6 \\
US Consumption & Macro & M & 12 & 10 & 792 & 31 & 1 \\
US Consumption & Macro & Q & 8 & 10 & 262 & 31 & 1 \\
US Consumption & Macro & Y & 5 & 10 & 64 & 31 & 1 \\
World CO2 Emissions & Macro & Y & 5 & 9 & 60 & 191 & 1 \\
World Life Expectancy & Macro & Y & 5 & 10 & 74 & 237 & 1 \\
World Tourism & Macro & Y & 5 & 2 & 21 & 178 & 1 \\

\rowcolor{blue!8}\multicolumn{8}{@{}l}{\textbf{Energy}} \\
ETT & GIFT-Eval & 15T & 96 & 20 & 69,680 & 2 & 7 \\
ETT & GIFT-Eval & H & 168 & 20 & 17,420 & 2 & 7 \\
ETT & GIFT-Eval & D & 28 & 20 & 724 & 2 & 7 \\
ETT & GIFT-Eval & W & 13 & 5 & 103 & 2 & 7 \\
Solar & GIFT-Eval & W & 13 & 1 & 52 & 137 & 1 \\
Solar & GIFT-Eval & D & 28 & 10 & 365 & 137 & 1 \\
ENTSO-e Load & Energy & 15T & 96 & 20 & 175,292 & 6 & 1 \\
ENTSO-e Load & Energy & 30T & 96 & 20 & 87,645 & 6 & 1 \\
ENTSO-e Load & Energy & H & 168 & 20 & 43,822 & 6 & 1 \\
EPF-BE & Energy & H & 24 & 20 & 52,416 & 1 & 1 \\
EPF-DE & Energy & H & 24 & 20 & 52,416 & 1 & 1 \\
EPF-FR & Energy & H & 24 & 20 & 52,416 & 1 & 1 \\
EPF-NP & Energy & H & 24 & 20 & 52,416 & 1 & 1 \\
EPF-PJM & Energy & H & 24 & 20 & 52,416 & 1 & 1 \\
ERCOT & Energy & D & 28 & 20 & 6,452 & 8 & 1 \\
ERCOT & Energy & H & 168 & 20 & 154,872 & 8 & 1 \\
ERCOT & Energy & M & 12 & 15 & 211 & 8 & 1 \\
ERCOT & Energy & W & 13 & 20 & 921 & 8 & 1 \\
GFC12 & Energy & H & 168 & 10 & 39,414 & 11 & 1 \\
GFC14 & Energy & H & 168 & 20 & 17,520 & 1 & 1 \\
GFC17 & Energy & H & 168 & 20 & 17,544 & 8 & 1 \\
Solar with Weather & Energy & 15T & 96 & 20 & 198,600 & 1 & 1 \\
Solar with Weather & Energy & H & 24 & 20 & 49,648 & 1 & 1 \\
KDD Cup 2022 & Competitions & D & 14 & 10 & 243 & 134 & 1 \\
KDD Cup 2022 & Competitions & 10T & 288 & 10 & 35,279 & 134 & 1 \\
KDD Cup 2022 & Competitions & 30T & 96 & 10 & 11,758 & 134 & 1 \\

\rowcolor{blue!8}\multicolumn{8}{@{}l}{\textbf{Healthcare}} \\
Hospital & GIFT-Eval & M & 12 & 4 & 84 & 767 & 1 \\
ECDC ILI & Other & W & 13 & 10 & 201 & 25 & 1 \\
Hospital Admissions & Other & D & 28 & 20 & 1,731 & 8 & 1 \\
Hospital Admissions & Other & W & 13 & 16 & 246 & 8 & 1 \\
UK COVID-Nation-Cumulative & Other & D & 28 & 20 & 729 & 4 & 3 \\
UK COVID-Nation-Cumulative & Other & W & 8 & 4 & 105 & 4 & 3 \\
UK COVID-Nation-New & Other & D & 28 & 20 & 729 & 4 & 3 \\
UK COVID-Nation-New & Other & W & 8 & 4 & 105 & 4 & 3 \\
UK COVID-UTLA-Cumulative & Other & W & 13 & 5 & 104 & 214 & 1 \\
UK COVID-UTLA-New & Other & D & 28 & 10 & 721 & 214 & 1 \\

\rowcolor{blue!8}\multicolumn{8}{@{}l}{\textbf{Mobility}} \\
Loop Seattle & GIFT-Eval & D & 28 & 10 & 365 & 323 & 1 \\
Loop Seattle & GIFT-Eval & 5T & 288 & 10 & 105,120 & 323 & 1 \\
Loop Seattle & GIFT-Eval & H & 168 & 10 & 8,760 & 323 & 1 \\
M-DENSE & GIFT-Eval & D & 28 & 10 & 730 & 30 & 1 \\
M-DENSE & GIFT-Eval & H & 168 & 10 & 17,520 & 30 & 1 \\
SZ Taxi & GIFT-Eval & 15T & 96 & 10 & 2,976 & 156 & 1 \\
SZ Taxi & GIFT-Eval & H & 168 & 2 & 744 & 156 & 1 \\

\rowcolor{blue!8}\multicolumn{8}{@{}l}{\textbf{Nature}} \\
Jena Weather & GIFT-Eval & 10T & 144 & 20 & 52,704 & 1 & 21 \\
Jena Weather & GIFT-Eval & D & 28 & 11 & 366 & 1 & 21 \\
Jena Weather & GIFT-Eval & H & 24 & 20 & 8,784 & 1 & 21 \\
UCI Air Quality & Other & H & 168 & 20 & 9,357 & 1 & 4 \\
UCI Air Quality & Other & D & 28 & 11 & 389 & 1 & 4 \\

\rowcolor{blue!8}\multicolumn{8}{@{}l}{\textbf{Retail}} \\
Hierarchical Sales & GIFT-Eval & D & 28 & 10 & 1,825 & 118 & 1 \\
Hierarchical Sales & GIFT-Eval & W & 13 & 10 & 260 & 118 & 1 \\
Favorita Store Sales & Competitions & M & 12 & 2 & 54 & 1,579 & 1 \\
Favorita Store Sales & Competitions & W & 13 & 10 & 240 & 1,579 & 1 \\
Favorita Store Sales & Competitions & D & 28 & 10 & 1,688 & 1,579 & 1 \\
Favorita Transactions & Competitions & M & 12 & 2 & 54 & 51 & 1 \\
Favorita Transactions & Competitions & W & 13 & 10 & 240 & 51 & 1 \\
Favorita Transactions & Competitions & D & 28 & 10 & 1,688 & 51 & 1 \\
M5 & Competitions & M & 12 & 1 & 58 & 30,490 & 1 \\
M5 & Competitions & W & 13 & 1 & 257 & 30,490 & 1 \\
M5 & Competitions & D & 28 & 1 & 1,810 & 30,490 & 1 \\
Restaurant & Competitions & D & 28 & 8 & 296 & 817 & 1 \\
Rohlik Orders & Competitions & W & 8 & 5 & 170 & 7 & 1 \\
Rohlik Orders & Competitions & D & 61 & 5 & 1,197 & 7 & 1 \\
Rohlik Sales & Competitions & W & 8 & 1 & 150 & 5,243 & 1 \\
Rohlik Sales & Competitions & D & 14 & 1 & 1,046 & 5,390 & 1 \\
Rossmann & Competitions & W & 13 & 8 & 133 & 1,115 & 1 \\
Rossmann & Competitions & D & 48 & 10 & 942 & 1,115 & 1 \\
Walmart & Competitions & W & 39 & 1 & 143 & 2,936 & 1 \\
Hermes & Other & W & 52 & 1 & 261 & 10,000 & 1 \\

\end{longtable}
}
\clearpage

\end{document}